\documentclass[preprint]{elsarticle}

\usepackage{amsmath, amssymb, amsfonts}
\usepackage{amsthm}            
\usepackage{graphicx}          
\usepackage{array}             
\usepackage{multirow}          
\usepackage{booktabs}          
\usepackage{subfig}            
\usepackage{textcomp}          
\usepackage{xcolor}            
\usepackage{threeparttable}    
\usepackage{url}               
\usepackage{hyperref}          
\usepackage{adjustbox} 
\usepackage{comment}
\usepackage{float}
\usepackage{algorithmic}       
\usepackage{enumitem}
\usepackage{algorithm}         
\usepackage[table]{xcolor} 
\usepackage{tikz} 
\usetikzlibrary{positioning}  
\newtheoremstyle{upright}
  {3pt}{3pt}{\normalfont}{}{ \bfseries}{.}{ }{}
\theoremstyle{upright}
\newtheorem{definition}{Definition}
\DeclareMathOperator*{\argmin}{arg\,min}
\usepackage{lineno,hyperref}
\modulolinenumbers[5]
\journal{Engineering Applications of Artificial Intelligence} 

\begin{document}

\begin{frontmatter}
\title{Beyond Prediction: Interval Neural Networks for Uncertainty-Aware System Identification}

\author[itu]{Mehmet Ali Ferah}
\ead{ferah23@itu.edu.tr}

\author[itu]{Tufan Kumbasar}
\ead{kumbasart@itu.edu.tr}
\address[itu]{Artificial Intelligence and Intelligent Systems Laboratory, Istanbul Technical University, Istanbul, Türkiye}

\begin{abstract}
System identification (SysID) is critical for modeling dynamical systems from experimental data, yet traditional approaches often fail to capture nonlinear behaviors. While deep learning offers powerful tools for modeling such dynamics, incorporating uncertainty quantification is essential to ensure reliable predictions. This paper presents a systematic framework for constructing and training interval Neural Networks (INNs) for uncertainty-aware SysID. By extending crisp neural networks into interval counterparts, we develop Interval LSTM and NODE models that propagate uncertainty through interval arithmetic without probabilistic assumptions. This design allows them to represent uncertainty and produce prediction intervals. For training, we propose two strategies: Cascade INN (C-INN), a two-stage approach converting a trained crisp NN into an INN, and Joint INN (J-INN), a one-stage framework jointly optimizing prediction accuracy and interval precision. Both strategies employ uncertainty-aware loss functions and parameterization tricks to ensure reliable learning. Comprehensive experiments on multiple SysID datasets demonstrate the effectiveness of both approaches and benchmark their performance against well-established uncertainty-aware baselines: C-INN achieves superior point prediction accuracy, whereas J-INN yields more accurate and better-calibrated prediction intervals. Furthermore, to reveal how uncertainty is represented across model parameters, the concept of channel-wise elasticity is introduced, which is used to identify distinct patterns across the two training strategies. The results of this study demonstrate that the proposed framework effectively integrates deep learning with uncertainty-aware modeling.
\end{abstract}


\end{frontmatter}

\newpage
\section{Introduction}

System identification (SysID) refers to the data-driven development of mathematical models that describe the behavior of dynamic systems, where the future evolution of the system is determined by its historical dynamics \cite{ljung1999system}. Traditional SysID methods are well-developed but mainly focus on linear system modeling \cite{dankers2016, zhou2006}, which may reduce their ability to capture the nonlinear dynamics encountered in real-world systems. Therefore, there is a crucial need for modeling structures that can capture such nonlinear dynamics. In this regard, integrating deep learning (DL) into SysID has emerged as a powerful approach, since DL can effectively represent complex mappings by learning high-dimensional and nonlinear relationships. Consequently, a growing body of recent studies has explored the combination of DL and SysID \cite{DLsytsid,dai2024deep,mansur2026solis}. Building on the integration of DL into SysID, various neural network (NN) architectures have been investigated for modeling system dynamics. Examples include convolutional NNs\cite{cnn_sysid}, neural state-space models\cite{cont_sysid}, feedforward NNs  \cite{feedforward_sysid}, and recurrent NNs \cite{rnn_sysid}. Among these, Long Short-Term Memory (LSTM) networks \cite{LSTM} and Neural Ordinary Differential Equations (NODEs) \cite{node} have gained particular popularity in SysID, owing to their ability to retain historical information and effectively capture nonlinear dynamics. Both LSTM\cite{tuna2022deep,lstm_sysid} and NODE-based \cite{rahman2022neural,node_sysid} SysID applications have demonstrated their effectiveness. These DL models achieve this by transforming past observations into hidden states, which are then mapped to outputs, enabling them to capture complex nonlinear behaviors more effectively than linear models\cite{saini2025nonlinear}. Furthermore, their complementary characteristics make them particularly appealing. LSTMs are well-suited for discrete-time simulations, while NODEs provide a natural framework for representing continuous-time dynamics \cite{coelho2024enhancing}.

A key limitation of NNs is that their point predictions may lack reliability. To overcome this issue, incorporating Uncertainty Quantification (UQ) is essential \cite{uq2,kabir2021optimal,uq_survey}, as it provides a more comprehensive assessment of prediction reliability by explicitly measuring the uncertainty associated with the model outputs \cite{uq,ak2015interval}. This is especially important in real-world SysID applications, such as aerospace, flight, and robotic control systems, where the reliability of NN predictions is critical \cite{msaddi2025expanding}. Rather than relying solely on point predictions, UQ provides Prediction Intervals (PIs), which define a range within which the true system response is likely to lie \cite{PI,PI2}. This is particularly valuable in SysID tasks, where accurately modeling complex system dynamics is essential; PIs allow for the inclusion of safety margins and help ensure the reliability of the identified models. Accordingly, recent studies have proposed and evaluated UQ methodologies specifically tailored for SysID applications \cite{msaddi2025expanding, guven2025fuzzy}. Popular approaches for UQ often model uncertainty in the learnable parameters (LPs) of NNs by assuming they follow prior distributions \cite{gausslp}. In the context of SysID, Gaussian-based neural networks have been applied to improve prediction reliability \cite{dynogp}. However, predefined probabilistic distributions can constrain model flexibility, potentially limiting generalization. A more general approach that does not rely on a specific probabilistic assumption is therefore preferable. Interval Neural Networks (INNs) offer such an alternative for constructing PIs, providing uncertainty bounds without requiring any probabilistic assumptions \cite{baker1998universal}. Recent studies have demonstrated the applicability of INNs across a variety of domains \cite{intervalbound, intervalimprecise, SADEGHI2019338, oala2020interval,harapanahalli2023toolbox}. In the context of SysID, implementations based on INN architectures have been investigated \cite{ferah2025introducing}.

This paper presents a systematic framework for constructing and training INNs for uncertainty-aware system identification, enabling predictions and UQ in SysID tasks as illustrated in  Fig.~\ref{fig:snap}. The main contributions/ outcomes of this paper are as follows:

\begin{figure}[t]
\centering

\subfloat[Heat Exchanger]{%
  \includegraphics[width=0.48\textwidth]{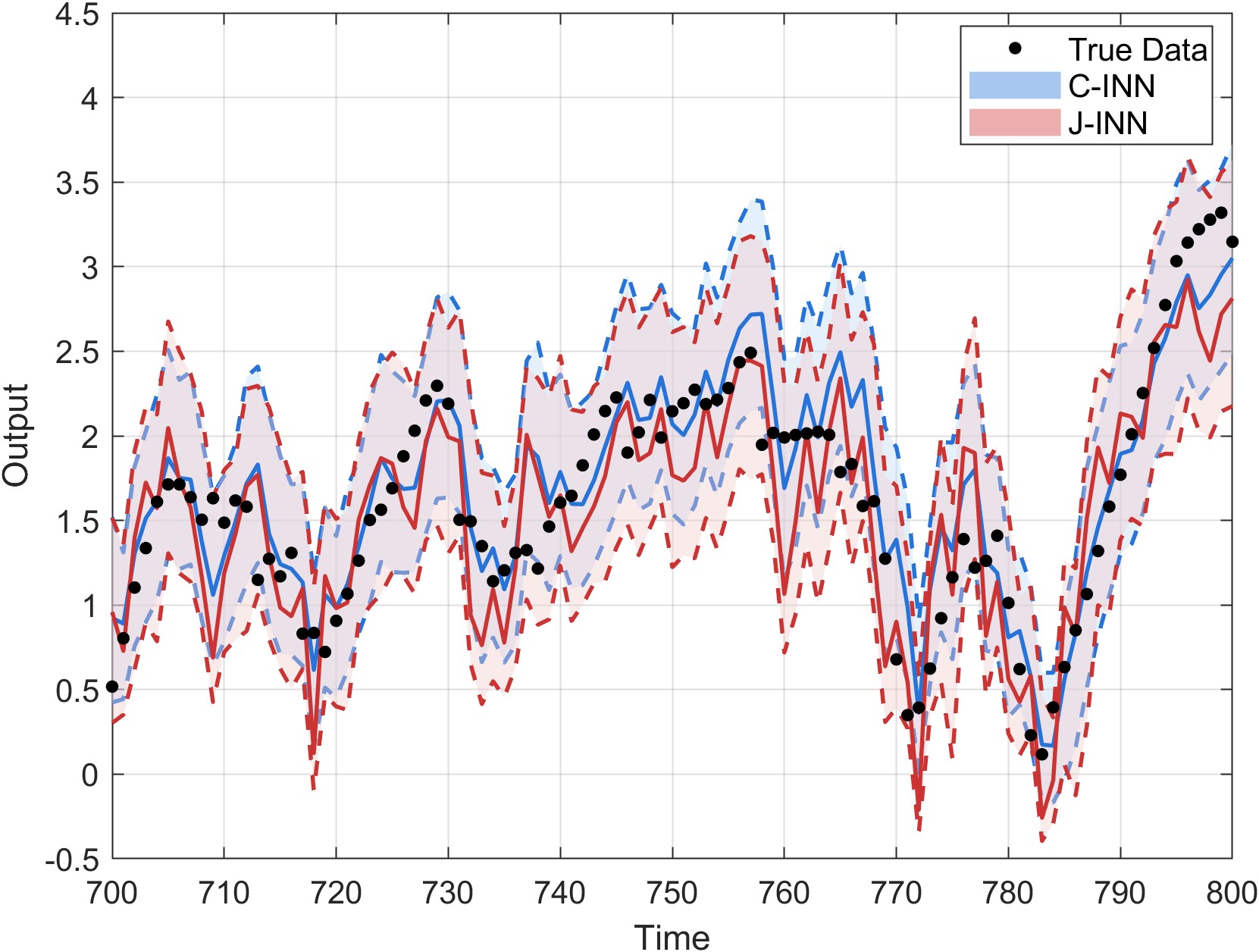}
  \label{fig:snap_exchanger}
}
\hfill
\subfloat[MR-Damper]{%
  \includegraphics[width=0.48\textwidth]{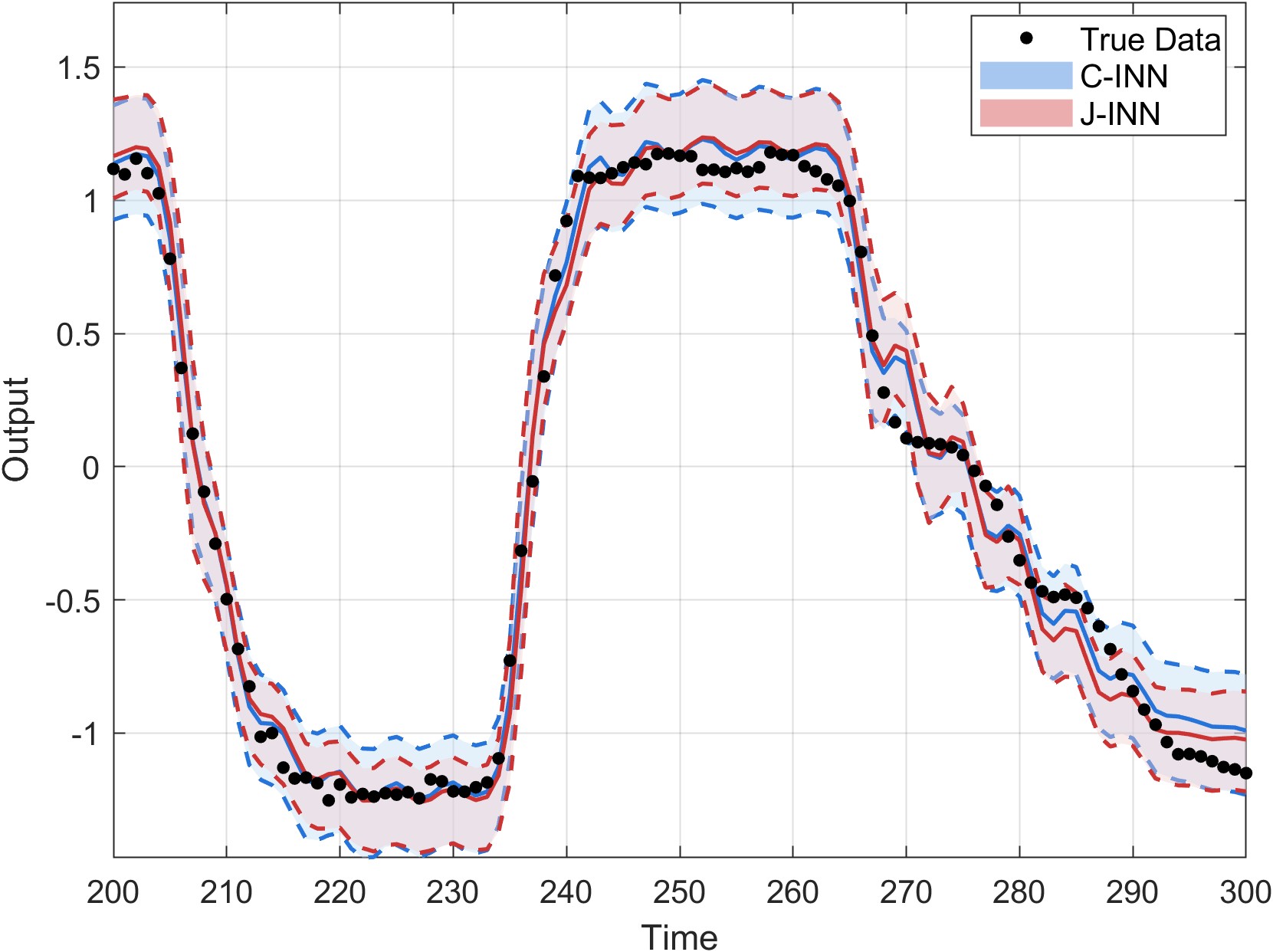}
  \label{fig:snap_damper}
}

\vspace{0.3cm}

\subfloat[Hair Dryer]{%
  \includegraphics[width=0.48\textwidth]{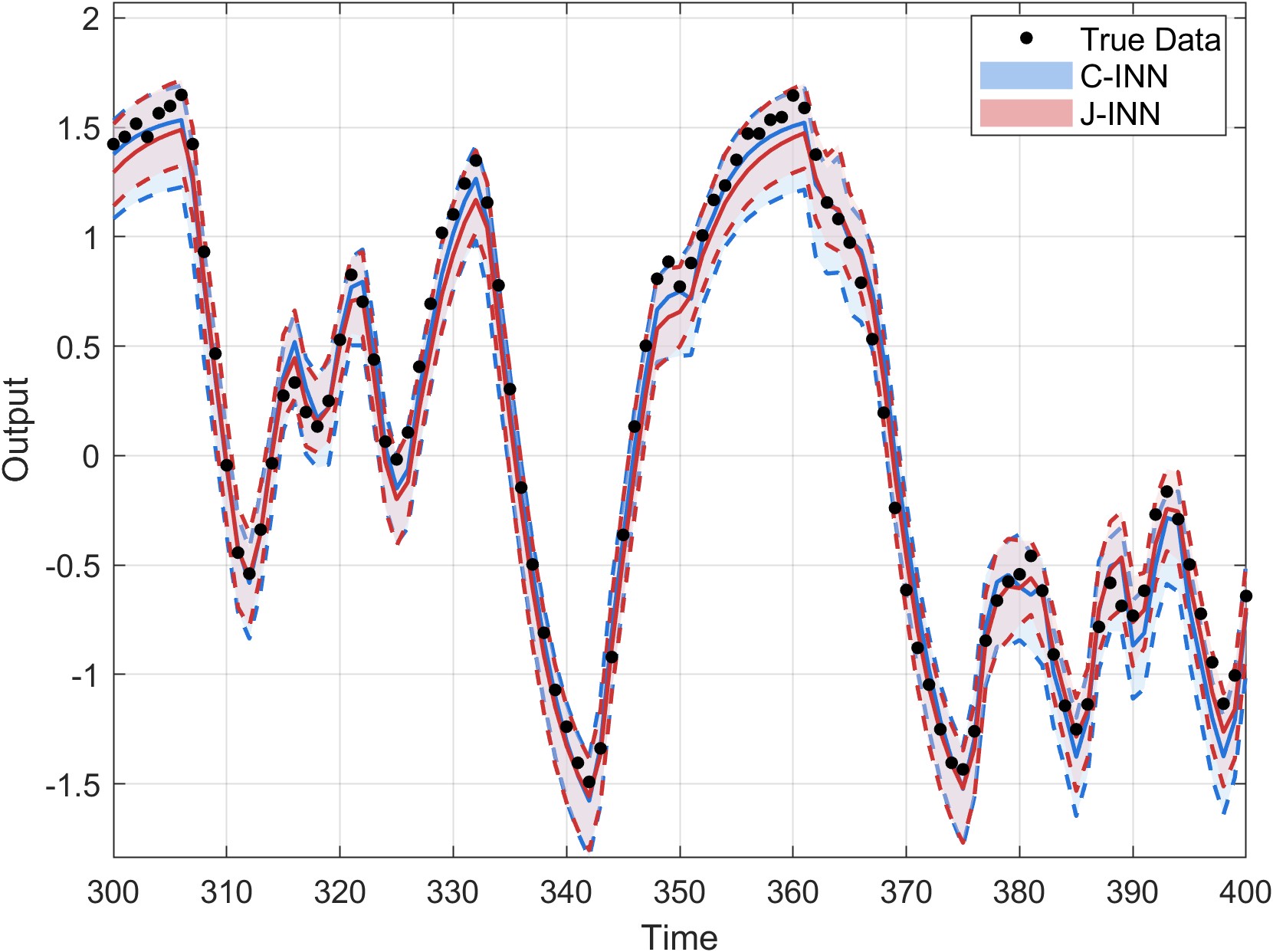}
  \label{fig:snap_dryer}
}
\hfill
\subfloat[Robot Arm]{%
  \includegraphics[width=0.48\textwidth]{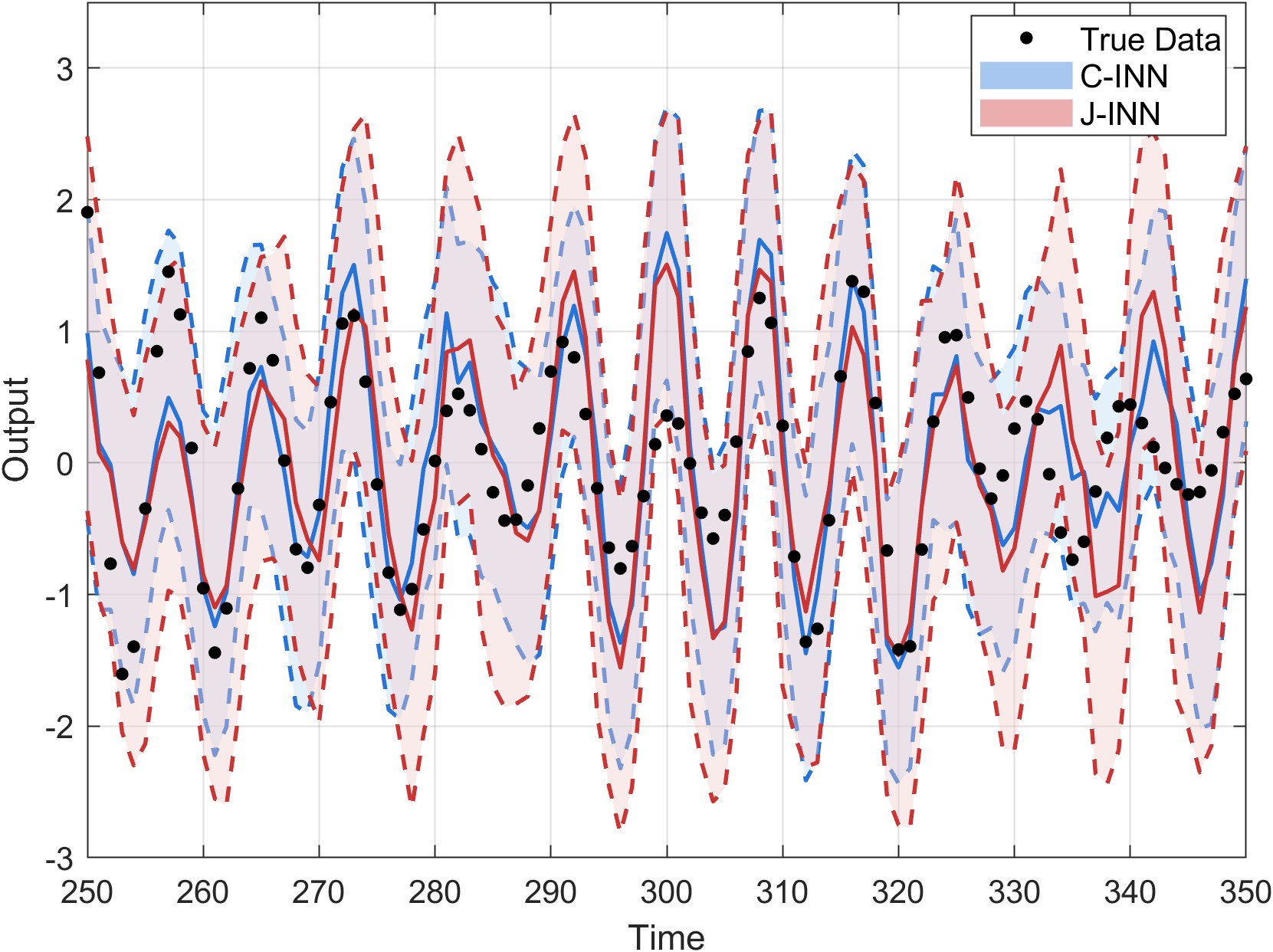}
  \label{fig:snap_robot}
}

\caption{Snapshot of predicted system responses from Interval NODE (INODE) models trained using the proposed Cascade (C-INN) and Joint (J-INN) frameworks on four benchmark SysID datasets. Solid lines (—) indicate point predictions, while dashed lines (– –) represent the 90\% prediction intervals, highlighting uncertainty-aware predictive performance.}
\label{fig:snap}
\end{figure}

\begin{enumerate}[label=(\roman*)]
    \item \textit{Construction of INNs:}
    We propose to systematically build INNs by transforming crisp NNs, replacing their crisp LPs with interval LPs, and avoiding probabilistic assumptions. By applying interval arithmetic across all layers, INNs can effectively propagate both input and parameter uncertainties. We provide the mathematical foundations for extending LSTM and NODE architectures into Interval LSTM (ILSTM) and Interval NODE (INODE). To generate PIs around the network outputs, ILSTM and INODE incorporate the outputs and hidden states of their crisp counterparts into the interval inputs, allowing uncertainty to be effectively propagated through time. Together, these extensions establish a framework for constructing INNs capable of uncertainty-aware modeling of nonlinear systems.
    \item \textit{Training of INNs:} We propose two complementary training strategies to effectively learn INNs while balancing point prediction accuracy (i.e., capturing system dynamics) and uncertainty quantification (i.e., generating PIs). 
    \begin{itemize}
        \item The Cascade INN (C-INN) strategy revisits and extends the two-stage approach introduced in \cite{ferah2025introducing}, where a crisp NN is first trained to achieve high predictive accuracy and then converted into an INN by freezing its structure and updating only the LPs responsible for uncertainty propagation. 
        \item The Joint INN (J-INN) strategy is a unified one-stage framework in which the INN is trained directly from scratch, jointly optimizing for both point prediction accuracy and interval precision. Training is formulated as a multi-objective optimization problem, leveraging Pareto analysis to balance the competing objectives of prediction accuracy and UQ performance. 
    \end{itemize}
    To ensure stable and effective learning, both strategies incorporate UQ-specific loss functions and parameterization techniques designed to address challenges associated with interval LPs.
    \item \textit{Evaluation of Uncertainty-Aware Learning Frameworks for SysID:} We conduct comprehensive SysID experiments to evaluate the proposed ILSTM and INODE models under both training frameworks and benchmark their performance against well-established NN models with UQ capability \cite{mahajan2024bayesian, akins2025monte, gal2016dropout, lakshminarayanan2017simple}. The results demonstrate that both learning strategies effectively develop uncertainty-aware SysID models, capturing system dynamics while generating reliable prediction intervals. Comparative analysis indicates that the Cascade INN strategy excels in point prediction accuracy, reflecting its advantage in leveraging a pre-trained crisp NN. In contrast, the J-INN strategy provides improved UQ, producing tighter and better-calibrated PIs. These findings highlight the trade-offs between prediction accuracy and interval reliability, guiding the selection of an appropriate training strategy based on application requirements.
    \item \textit{Understanding How INNs Represent Uncertainty:} To interpret the performance of the trained models, we analyze how uncertainty is distributed across the INNs. In this context, we employ the notion of \textit{elasticity}, which quantifies the level of uncertainty associated with each INN parameter \cite{ferah2025introducing}, and extend it to channel-wise elasticity to examine how uncertainty in individual input features propagates through the network. While standard elasticity can be challenging to interpret for complex neural networks, channel-wise elasticity provides a more granular, explainable AI perspective, revealing both model uncertainty and the contribution of specific input features. Comparative analysis shows that the C-INN strategy tends to localize uncertainty to specific weights or biases, particularly in the first and final layers, whereas the J-INN strategy promotes a more distributed representation of uncertainty, reflecting fundamental differences in how these training strategies propagate and encode uncertainty throughout the network.

\end{enumerate}

The remainder of this paper is organized as follows. Section~\ref{interval_arithmetic} reviews the preliminaries of interval arithmetic. Section~\ref{systemident} outlines the structure and inference of crisp NNs for SysID. Section~\ref{inn_section} introduces the design and inference of INNs for SysID. Section~\ref{sec5} describes the proposed C-INN and J-INN training strategies. Section~\ref{sec6} presents a comparative evaluation based on prediction accuracy and UQ metrics, complemented by an \textit{elasticity}-based analysis for uncertainty interpretation.

\section{Preliminaries on Interval Arithmetic}\label{interval_arithmetic}

\begin{definition}\label{def1}
For $x \in \mathbb{R}$, a \textit{crisp number} $a$ is a single, deterministic, and finite value such that:
\[
x = a, \quad a \in \mathbb{R}
\]
\end{definition}

\begin{definition}\label{def2}
Let $\underline{a}, \overline{a} \in \mathbb{R}$ and $\underline{a} \leq \overline{a}$.  
An \textit{interval number} $\tilde{a} = [\underline{a}, \overline{a}]$ is defined as the closed subset of crisp numbers, defined in Definition \ref{def1}, given by \cite{interval_artihmetic}:
\[
\tilde{a} = [\underline{a}, \overline{a}] = \{ x \in \mathbb{R} \mid \underline{a} \leq x \leq \overline{a} \}.
\]
\noindent\textit{Special Case:}  
When $\underline{a} = \overline{a}$, the interval $[\underline{a}, \overline{a}]$ collapses to a single point $a$, representing a crisp number \cite{interval_artihmetic}.
\end{definition}

Given interval numbers \([\underline{a}, \overline{a}] \) and \( [\underline{b}, \overline{b}] \), the following operations are defined for interval arithmetic \cite{interval_artihmetic}:

\begin{itemize}
    \item Addition:  
$[\underline{a}, \overline{a}] + [\underline{b}, \overline{b}] = [\underline{a} + \underline{b}, \overline{a} + \overline{b}]$ 
    \item Subtraction:  
$[\underline{a}, \overline{a}] - [\underline{b}, \overline{b}] = [\underline{a} - \overline{b}, \overline{a} - \underline{b}]$ 
\item Multiplication:  
$[\underline{a}, \overline{a}] [\underline{b}, \overline{b}] = [\min(S), \max(S)],$ \label{intmul}  
\text{where } \noindent$S = \{\underline{a} \cdot \underline{b}, \underline{a} \cdot \overline{b}, \overline{a} \cdot \underline{b}, \overline{a} \cdot \overline{b}\}$.
\end{itemize}

We define an interval matrix with \(\underline{A}, \overline{A} \in \mathbb{R}^{m \times n}\), as follows:

\begin{equation}\label{intMatrix}
[\underline{A}, \overline{A}] =
\left\{ X \in \mathbb{R}^{m \times n} \;\middle|\;
\underline{A}_{ij} \leq X_{ij} \leq \overline{A}_{ij}, \;
\forall i = 1,\dots,m,\; j = 1,\dots,n
\right\}.
\end{equation}
Basic matrix operations, such as addition, subtraction, and multiplication, can be easily extended to interval matrices.
\begin{itemize}
    \item Addition and Subtraction:  Let \([\underline{A}, \overline{A}]\) and \([\underline{B}, \overline{B}]\) be as defined in \eqref{intMatrix}. The interval operations are defined as: 
\begin{equation} \label{matrixsum}
[\underline{A}, \overline{A}] + [\underline{B}, \overline{B}] = [\underline{A} + \underline{B}, \overline{A} + \overline{B}]
\end{equation}
\begin{equation} \label{matrixsubs}
[\underline{A}, \overline{A}] - [\underline{B}, \overline{B}] = [\underline{A} - \overline{B}, \overline{A} - \underline{B}]
\end{equation}
\item Multiplication: Let \([ \underline{A}, \overline{A} ]\) and \([ \underline{B}, \overline{B} ]\) as in \eqref{intMatrix} but with \(\underline{A}, \overline{A} \in \mathbb{R}^{m \times p}\) and \(\underline{B}, \overline{B} \in \mathbb{R}^{p \times n}\). It is defined as:
\begin{equation} \label{intmatrixMultiplication}
[ \underline{C}, \overline{C} ] = [ \underline{A}, \overline{A} ] [ \underline{B}, \overline{B} ]
\end{equation}
To compute each entry $[\underline{C}_{ij}, \overline{C}_{ij}]$, we take the interval dot product of the $i$-th row of $[\underline{A}, \overline{A}]$ and the $j$-th column of $[\underline{B}, \overline{B}]$. Each element-wise product yields four boundary combinations, which are captured by four intermediate vectors $\mathbf{v}^{(1)}_{ij}, \mathbf{v}^{(2)}_{ij}, \mathbf{v}^{(3)}_{ij}, \mathbf{v}^{(4)}_{ij} \in \mathbb{R}^p$.
\begin{equation}
\begin{aligned}
\mathbf{v}^{(1)}_{ij} &= [ \underline{a}_{i1}\underline{b}_{1j}, \; \underline{a}_{i2}\underline{b}_{2j}, \; \dots, \; \underline{a}_{ip}\underline{b}_{pj} ] \\
\mathbf{v}^{(2)}_{ij} &= [ \underline{a}_{i1}\overline{b}_{1j}, \; \underline{a}_{i2}\overline{b}_{2j}, \; \dots, \; \underline{a}_{ip}\overline{b}_{pj} ] \\
\mathbf{v}^{(3)}_{ij} &= [ \overline{a}_{i1}\underline{b}_{1j}, \; \overline{a}_{i2}\underline{b}_{2j}, \; \dots, \; \overline{a}_{ip}\underline{b}_{pj} ] \\
\mathbf{v}^{(4)}_{ij} &= [ \overline{a}_{i1}\overline{b}_{1j}, \; \overline{a}_{i2}\overline{b}_{2j}, \; \dots, \; \overline{a}_{ip}\overline{b}_{pj} ]
\end{aligned}
\end{equation}
The lower and upper bounds for the intermediate vectors are then obtained by taking the element-wise minimum and maximum across these four vectors:
\begin{equation}
\begin{aligned}
\mathbf{\underline{v}}_{ij} &= \min \left( \mathbf{v}^{(1)}_{ij}, \mathbf{v}^{(2)}_{ij}, \mathbf{v}^{(3)}_{ij}, \mathbf{v}^{(4)}_{ij} \right) \\
\mathbf{\overline{v}}_{ij}  &= \max \left( \mathbf{v}^{(1)}_{ij}, \mathbf{v}^{(2)}_{ij}, \mathbf{v}^{(3)}_{ij}, \mathbf{v}^{(4)}_{ij} \right)
\end{aligned}
\end{equation}
Finally, mirroring the summation step of classical matrix multiplication, the bounds for the $ij$-th entry of the resulting interval matrix are computed by summing the elements of these extreme vectors:
\begin{equation} \label{eq:lower_upper_bounds}
[\underline{C}_{ij}, \overline{C}_{ij}] = \left[ \sum_{k=1}^p \left(\mathbf{\underline{v}}_{ij}\right)_k, \; \sum_{k=1}^p \left(\mathbf{\overline{v}}_{ij}\right)_k \right].
\end{equation}

\end{itemize}

\section{Background on Neural Networks for System Identification}\label{systemident}

Here, we briefly outline how the NNs are defined and learned for SysID problems. We assume having a dataset, \( \{(u(k), \hat{y}(k))\}_{k=1}^{K} \), with an input \( u(k) \) and an output $\hat{y}(k)$. To represent the dynamics, we aim to learn an NN as follows: 
\begin{equation}\label{systemid}
{y}(k) = g(x(k), \theta)
\end{equation}
with
\begin{equation}\label{NNinput}
x(k) = \begin{aligned}
& \big[ u(k-n_d), \, \dots, \, u(k-n_d-n_x), \\
& \quad y(k-1), \, \dots, \, y(k-n_y) \big]
\end{aligned}
\end{equation}
where \(y(k)\) is the predicted output, \( n_x \) is input lag , \( n_y \) is output lag and  \( n_d \) is  dead time. During training and inference, the model operates in simulation mode, with \( x(k) \) updated after each prediction via the predictions.

This paper uses LSTM and NODE networks to represent \(g(.)\) in \eqref{systemid}. All networks are implemented using feedforward NNs as layers. The remainder of this section provides a brief overview of the feedforward NNs, LSTMs, and NODEs.

\subsection{Feedforward Neural Networks}\label{NNsec}
An \(n\)-layer NN is a function with an input \(x\) as follows:
\begin{equation}\label{NN}
y = g(x; \theta)
\end{equation}
with crisp LPs $\theta = \{\theta_i\}_{i=1}^{n}$. Here, \(g\) is defined as:
\begin{equation}\label{NNlayer}
g(x; \theta) = g_n(  g_{n-1}(g_1(x; \theta_1); \dots; \theta_{n-1}); \theta_n)
\end{equation}
The inference of $g_i$ is defined for \(x_i \in \mathbb{R}^{m}\) as follows:  
\begin{equation}\label{NNforward}
g_i(x_i; \theta_i) = \sigma_i(W_i x_i + b_i)
\end{equation}
with \(\sigma = \{ \sigma_i \}_{i=1}^{n}\) represents the activation functions and \(\theta_i = (W_i, b_i)\) consists \(W_i \in \mathbb{R}^{p \times m}\) and \(b_i \in \mathbb{R}^p\).

\subsection{Long Short Term Memory Network}\label{LSTM}
For a sampled input $x(k)$, an \(n\)-layer LSTM is defined : 
\begin{equation}\label{lstmstacked}
(h(k), c(k)) = g(x(k), h(k-1), c(k-1); \theta)
\end{equation}
where each layer is characterized by the input gate \(i(k)\), forget gate \(f(k)\), output gate \(o(k)\), hidden state \(h(k)\), and cell state \(c(k)\) at time step \(k\), which are defined as \cite{tuna2022deep}: 
\begin{equation}\label{lstm}
\begin{aligned} 
i_i(k) &= \sigma^{\text{sig}}(W_i^\textit{i} x_i(k) + U_i^\textit{i} h_i(k-1) + b_i^\textit{i}) \\
f_i(k) &= \sigma^{\text{sig}}(W_i^f x_i(k) + U_i^f h_i(k-1) + b_i^f) \\
o_i(k) &= \sigma^{\text{sig}}(W_i^o x_i(k) + U_i^o h_i(k-1) + b_i^o) \\
q_i(k) &= \sigma^{\text{tanh}}(W_i^c x_i(k) + U_i^c h_i(k-1) + b_i^c) \\
c_i(k) &= f_i(k) \odot c_i(k-1) + i_i(k) \odot q_i(k) \\
h_i(k) &= o_i(k) \odot \sigma^{\text{tanh}}(c_i(k))
\end{aligned}
\end{equation}
with 
$\theta_{i} = (W_{i}^{f}, U_{i}^{f}, b_{i}^{f}, W_{i}^{\textit{i}}, U_{i}^{\textit{i}}, b_{i}^{\textit{i}}, W_{i}^{c}, U_{i}^{c}, b_{i}^{c}, W_{i}^{o}, U_{i}^{o}, b_{i}^{o})$, where \(i\), \(f\), \(o\) and \(c\) indicate gates. In \eqref{lstm},  \(\sigma^{\text{sig}}\) \ and \(\sigma^{tanh}\) are the sigmoid and tanh activation functions, respectively. For point prediction, the final layer is defined as:
\begin{equation}\label{LSTM_N}
y(k) = W_n h(k) + b_n
\end{equation}

\subsection{Neural Ordinary Differential Equations Networks}\label{NODE}
Given an input \(({x}(t), t)\), state \(h(t)\) at time \(t\), the \(n\)-layer NODE is defined as \cite{node}:
\begin{equation}\label{nodecont}
\frac{d{h}(t)}{dt} = g({x}(t), t; \theta)
\end{equation}
To obtain \({h}(T)\) at time \(T\), we solve the following integral:
\begin{equation}\label{nodesolve}
{h}(T) = {h}(t_0) + \int_{t_0}^{T} g({x}(t), t; \theta) \, dt
\end{equation}

In this study, we apply Euler's method and rewrite \eqref{nodesolve} as \cite{zhang2019anodev2}:
\begin{equation}\label{nodediscrete}
{h}(k) = {h}(k-1) + g({x}(k); \theta)
\end{equation}
which is defined with $g_i$ \((i = 1, \dots, n-1)\) as:
\begin{equation}\label{nodelayer}
g_i(x_i(k); \theta_i) = \sigma_i(W_i x_i(k) + b_i)
\end{equation}
and the final layer \(i = n\) with :
\begin{equation}\label{nodelastlayer}
y(k)  = W_n x_n(k) + b_n
\end{equation}

\section{Structure of Interval Neural Networks for UQ in SysID}\label{inn_section}
This section presents the structure and inferences of INNs for SysID with the capability of generating PIs. We first extend the feedforward NN into its INN counterpart, as it is used to define layers of ILSTM and INODE in detail. Then, we define the equations of ILSTM and INODE for PI generation. We base our INN construction approach on an NN with  LP \(\theta\).

\subsection{Feedforward Interval Neural Network Construction}\label{INN}

The deployed INN extends the crisp NN by enabling it to generate interval-valued predictions that can be used to quantify uncertainty at each output. Formally, the INN maps an input $x$ to an interval-valued output \( \tilde{y} = [\underline{y}, \overline{y}] \) through a composition of interval-valued layers:
\begin{equation}\label{intervalnn}
\tilde{y} = \tilde{g}(x; \tilde{\theta}) = \tilde{g}_n\left( \tilde{g}_{n-1}\left( \dots \tilde{g}_1(x; \tilde{\theta}_1); \tilde{\theta}_{n-1} \right); \tilde{\theta}_n \right),
\end{equation}
where each layer’s function \( \tilde{g}_i(\cdot) \) is defined using interval arithmetic as:
\begin{equation}\label{intervalforward}
\tilde{g}_i(\tilde{x}_i; \tilde{\theta}_i) = \sigma_i\left( \tilde{W}_i \tilde{x}_i + \tilde{b}_i \right),
\end{equation}
with interval-valued weights \( \tilde{W}_i = [\underline{W}_i, \overline{W}_i] \) and biases \( \tilde{b}_i = [\underline{b}_i, \overline{b}_i] \). The nonlinear activation \( \sigma_i(\cdot) \) is applied component-wise using interval arithmetic operations as presented in Section \ref{interval_arithmetic}. Note that, to handle a crisp input at the first layer, we apply the special condition described in Definition~\ref{def2}, interpreting the input as a degenerate interval: \( \tilde{x}_1 = [x, x] \).

The interval output \( \tilde{y} = [\underline{y}, \overline{y}] \) must be constructed to enclose the crisp output \( y \) produced by the crisp NN $g(.)$ satisfying the condition $y \in \tilde{y}$. This formulation allows the model to retain the point prediction capability of the underlying crisp NN while simultaneously providing an interval output suitable for UQ. To accomplish such a goal, the INN introduces uncertainty directly into the LPs of a crisp NN by replacing each scalar parameter \( \theta_i \) with an interval-valued counterpart \( \tilde{\theta}_i \), the interval LPs, defined as:
\begin{equation}\label{interval_theta}
\tilde{\theta}_i = [\underline{\theta}_i, \overline{\theta}_i] = [\theta_i - \underline{\Delta}_i, \theta_i + \overline{\Delta}_i],
\end{equation}
where \( \underline{\Delta}_i \) and \( \overline{\Delta}_i \) are the uncertainty LPs of the INN forming the interval margin \( \tilde{\Delta}_i = [\underline{\Delta}_i, \overline{\Delta}_i] \). The full parameter set of the INN is thus \( \{ \theta_i, \tilde{\Delta}_i \}_{i=1}^n \). To satisfy the basic definition of an interval (Definition~\ref{def2}), it must hold that \( \underline{\theta}_i < \overline{\theta}_i \).

\subsection{Inference of Interval LSTM}\label{ILSTM}
This section introduces ILSTM based on the INN design. We extend \eqref{lstm} and \eqref{LSTM_N} by defining \(\tilde{\theta}\) as in \eqref{interval_theta} and obtain:

\begin{equation}\label{ilstmi}
\begin{aligned}
\tilde{i}_i(k) &= \sigma^{sig}(\tilde{W}_i^\textit{i} \tilde{x}_i(k) + \tilde{U}_i^\textit{i} \tilde{h}_i(k-1) + \tilde{b}_i^\textit{i}) \\
\tilde{f}_i(k) &= \sigma^{sig}(\tilde{W}_i^f \tilde{x}_i(k) + \tilde{U}_i^f \tilde{h}_i(k-1) + \tilde{b}_i^f) \\
\tilde{o}_i(k) &= \sigma^{sig}(\tilde{W}_i^o \tilde{x}_i(k) + \tilde{U}_i^o \tilde{h}_i(k-1) + \tilde{b}_i^o) \\
\tilde{q}_i(k) &= \sigma^{tanh}(\tilde{W}_i^c \tilde{x}_i(k) + \tilde{U}_i^c \tilde{h}_i(k-1) + \tilde{b}_i^c) \\
\tilde{c}_i(k) &= \tilde{f}_i(k) \odot \tilde{c}_i(k-1) + \tilde{i}_i(k) \odot \tilde{q}_i(k) \\
\tilde{h}_i(k) &= \tilde{o}_i(k) \odot \sigma^{tanh}(\tilde{c}_i(k))
\end{aligned}
\end{equation}
Then, we can define the overall ILSTM network as:
\begin{equation}\label{ILSTM_N-1}
(\tilde{h}(k), \tilde{c}(k)) =  
 \tilde{g}(\tilde{x}(k), \tilde{h}(k-1),
 \tilde{c}(k-1); \tilde{\theta})
\end{equation}
with
\begin{equation}\label{ILSTM_N}
\tilde{y}(k) = \tilde{W}_n \tilde{h}(k) + \tilde{b}_n.
\end{equation}

As the aim is to generate a PI around the output of the LSTM (i.e., $y \in \tilde{y}$), we define the input \(\tilde{x}(k)\) in the ILSTM by updating its output and states by using the crisp LSTM's output and state as follows:

\begin{equation}\label{centerh}
[\underline{h}(k-1), \overline{h}(k-1)] = [h(k-1), h(k-1)]
\end{equation}
\begin{equation}\label{centerc}
[\underline{c}(k-1), \overline{c}(k-1)] = [c(k-1), c(k-1)]
\end{equation}
\begin{equation}\label{centerylstm}
[\underline{y}(k-1), \overline{y}(k-1)] = [y(k-1), y(k-1)]
\end{equation}

\subsection{Inference of Interval NODE}\label{INODE}
We construct INODE similar to ILSTM by $\tilde{\theta}$ using \eqref{interval_theta}. The equations of INODE are expressed as follows:
\begin{equation}\label{intervalnode}
\tilde{h}(k+1) = \tilde{h}(k)+ \tilde{g}(\tilde{x}(k); \tilde{\theta})
\end{equation}
For layers \(i = 1, \dots, n-1\), we define:
\begin{equation}\label{intervalnodelayer}
\tilde{h}_i(k+1) = \tilde{h}_i(k) + \sigma_i(\tilde{W}_i\tilde{x}_i(k)  + \tilde{b}_i)
\end{equation}
while for the final layer \(i = n\), we define:
\begin{equation}\label{intervalnodelast}
\tilde{h}_n(k+1) = \tilde{h}_n(k) + (\tilde{W}_n \tilde{x}_n(k) + \tilde{b}_n)
\end{equation}

As we have done in ILSTM, to generate a PI around the output of the NODE (i.e. $y \in \tilde{y}$), we update at each step INODE output with:
 \begin{equation}\label{nodecenter}
[\underline{y}(k-1),\overline{y}(k-1)] = [y(k-1),y(k-1)]
\end{equation}

\section{Learning Interval Neural Networks for UQ in SysID}\label{sec5}

In SysID, incorporating UQ is crucial for providing reliable and informative models that capture both predictions and their confidence bounds. A key method to achieve this is through constructing PIs, which estimate the range
\begin{equation}
    {\tilde{y}}(k) = \left[\underline{y}(k), \bar{y}(k)\right]
\end{equation}
where an target output \(\hat{y}(k)\) is expected to fall with target coverage level \(\alpha\): 
\begin{equation}
  \text{P}\{ \hat{y}(k) \in \tilde{y}(k) \} \geq \alpha.
\end{equation}
While valid PIs must satisfy this coverage condition, it is equally important for PIs to be narrow, providing high-quality uncertainty estimates. The width
\begin{equation}
    \Delta \tilde{y}(k) = \bar{y}(k) - \underline{y}(k)
\end{equation}
directly affects the informativeness of the PI. In addition to UQ, achieving high point-wise prediction accuracy remains a primary goal to ensure the model’s practical utility.

In this section, before detailing each training strategy, we first describe the data preparation, loss functions, parameterization techniques, and initialization methods common to both frameworks. Then, we present two different  DL frameworks to train the presented ILSTM and INODE networks to develop models that generate high-quality PIs—intervals that are both reliable (valid coverage) and precise (minimal width)—while maintaining strong point-wise accuracy. 
\begin{itemize}
    \item \textbf{Cascade INN Learning Strategy (C-INN):} A two-stage learning strategy where a crisp NN is first trained for point predictions, followed by a second stage that learns uncertainty LPs to construct the INN while freezing the original parameters.
    \item \textbf{Joint INN Learning Strategy (J-INN):} A one-stage learning strategy that jointly optimizes both point prediction and UQ losses in an end-to-end fashion, using techniques such as GradNorm \cite{chen2018gradnorm} for balancing.
\end{itemize}
Both approaches aim to learn a model \(\tilde{g}(\cdot)\) that outputs PIs \(\tilde{y}(k)\) satisfying the marginal target coverage condition while keeping the interval width minimal and ensuring accurate point estimates \footnote{MATLAB implementation. [Online]. Available: GitHub repo will be shared upon acceptance of the paper.}

\subsection{Data Preparation} 

To train the models within a DL framework, we first extract \( B  \) trajectories with window length \( N \) to define the mini-batches via Algorithm-\ref{alg1}.Based on the sequence \( \{u(k), \hat{y}(k)\}_{k=1}^{K} \), the input matrix \( U \) and the target output matrix \( \hat{Y} \) are constructed with appropriate dimensions for mini batch training. The models are subsequently trained using a DL optimizer over mini-batches of size \( mbs \). During training, we assume that the input sequence \( u(k) \) is known, and the output at the first time step \( \hat{y}(1) \) is given. Since the regression input \( x(k) \), as defined in ~\eqref{NNinput}, is constructed using lagged values of both \( u(k) \) and \( y(k) \), zero-padding is applied to the unavailable past values at the beginning of the sequence to ensure consistent input dimensionality.

\begin{algorithm}[t]
\caption{Extracting Training Trajectories: Windowing Data}
\label{alg1}
\begin{algorithmic}[1]
\STATE \textbf{Input:}  \( \{u(k), \hat{y}(k)\}_{k=1}^{K} \), \(N\),\(n_{step}\)
\STATE \( i \gets 1 \)
\FOR{\( k = 1 \) \textbf{to} \( K - N \) \textbf{step} \( n_{step} \)}

    \STATE \( \{U(i,:), \hat{Y}(i,:)\} = \{u(k:k+N-1), \hat{y}(k:k+N-1)\} \)
    \STATE \( i \gets i + 1 \)
\ENDFOR
\STATE \textbf{Output:} \( U \)  and \( \hat{Y} \) 
\end{algorithmic}
\end{algorithm}

\subsection{Loss Functions}

The loss functions for prediction accuracy and UQ are briefly defined below.

\subsubsection{Accuracy Loss} To obtain accurate point predictions, \(L_{MSE}\) loss is used, which is also referred to as the regression loss:

\begin{equation}\label{l2oss}
L_{MSE}=\frac{1}{BN} \sum_{m=1}^{B}\sum_{k=1}^{N} ( \hat{Y}(m,k) - Y(m,k) )^2 
\end{equation}

\subsubsection{Prediction Interval Quality Loss} To obtain well-calibrated and informative PIs, $L_{\text{RQR-W}}$ loss is used \cite{pouplin2024relaxed}, which is also referred to as the UQ loss:

\begin{equation}\label{rqrw}
L_{RQR-W}=\frac{1}{BN} \sum_{m=1}^{B}\sum_{k=1}^{N} \mathcal{L}_{RQR-W}
\end{equation}
where
\begin{equation}\label{rqr_loss_sum}
\mathcal{L}_{\text{RQR-W}} = \mathcal{L}_{\text{RQR}} +\lambda \mathcal{L}_{\text{W}}
\end{equation}
Here, $\mathcal{L}_{\text{RQR}}$ is the loss function related to target coverage \(\alpha\) and is defined as:
\begin{equation}\label{rqr_loss}
\mathcal{L}_{\text{RQR}} = 
\begin{cases} 
\alpha \kappa & \text{if } \kappa \geq 0 \\
(\alpha - 1)\kappa & \text{if } \kappa < 0
\end{cases} \\
\end{equation}
where 
\begin{equation}
    \kappa = (\hat{Y}(m,k) - \underline{Y}(m,k))(\hat{Y}(m,k) - \overline{Y}(m,k)).
\end{equation}
The \(\mathcal{L}_{\text{W}}\) term penalizes the PI width to enforce narrower bounds, weighted by the hyperparameter \(\lambda\). $\mathcal{L}_{\text{W}}$ is defined as:
\begin{equation}\label{w_loss}
\mathcal{L}_{\text{W}} = {(\overline{Y}(m,k) - \underline{Y}(m,k))^2}/{2}
\end{equation}

\subsubsection{Parametrization Tricks and Initialization} 
During the training of the INN, we must ensure that all LPs \(\tilde{\Delta}\) result in \(\tilde{\theta}\) that satisfy the conditions of interval numbers (i.e., Definition-\ref{def2}). Thus, we must guarantee that \(\Delta_i \geq 0\). Yet, enforcing this constraint transforms \eqref{rqr_loss_sum} into a constraint optimization problem. Thus, as all the built-in DL optimizers are unconstrained ones, we perform the following parameterization tricks to eliminate the constraints:
\begin{equation}\label{trick}
\underline{\Delta}_i= \sigma_{\Delta}(\underline{\Delta}'_i), \quad
\overline{\Delta}_i= \sigma_{\Delta}(\overline{\Delta}'_i)
\end{equation}
where $\underline{\Delta}'_i$ and $\overline{\Delta}'_i$ are the new LPs, and \(\sigma_{\Delta}(.)\) is a function that ensures outputs remain strictly positive. In this paper, we propose two functions for $\sigma_{\Delta}$, which are absolute function $\sigma^{abs}$ and the ReLU function $\sigma^{ReLU}$.

The initialization of \(\tilde{\Delta}\) is important as it might increase the training time. Given the parameters \(\theta\), initialized appropriately using MATLAB’s default initialization methods for both linear and LSTM layers~\cite{MATLAB}, the initialization of \(\tilde{\Delta}\) is performed for the hidden layers as follows: 
 \begin{equation}\label{rhidden}
        \{\tilde{\Delta}_1,...,\tilde{\Delta}_{n-1}\} = \{[{\theta_1}, {\theta_1}], \dots, [\theta_{n-1}, {\theta_{n-1}}]\} r_{\text{h}} 
 \end{equation}
where  $r_{\text{h}} \in [0,1]$ is the hidden layer uncertainty rate. The output layer is initialized with: 
\begin{equation}\label{rout}
\tilde{\Delta}_n\ = [{\theta_n},{\theta_n}] r_{\text{o}} 
  \end{equation}
where $r_{\text{o}} \in [0,1]$ is the 
is the output layer uncertainty rate. $r_{\text{h}}$ and $r_{\text{o}}$ are hyperparameters to be determined.

\subsection{Cascade INN Learning Strategy: A two-stage learning strategy}

In the first approach, we propose a C-INN that systematically transforms a crisp NN into an INN. The training procedure for the two-stage approach is outlined in Algorithms \ref{alg2} and \ref{alg2-helper}. The two main stages are as follows: 
\begin{enumerate}
    \item A crisp NN \(g\) is trained to learn \(\theta\) using only the accuracy loss function in defined in \eqref{l2oss}.  
     The optimization problem to be minimized at every epoch for the NN is defined as follows: 
    \begin{equation}\label{loss}
    \theta^* = \argmin_{\theta} L_{MSE} 
    \end{equation}
    \item  The trained crisp NN is extended to an INN by introducing uncertainty LPs \(\tilde{\Delta}\).  The uncertainty LPs are trained while freezing the previously learned \(\theta^*\) using only the UQ loss function defined in \eqref{rqrw}. Thus, the following optimization problem is solved at each epoch:
\begin{equation}\label{rqrw_loss} 
\tilde{\Delta}^* =\argmin_{\tilde{\Delta}} L_{\text{RQR-W}}
\end{equation}  
\end{enumerate}

\begin{algorithm}[t]
\small  
\caption{Cascade INN Learning Strategy}
\label{alg2}
\begin{algorithmic}[1]

\STATE \textbf{Input:} Training data: \( \{u(k), \hat{y}(k)\}_{k=1}^{K} \)
\STATE Preprocessed data : \( U,\hat{Y} \) \hfill $\triangleright$ \(\textnormal{Algorithm-} \ref{alg1}\) 
\STATE Mini-batch size: \(mbs\) 
\STATE Number of trajectory: \(B\) 
\STATE Number of epochs: \(E\)
\STATE Coverage: $\alpha$, penalty: $\lambda$
   \STATE \textbf{Output:} LP sets \(\theta\) and \(  \tilde{\Delta}\) 

   \STATE \textbf{Initialize \(\theta, \tilde{\Delta} \)}\hfill $\triangleright$ 
   \textnormal{Eq}.\eqref{rhidden}-\eqref{rout} 
\STATE \( \theta^* \gets \) \texttt{S1}$(U, \hat{Y}, \theta, mbs, B, E)$\textit{ //Stage 1: Learning NN}  \hfill $\triangleright$ \(\textnormal{Algorithm-} \ref{alg2-helper}\) 

\STATE \( \tilde{\Delta}^* \gets \) \texttt{S2}$(U, \hat{Y}, \theta^*, \tilde{\Delta}, mbs, B, E, \alpha, \lambda)$\textit{ // Stage 2: NN to INN}  \hfill $\triangleright$ \(\textnormal{Algorithm-} \ref{alg2-helper}\)

\RETURN \( \theta = \theta^*, \tilde{\Delta} = \tilde{\Delta}^* \)

\end{algorithmic}
\end{algorithm}

\begin{algorithm}[t]
\small  
\caption{Cascade INN Learning Strategy: The two stages}
\label{alg2-helper}
\begin{algorithmic}[1]
\STATE\textit{----------------------------- Stage-1: Crisp NN Training ----------------------------- }
\STATE \textbf{function} \texttt{S1}($U, \hat{Y},\theta, mbs, B, E$)
    \STATE \textbf{Output:} LP \( \theta \)
\FOR{$e = 1$ to $E$}
    \FOR{each mini-batch in $B$}
        \STATE Select mini-batch $\{U(1:mbs,1:N), \hat{Y}(1:mbs,1:N)\}$
            \STATE Construct $X(1:mbs,1:N)$ \hfill $\triangleright$ Eq.~\eqref{NNinput}
            \STATE Compute $Y \leftarrow g(X(1:mbs,1:N); \theta)$ \hfill $\triangleright$ Sec.~\ref{NNsec}
            \STATE Compute loss: $L_{MSE}$ \hfill $\triangleright$ Eq.~\eqref{rqrw_loss}
            \STATE Compute gradients: $\partial L_{MSE} / \partial \theta$
            \STATE Update $\theta$ via Adam optimizer
    \ENDFOR
\ENDFOR
\STATE \( \theta^* = \arg\min(L_{MSE}) \)
\STATE \textbf{return:} $\theta = \theta^*$
\STATE\textit{----------------------------- Stage-2: INN Training ----------------------------- }
\STATE \textbf{function}  \texttt{S2}($U, \hat{Y}, \theta^*,\tilde{\Delta}, mbs, B, E,\alpha, \lambda$)
\STATE \textbf{Output:} LP set \( \tilde{\Delta} \)
\FOR{$e = 1$ to $E$}
    \FOR{each mini-batch in $B$}
              \STATE Select mini-batch $\{U(1:mbs,1:N), \hat{Y}(1:mbs,1:N)\}$
            \STATE Construct $X(1:mbs,1:N)$ \hfill $\triangleright$ Eq.~\eqref{NNinput}
            \STATE Compute $Y \leftarrow g(X(1:mbs,1:N); \theta^*)$ \hfill $\triangleright$ Sec.~\ref{NNsec}
            \STATE Perform tricks and compute $\underline{\Delta}_i, \overline{\Delta}_i$ \hfill $\triangleright$ Eq.~\eqref{trick}
            \STATE Define \( \tilde{\theta} \) \hfill $\triangleright$ Eq.~\eqref{interval_theta}
            \STATE Compute \( \tilde{Y} \leftarrow \tilde{g}(X(1:mbs,1:N); \tilde{\theta}) \) \hfill $\triangleright$ Sec.~\ref{INN}
            \STATE Compute loss: \( L_{\text{RQR-W}} \) \hfill $\triangleright$ Eq.~\eqref{rqrw_loss}
            \STATE Compute gradients: \( \partial L / \partial \tilde{\Delta} \)
            \STATE Update \( \tilde{\Delta} \) via Adam optimizer
    \ENDFOR
\ENDFOR
\STATE \( \tilde{\Delta}^* = \arg\min(L_{\text{RQR-W}}) \)
\STATE \textbf{return:} $ \tilde{\Delta}=  \tilde{\Delta}^*$

\end{algorithmic}
\end{algorithm}

\subsection{Joint INN Learning Strategy: A one-stage learning strategy}

The J-INN framework aims to leverage the two-stage approach within a unified one-stage framework by seeking a Pareto-optimal solution across the optimization problems defined in \eqref{loss} and \eqref{rqrw_loss}. To accomplish such a goal, we define a Pareto loss function as a weighted combination of $L_{MSE}$ and $L_{RQR-W}$:
\begin{equation}
    L_{pareto} = s_1L_{MSE} + s_2L_{RQR-W}
\end{equation}
where \(s_1\) and \( s_2\) are dynamic loss scales. The shared LPs (i.e., the ones of the crisp NN) \( \theta \) are updated using the gradients of \( L_{\text{pareto}} \), as both loss functions contribute to the underlying representation learning. Meanwhile, the uncertainty LPs \( \tilde{\Delta} \), which directly influence \( L_{\text{RQR-W}} \), are updated based solely on gradients of the UQ loss.

To adaptively balance the influence of each loss (i.e., \( L_{\text{MSE}} \) and \( L_{\text{RQR-W}} \)) during training, we employ the GradNorm strategy~\cite{chen2018gradnorm}, which is implemented by minimizing the discrepancy between the actual and desired gradient norms of each loss for a selected set of shared parameters \( \theta^{sh} \subset \theta \). For both our LSTM and NODE-based architectures, we define \( \theta^{sh} \) as the weights of the penultimate layer \( \theta_{n-1} \), which encode richer temporal features compared to the output layer and provide more informative gradient signals for loss balancing. This design choice ensures stable and meaningful gradient measurements for effective task loss balancing.

\begin{algorithm}[h!]
\small  

\caption{Joint INN Learning Strategy}
\label{alg3}
\begin{algorithmic}[1]
\STATE \textbf{Input:} Training data: \( \{(u(k), \hat{y}(k)\}_{k=1}^{K} \)
\STATE Preprocessed data : \( U,\hat{Y} \) \hfill $\triangleright$ \(\textnormal{Algorithm-} \ref{alg1}\)   
\STATE Mini-batch size: \(mbs\) 
\STATE Number of trajectory: \(B\) 
\STATE Number of epochs: \(E\) 

\STATE Coverage: $\alpha$, penalty: $\lambda$ 
\STATE Scales: \(s=\{{s_1,s_2}\}\)

   \STATE \textbf{Output:} LP sets \(\theta\) and \(  \tilde{\Delta}\) 

\STATE \textbf{Initialize \(\theta, \tilde{\Delta} \)}\hfill $\triangleright$ \textnormal{Eq}.\eqref{rhidden}-\eqref{rout}

\FOR {\(e=1\) to \(E\) }
    \FOR{each \(mbs\) in \(B\) }
        \STATE Select mini-batch $\{U(1:mbs,1:N), \hat{Y}(1:mbs,1:N)\}$
         \STATE Construct X(1:\(mbs\),1:N) \hfill $\triangleright$ \(\textnormal{Eq}.\eqref{NNinput}\)    

        \STATE Compute Y$\leftarrow g($X(1:\(mbs\),1:N)$; \theta)$  \hfill $\triangleright$ Sec.  \ref{NNsec}
        \STATE Perform tricks and compute $\underline{\Delta}_i,\overline{\Delta}_i$ \hfill $\triangleright$ \(\textnormal{Eq}.\eqref{trick}\)

        \STATE Define \(\tilde{\theta}\) 
 \hfill $\triangleright$ \(\textnormal{Eq}.\eqref{interval_theta}\)

        \STATE Compute $\tilde{Y}$  $\leftarrow \tilde{g}($X(1:\(mbs\),1:N)$;\tilde{\theta})$ \hfill $\triangleright$ Sec. \ref{INN}     
\STATE Compute task losses:
\\ \hspace{1em}Regression loss: \(L_{MSE} \) \hfill $\triangleright$ Eq.~\eqref{l2oss}
\\ \hspace{1em}UQ loss: \(L_{\text{RQR-W}} \) \hfill $\triangleright$ Eq.~\eqref{rqrw_loss}
         \STATE Compute  \(L_{pareto} = s_1L_{MSE} + s_2L_{RQR-W}\)        
        \STATE Compute 
         \({\partial {L_{pareto}}}/{\partial {\theta}}, {\partial {L}_{RQR-W}}/{\partial \tilde{\Delta}} \)  
        \STATE \text{Compute  \( {\partial {L_{grad}}}/{\partial s} \leftarrow GradNorm( L_{MSE},L_{RQR-W} , \theta^{sh}, s)\)} 
\STATE Update parameters via Adam optimizer:
  \\ \hspace{1em}Update \( \theta \) using \( \partial L_{\text{pareto}}/\partial \theta \)
 \\ \hspace{1em}Update \( \tilde{\Delta} \) using \( \partial L_{\text{RQR-W}}/\partial \tilde{\Delta} \)
 \\ \hspace{1em}Update \( s \) using \( \partial L_{\text{grad}}/\partial s \)

      \STATE Renormalize \(s\) so that \(s_1+s_2=1\)       
\ENDFOR
\ENDFOR
\STATE \({\theta}^* = \argmin(L_{pareto}) \)
\STATE \(\tilde{\Delta}^* = \argmin({L_{RQR-W}}) \)
\RETURN $\theta=\theta^*, \tilde{\Delta}=\tilde{\Delta}^* $

\end{algorithmic}
\end{algorithm}

The full training procedure incorporating GradNorm for adaptive loss weighting and the joint update of parameters \( \theta, \tilde{\Delta} \), and \( s \) is detailed in Algorithm~\ref{alg3}. Here, the \textit{GradNorm} (step 21 in Algorithm~\ref{alg3}) is defined with the following loss at iteration \(\tau\) is formulated as:
\begin{equation}
L_{\text{grad}}(\tau) = \sum_{j=1}^2 \left| G_j(\tau) - \bar{G}(\tau) \cdot r_j(\tau)^\beta \right|_1,
\end{equation}
where \( G_j(\tau) \) denotes the gradient norm of the scaled loss \( L_j \) with respect to the shared parameters \( \theta^{sh} \) at iteration \( \tau \). 
For each loss function \( L_j \) (i.e., \( L_{\text{MSE}} \) and \( L_{\text{RQR-W}} \)), it is defined as:
\begin{equation}
G_j(\tau) = \left\| \nabla_{\theta^{sh}} \left( s_j(\tau) L_j(\tau) \right) \right\|_2
\end{equation}
The average gradient norm \(\bar{G}(\tau)\) is defined as:
\begin{equation}
\bar{G}(\tau) = \frac{1}{2} \sum_{j=1}^2 G_j(\tau)
\end{equation}
The relative inverse training rate \(r_j(\tau)\) is defined as:
\begin{equation}
r_j(\tau) = \frac{ \frac{L_j(\tau)}{L_j(0)} }{ \frac{1}{2} \sum_{k=1}^2 \frac{L_k(\tau)}{L_k(0)} }
\end{equation}
The hyperparameter \(\beta \geq 0\) controls the strength of balancing.

To sum up, by minimizing \(L_{\text{grad}}\), the scales \(s_1, s_2\) are updated to balance the gradient magnitudes according to relative loss progress. Note that, the scales \(s_j\) are renormalized after each update to satisfy \(s_1 + s_2 = 1\). This process ensures that the crisp LPs \(\theta\) and uncertainty LPs \(\tilde{\Delta}\) are trained jointly and adaptively, resulting in balanced and efficient learning.

\section{Comparative Performance Analysis}\label{sec6}
This section evaluates the learning performance of INNs on four SysID benchmark datasets: MR Damper~\cite{wang2009identification}, Heat Exchanger~\cite{DeMoorBLR1997}, Hair Dryer~\cite{ljung1999system}, and Robot Arm~\cite{ljung1999system}. Two base architectures---ILSTM and INODE---are implemented as described in Sections~\ref{ILSTM} and~\ref{INODE}. For each architecture, we apply two parameterization tricks during learning to analyze their effect on the SysID performance. Thus, four INN training configurations are defined:
\begin{itemize}
    \item ILSTM-1: ILSTM trained with  \(\sigma^{\text{ReLU}}\).
    \item ILSTM-2: ILSTM trained with  \(\sigma^{\text{abs}}\).
    \item INODE-1: INODE trained with  \(\sigma^{\text{ReLU}}\).
    \item INODE-2: INODE trained with  \(\sigma^{\text{abs}}\).
\end{itemize}
The C-INN and J-INN strategies are compared across these configurations to assess both point prediction and UQ capabilities. In addition, their results are benchmarked against commonly used probabilistic NN approaches presented in ~\ref{AppendixA}.

All experiments were conducted in MATLAB\textsuperscript{\textregistered} with ten different initial seeds for statistical analysis, using the dataset configurations and hyperparameters of all models are given in ~\ref{AppendixB} \footnote{MATLAB implementation. [Online]. Available: GitHub repo will be shared upon acceptance of the paper.}. All datasets were preprocessed using $z$-score normalization before training. For each experiment, the point prediction performance is assessed using the Root Mean Square Error (RMSE). UQ performance is examined using 
\begin{itemize}
    \item The Prediction Interval Coverage Probability (PICP)~\cite{6581855} measures the percentage of target values that lie within the PIs:  
    \begin{equation}
    \mathrm{PICP} = \frac{1}{N} \sum_{k=1}^{N} d(k) \times 100\%, 
    \quad 
    d(k) =
    \begin{cases}
    1, & \text{if } \underline{y}(k) \le   \hat{y}(k) \le \overline{y}(k), \\
    0, & \text{otherwise,}
    \end{cases}
    \end{equation}
        
    \item The PI Normalized Average Width (PINAW)~\cite{6581855} evaluates the average width of the PIs, normalized by the range of the observed target values:
    \begin{equation}
    \mathrm{PINAW} = \frac{1}{N \cdot R} \sum_{k=1}^{N} w(k),
    \quad
    w(k) =  \overline{y}(k) -  \underline{y}(k) ,
    \end{equation}
    where $w(k)$ is the interval width for the $k$th sample, and 
    \begin{equation}
    R = \max_{1 \le k \le N} \hat{y}(k) - \min_{1 \le k \le N} \hat{y}(k)
    \end{equation}
    is the range of the observed target values.

    \item The Coverage Width-based Criterion (CWC) ~\cite{cwc} is employed as a metric to evaluate the quality of PIs. It is defined as follows:

\begin{equation}
\text{CWC} = \text{PINAW} \cdot \left(1 + I(\text{PICP} < \alpha) \cdot e^{-
\eta  (\text{PICP}-\alpha)} \right),
\end{equation}
where \( I(\cdot) \) is the indicator function, which returns 1 if the condition inside holds and 0 otherwise. When the PICP meets or exceeds the coverage level \(\alpha\), the CWC closely matches the PINAW. If the PICP falls below \(\alpha\), a penalty increases the CWC value. Therefore, lower CWC values indicate better quality of the PIs. In this study, the penalty parameter \(\eta\) is set to 25~\cite{cwc}. This metric is particularly relevant for evaluating the overall UQ performance, as it simultaneously accounts for both PICP and PINAW.
\end{itemize}

The SysID results are examined for two coverage levels, \(\alpha = 0.90\) and \(\alpha = 0.95\). We report the mean and standard deviation of all metrics, along with boxplots for statistical comparison.  PICP values in the boxplots represent deviations from the target coverage. To comprehensively assess the results, the investigation is structured around a fourfold analysis:

\begin{enumerate}[label=(\roman*)]
    \item \textbf{Sections \ref{cascade} and \ref{joint}} evaluate and compare the performance of the INNs on SysID tasks, with models trained under the proposed learning strategies, C-INN and J-INN, respectively. These analyses provide an inter-architecture comparison by fixing the learning strategy and examining differences across model architectures.
    \item \textbf{Section \ref{casjoin}} summarizes the strengths of the C-INN and J-INN learning strategies, evaluating how the choice of training method influences SysID performance in terms of point prediction and UQ. Furthermore, their performance is compared against trained baseline NN models to provide a comprehensive benchmark.
    \item \textbf{Section \ref{elas}} introduces the novel concepts of \textit{elasticity} and \textit{channel-wise elasticity}, which quantify parameter uncertainty within the INNs, offering deeper insights into the sources of uncertainty.
\end{enumerate} 

\subsection{Cascade INN Approach Performance Analysis}
\label{cascade}

\begin{figure}[t]

\centering
\subfloat[\(\alpha = 0.90\)]{%
    \includegraphics[width=0.48\textwidth]{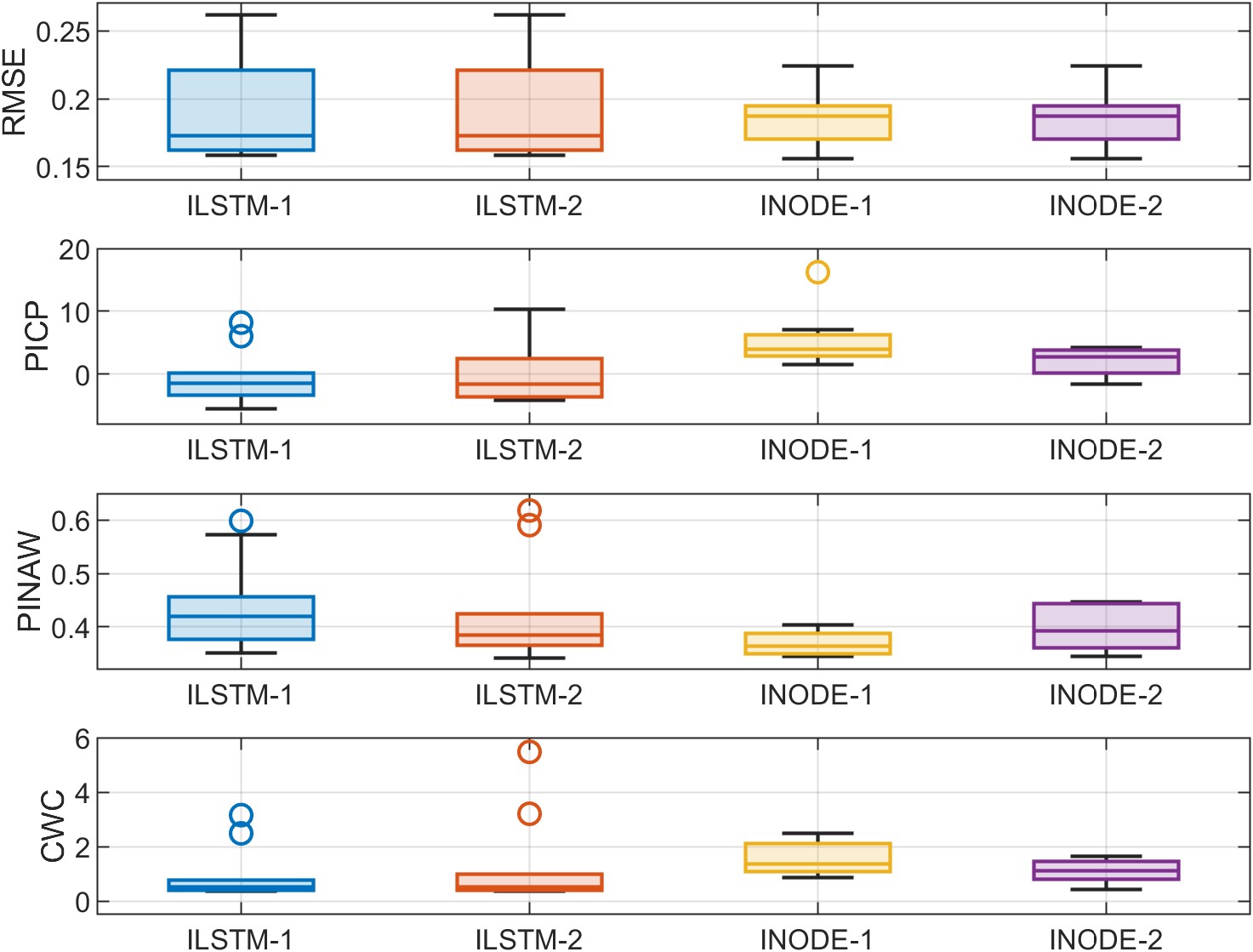}
}
\hfill
\subfloat[\(\alpha = 0.95\)]{%
    \includegraphics[width=0.48\textwidth]{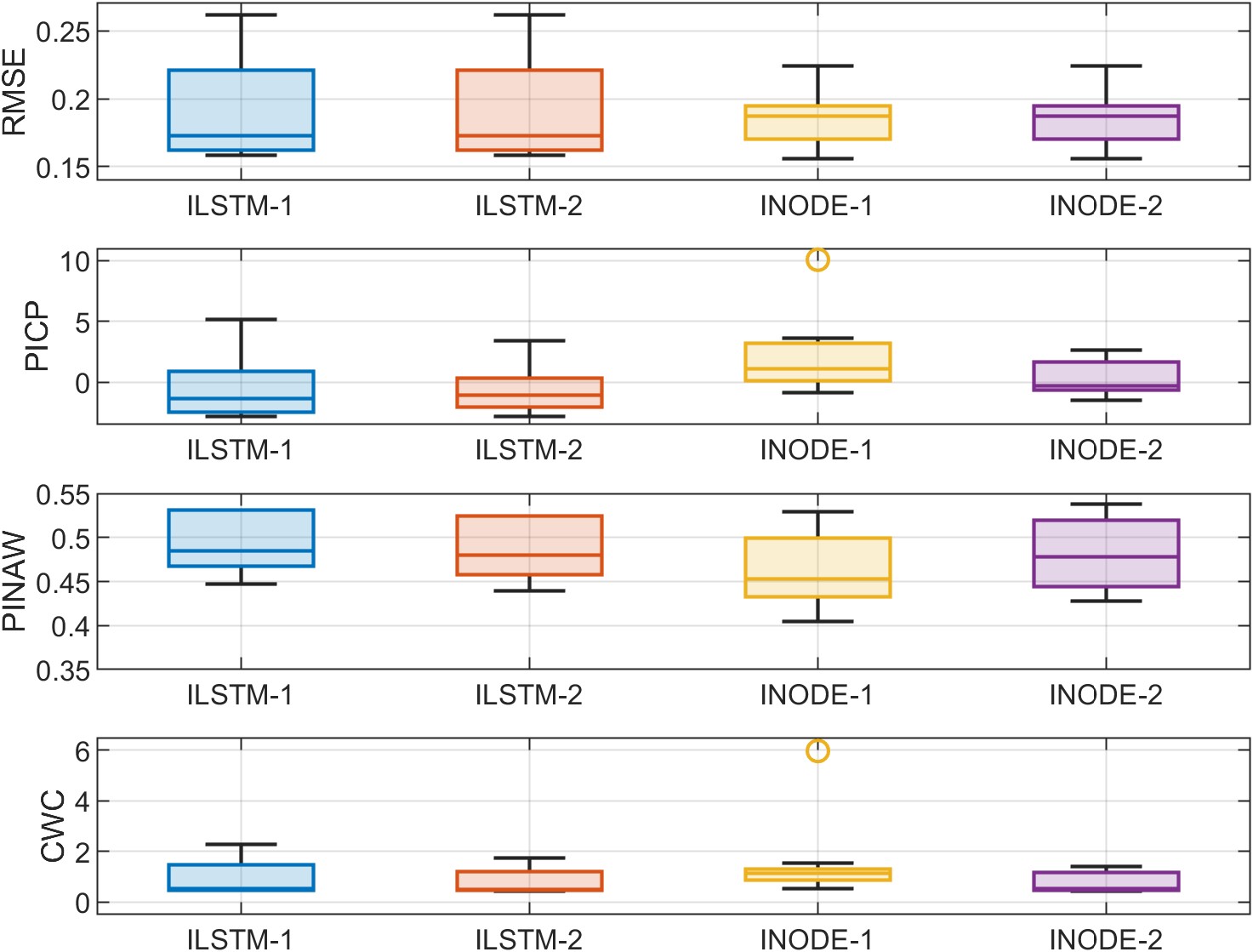}
}
\caption{Robot Arm Dataset: Performance Comparison for C-INN Strategy}
\label{fig:robot1}
\centering
\subfloat[\(\alpha = 0.90\)]{%
    \includegraphics[width=0.48\textwidth]{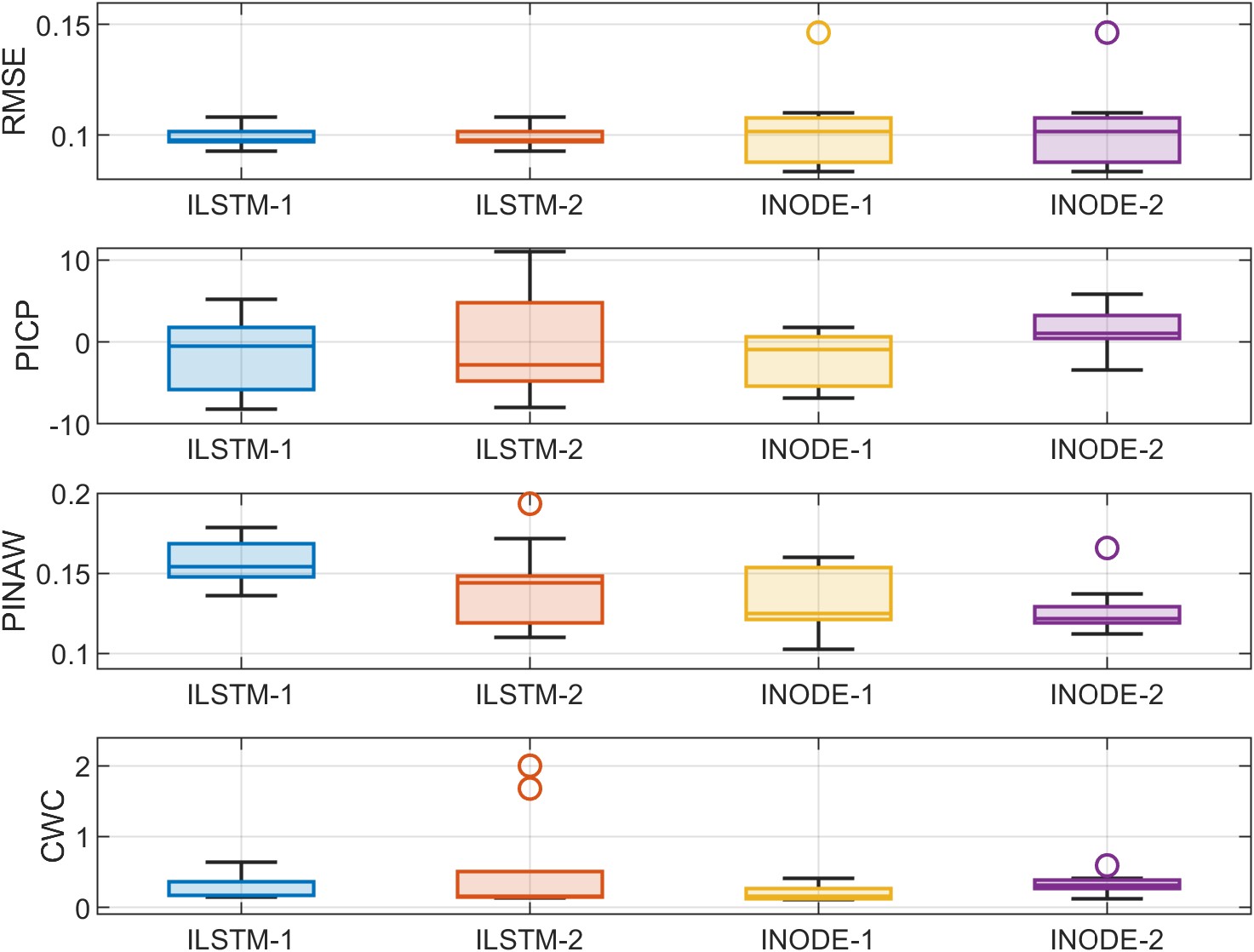}
}
\hfill
\subfloat[\(\alpha = 0.95\)]{%
 \includegraphics[width=0.48\textwidth]{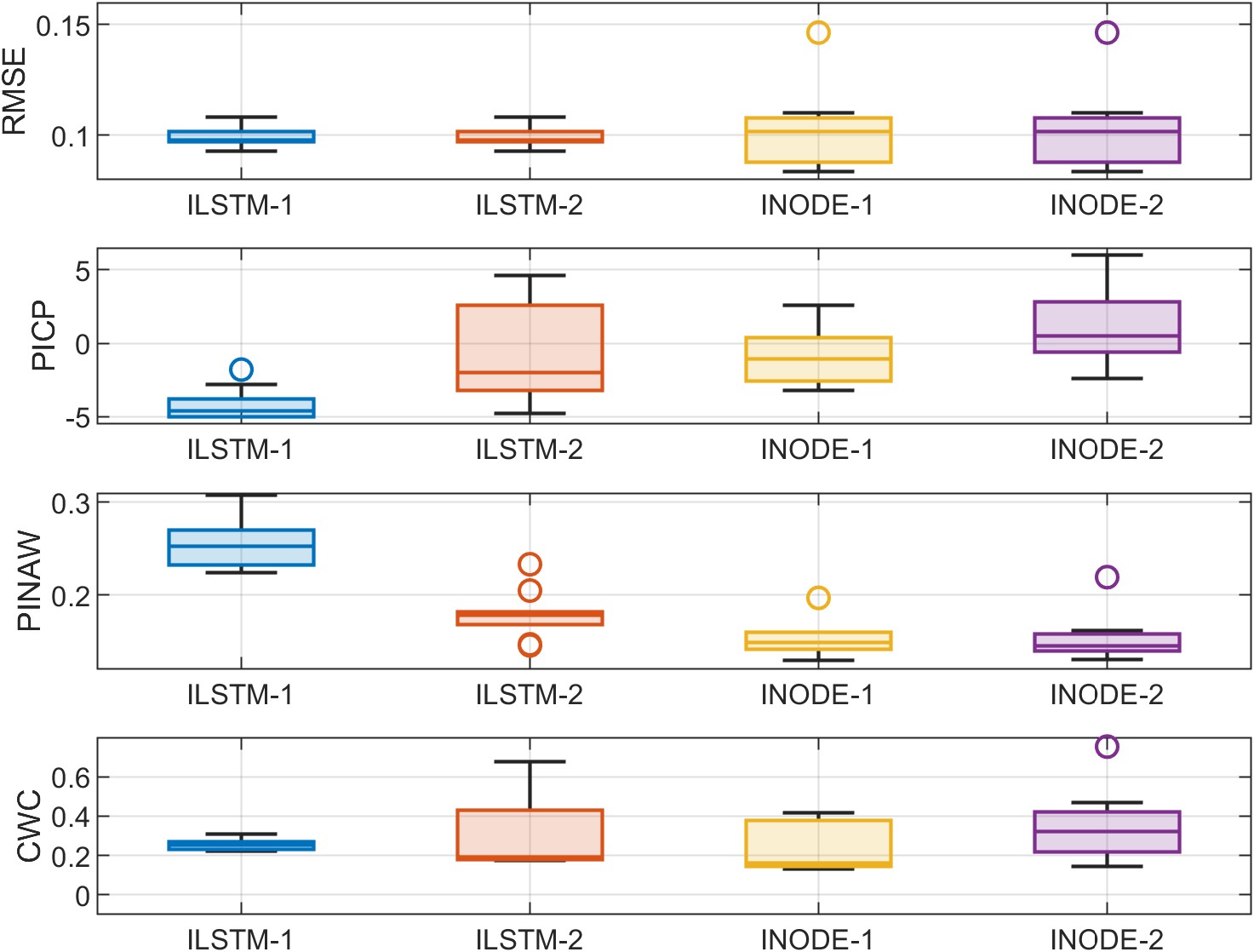}
}
\caption{Hair Dryer Dataset: Performance Comparison for C-INN Strategy}
\label{fig:dryer1}
\end{figure}

\begin{figure}[t]

\centering
\subfloat[\(\alpha = 0.90\)]{%
    \includegraphics[width=0.48\textwidth]{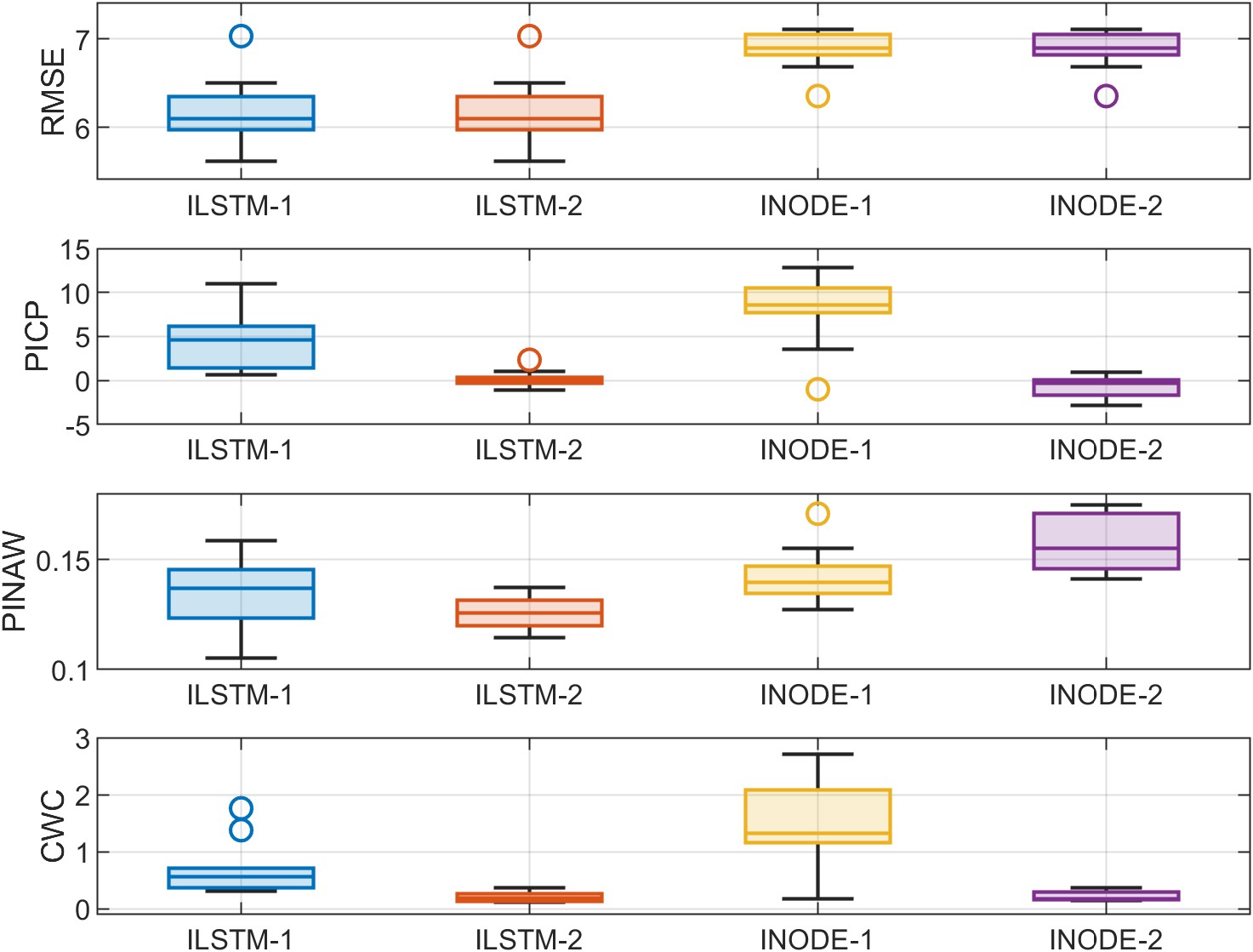}
}
\hfill
\subfloat[\(\alpha = 0.95\)]{%
    \includegraphics[width=0.48\textwidth]{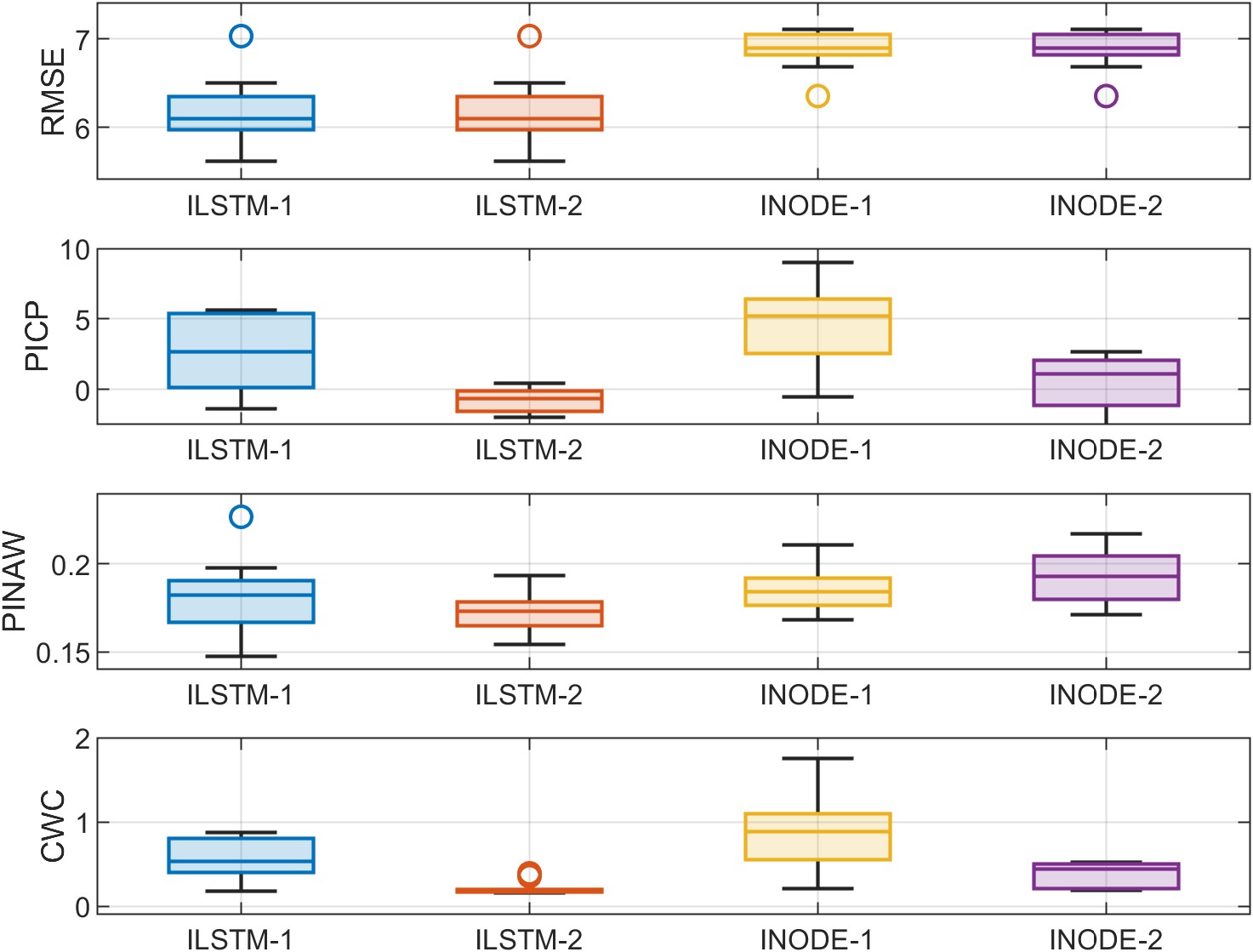}
}
\caption{MR-Damper Dataset: Performance Comparison for C-INN Strategy}
\label{fig:damper1}

\centering
\subfloat[\(\alpha = 0.90\)]{%
    \includegraphics[width=0.48\textwidth]{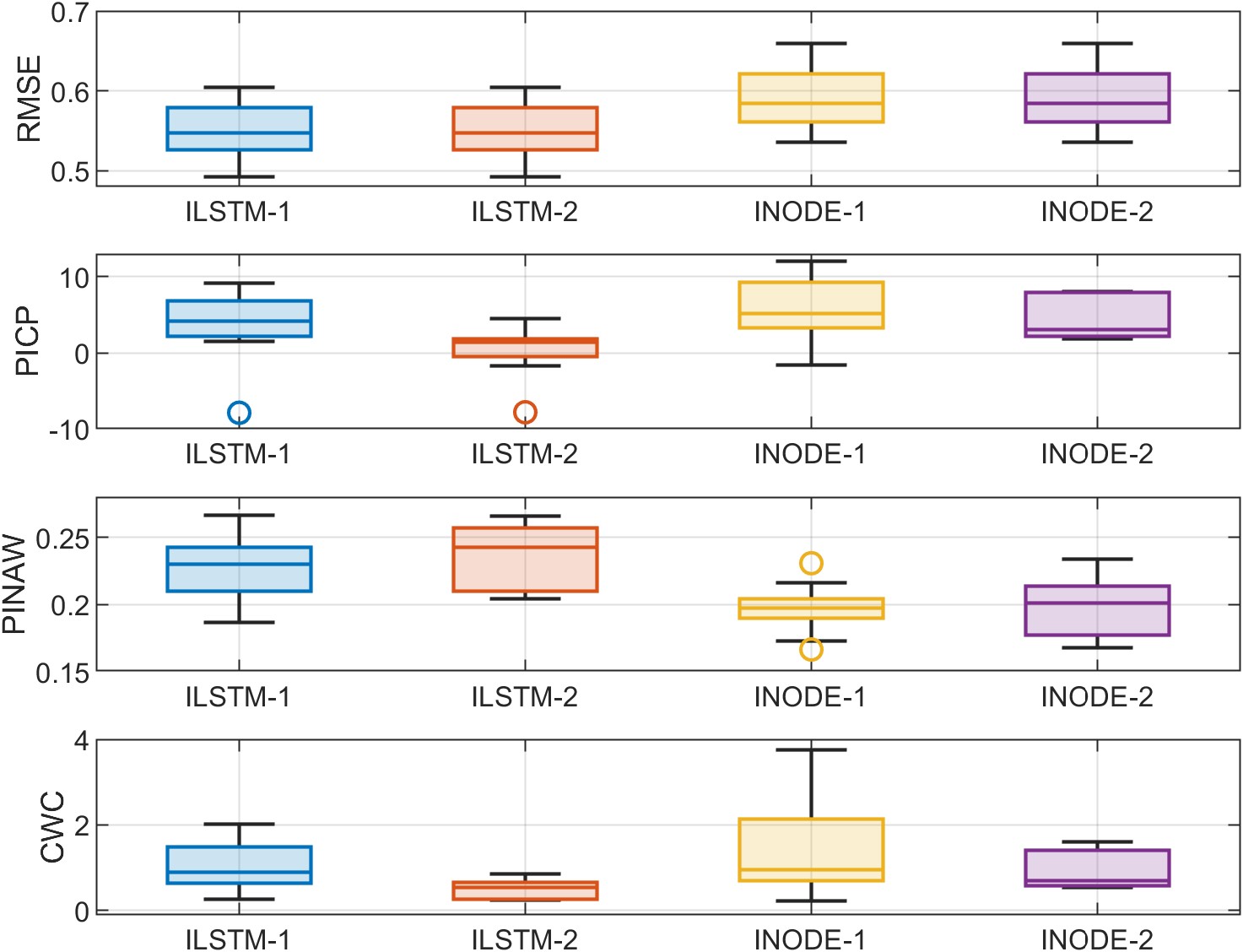}
}
\hfill
\subfloat[\(\alpha = 0.95\)]{%
    \includegraphics[width=0.48\textwidth]{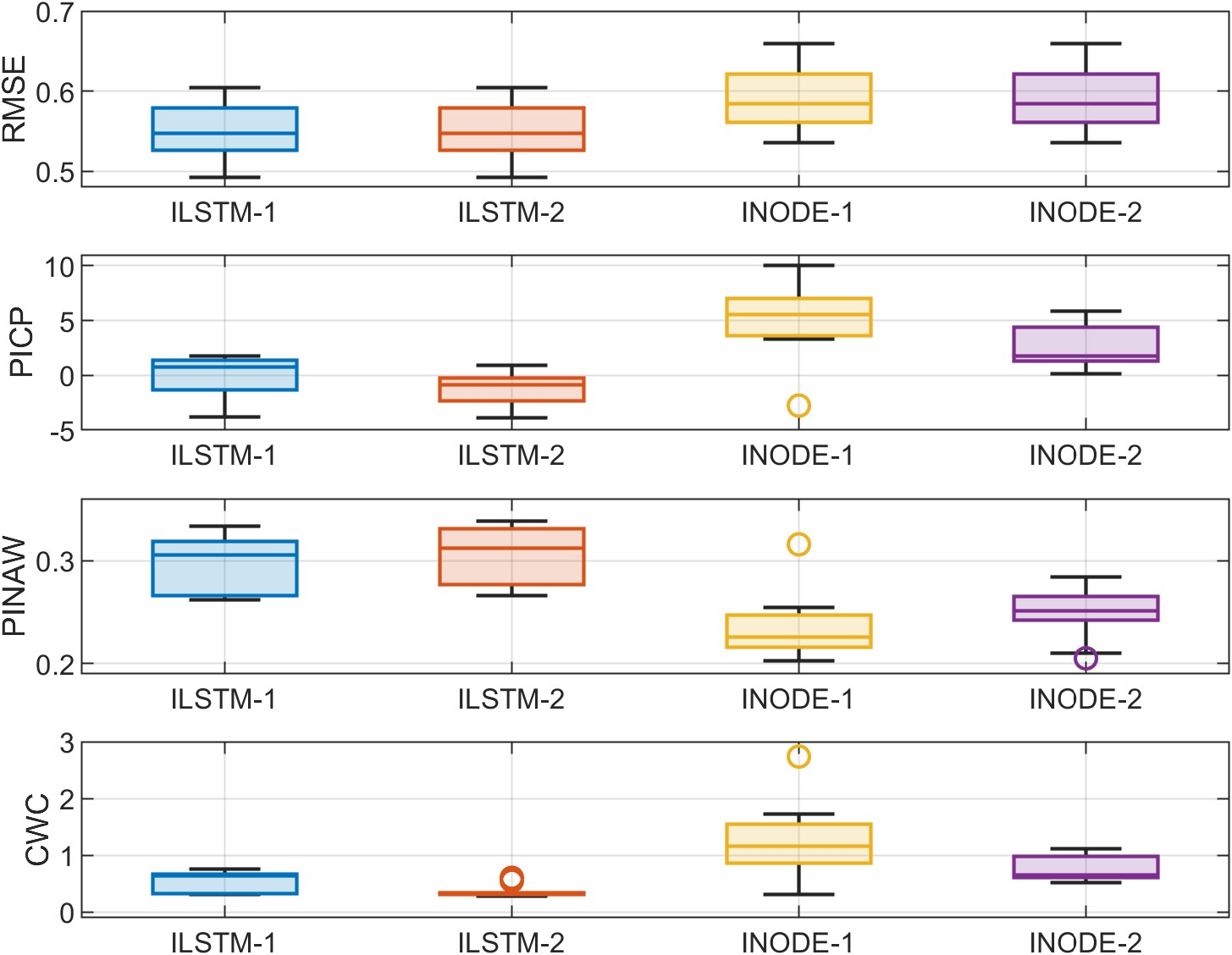}
}
\caption{Heat Exchanger Dataset: Performance Comparison for C-INN Strategy}
\label{fig:exchanger1}

\end{figure}

The overall performance results are presented in Table~\ref{tab:cnn_picp_pinaw-90} and Table~\ref{tab:cnn_picp_pinaw-95} for $\alpha=0.9$ and $\alpha=0.95$, respectively. The corresponding boxplots are illustrated in Figs.~\ref{fig:robot1}--\ref{fig:exchanger1}. We observe that: 
\begin{itemize}  
    \item The point prediction performance, measured by RMSE, is generally comparable across all models, except for the MR-Damper dataset (Fig.~\ref{fig:damper1}), where LSTM-based models show better performance. 
 \item Examining the boxplots, which illustrate the UQ performance, ILSTM-2 and INODE-2 demonstrate the best performance in terms of PICP, as they consistently reach the target coverage. Their PINAW performance is also comparable and closely aligned, which shows these models perform well. In particular, INODE-2 shows statistically robust behavior—evidenced by the small variance in Fig.\ref{fig:robot1} and Fig.\ref{fig:dryer1}. 
 
\item The CWC results in Tables~\ref{tab:cnn_picp_pinaw-90} and \ref{tab:cnn_picp_pinaw-95}, together with the boxplots in Figs.~\ref{fig:robot1}–\ref{fig:exchanger1}, indicate that ILSTM-2 and INODE-2 achieve comparable performance overall, while ILSTM-2 provides superior results for the MR-Damper and Heat Exchanger datasets.

 \item The results show a clear advantage of ILSTM-2 and INODE-2 over ILSTM-1 and INODE-1 across all datasets. Since the sole difference lies in the use of $\sigma^{\text{abs}}$ instead of $\sigma^{\text{ReLU}}$ during training, the improvement can be attributed to enhanced convergence behavior and reduced training instabilities, rather than architectural changes.
 
\end{itemize}

Under the C-INN strategy, ILSTM variants achieve identical prediction performance to INODE variants, as both are optimized solely for accuracy and then frozen. Their uncertainty LPs are learned independently to capture uncertainty. We can conclude that in the C-INN, as it is a two-stage training framework, point accuracy and UQ are effectively decoupled, with the parameterization trick $\sigma^{\text{abs}}$ enhancing the quality of uncertainty estimates (i.e., ILSTM-2 and INODE-2).

\begin{table}[t]
\caption{Evaluation of ILSTM and INODE performance at 90\% coverage under the C-INN strategy.}

\label{tab:cnn_picp_pinaw-90}
\centering
\begin{adjustbox}{max width=\textwidth}
\footnotesize
\renewcommand{\arraystretch}{0.8}
\setlength{\tabcolsep}{4pt}

\renewcommand{\arraystretch}{1.2}

\begin{threeparttable}
\begin{tabular}{|l| l|c|c|c|c|}
\hline
\textbf{Dataset} & \textbf{Metric} & \textbf{ILSTM-1} & \textbf{ILSTM-2} & \textbf{INODE-1} & \textbf{INODE-2} \\
\hline

\multirow{4}{*}{Robot Arm} 
& RMSE  & 19.39\((\pm3.85)\) & 19.39\((\pm3.85)\) & 18.57\((\pm1.82)\) & 18.57\((\pm1.82)\) \\
& PICP  & 90.41\((\pm4.15)\) & 89.75\((\pm4.87)\) & 84.74\((\pm4.06)\) & 87.95\((\pm2.13)\) \\
& PINAW & 43.83\((\pm8.17)\) & 42.31\((\pm9.33)\) & 36.97\((\pm2.03)\) & 39.86\((\pm3.88)\) \\
& CWC   & 0.96\((\pm0.95)\) & 1.28\((\pm1.62)\) & 3.44\((\pm5.82)\) & 1.08\((\pm0.42)\) \\

\hline

\multirow{4}{*}{Hair Dryer} 
& RMSE  & 9.88\((\pm0.42)\) & 9.88\((\pm0.42)\) & 10.27\((\pm1.72)\) & 10.27\((\pm1.72)\) \\
& PICP  & 91.44\((\pm4.54)\) & 90.32\((\pm6.55)\) & 91.98\((\pm2.97)\) & 88.68\((\pm2.45)\) \\
& PINAW & 15.66\((\pm1.33)\) & 14.27\((\pm2.41)\) & 13.08\((\pm1.88)\) & 12.68\((\pm1.45)\) \\
& CWC   & 0.30\((\pm0.18)\) & 0.53\((\pm0.67)\) & 0.20\((\pm0.09)\) & 0.32\((\pm0.13)\) \\

\hline

\multirow{4}{*}{MR-Damper}
& RMSE  & 617.40\((\pm36.73)\) & 617.40\((\pm36.73)\) & 686.86\((\pm21.36)\) & 686.86\((\pm21.36)\) \\
& PICP  & 85.24\((\pm3.40)\) & 89.79\((\pm0.89)\) & 81.98\((\pm3.88)\) & 90.73\((\pm1.21)\) \\
& PINAW & 13.25\((\pm1.68)\) & 12.59\((\pm0.72)\) & 14.24\((\pm1.23)\) & 15.70\((\pm1.25)\) \\
& CWC   & 0.71\((\pm0.46)\) & 0.21\((\pm0.09)\) & 1.56\((\pm0.95)\) & 0.22\((\pm0.08)\) \\

\hline

\multirow{4}{*}{Heat Exchanger}
& RMSE  & 54.86\((\pm3.35)\) & 54.86\((\pm3.35)\) & 59.06\((\pm4.13)\) & 59.06\((\pm4.13)\) \\
& PICP  & 86.21\((\pm4.58)\) & 89.51\((\pm3.24)\) & 84.10\((\pm3.92)\) & 85.39\((\pm2.73)\) \\
& PINAW & 22.79\((\pm2.46)\) & 23.69\((\pm2.30)\) & 19.71\((\pm1.77)\) & 19.95\((\pm2.10)\) \\
& CWC   & 1.07\((\pm0.54)\) & 0.52\((\pm0.20)\) & 1.43\((\pm1.03)\) & 0.95\((\pm0.42)\) \\

\hline
\end{tabular}
\begin{tablenotes}
\footnotesize
\item Note: RMSE and PINAW values are scaled by 100.
\end{tablenotes}
\end{threeparttable}
\end{adjustbox}
\end{table}

\begin{table}[t]
\caption{Evaluation of ILSTM and INODE performance at 95\% coverage under the C-INN strategy.}

\label{tab:cnn_picp_pinaw-95}
\centering
\begin{adjustbox}{max width=\textwidth}
\footnotesize
\renewcommand{\arraystretch}{0.8}
\setlength{\tabcolsep}{4pt}
\begin{threeparttable}

\renewcommand{\arraystretch}{1.2}

\begin{tabular}{|l| l|c|c|c|c|}
\hline
\textbf{Dataset} & \textbf{Metric} & \textbf{ILSTM-1} & \textbf{ILSTM-2} & \textbf{INODE-1} & \textbf{INODE-2} \\
\hline

\multirow{4}{*}{Robot Arm} 
& RMSE  & 19.39\((\pm3.85)\) & 19.39\((\pm3.85)\) & 18.57\((\pm1.82)\) & 18.57\((\pm1.82)\) \\
& PICP  & 95.44\((\pm2.57)\) & 95.62\((\pm1.90)\) & 92.97\((\pm3.02)\) & 94.81\((\pm1.35)\) \\
& PINAW & 52.20\((\pm8.40)\) & 52.23\((\pm9.61)\) & 46.54\((\pm4.17)\) & 48.23\((\pm3.73)\) \\
& CWC   & 0.89\((\pm0.60)\) & 0.79\((\pm0.46)\) & 1.52\((\pm1.52)\) & 0.78\((\pm0.38)\) \\

\hline

\multirow{4}{*}{Hair Dryer} 
& RMSE  & 9.88\((\pm0.42)\) & 9.88\((\pm0.42)\) & 10.27\((\pm1.72)\) & 10.27\((\pm1.72)\) \\
& PICP  & 99.20\((\pm1.04)\) & 95.83\((\pm3.27)\) & 95.77\((\pm1.98)\) & 93.93\((\pm2.49)\) \\
& PINAW & 25.75\((\pm2.83)\) & 17.92\((\pm2.43)\) & 15.58\((\pm2.27)\) & 15.32\((\pm2.38)\) \\
& CWC   & 0.26\((\pm0.03)\) & 0.30\((\pm0.18)\) & 0.23\((\pm0.11)\) & 0.35\((\pm0.17)\) \\

\hline

\multirow{4}{*}{MR-Damper}
& RMSE  & 617.40\((\pm36.73)\) & 617.40\((\pm36.73)\) & 686.86\((\pm21.36)\) & 686.86\((\pm21.36)\) \\
& PICP  & 92.31\((\pm2.50)\) & 95.77\((\pm0.75)\) & 90.24\((\pm2.69)\) & 94.55\((\pm1.76)\) \\
& PINAW & 18.22\((\pm2.00)\) & 17.32\((\pm1.23)\) & 18.54\((\pm1.23)\) & 19.27\((\pm1.46)\) \\
& CWC   & 0.56\((\pm0.23)\) & 0.21\((\pm0.08)\) & 0.89\((\pm0.42)\) & 0.37\((\pm0.14)\) \\

\hline

\multirow{4}{*}{Heat Exchanger}
& RMSE  & 54.86\((\pm3.35)\) & 54.86\((\pm3.35)\) & 59.06\((\pm4.13)\) & 59.06\((\pm4.13)\) \\
& PICP  & 94.95\((\pm1.71)\) & 96.18\((\pm1.38)\) & 89.85\((\pm3.27)\) & 92.40\((\pm1.92)\) \\
& PINAW & 29.63\((\pm2.61)\) & 30.47\((\pm2.72)\) & 23.46\((\pm3.09)\) & 24.82\((\pm2.36)\) \\
& CWC   & 0.56\((\pm0.17)\) & 0.37\((\pm0.11)\) & 1.26\((\pm0.64)\) & 0.76\((\pm0.21)\) \\

\hline
\end{tabular}
\begin{tablenotes}
\footnotesize
\item Note: RMSE and PINAW values are scaled by 100.
\end{tablenotes}
\end{threeparttable}
\end{adjustbox}
\end{table}

\subsection{Joint INN Approach Performance Analysis} \label{joint}

\begin{figure}[t]

\centering
\subfloat[\(\alpha = 0.90\)]{%
    \includegraphics[width=0.48\textwidth]{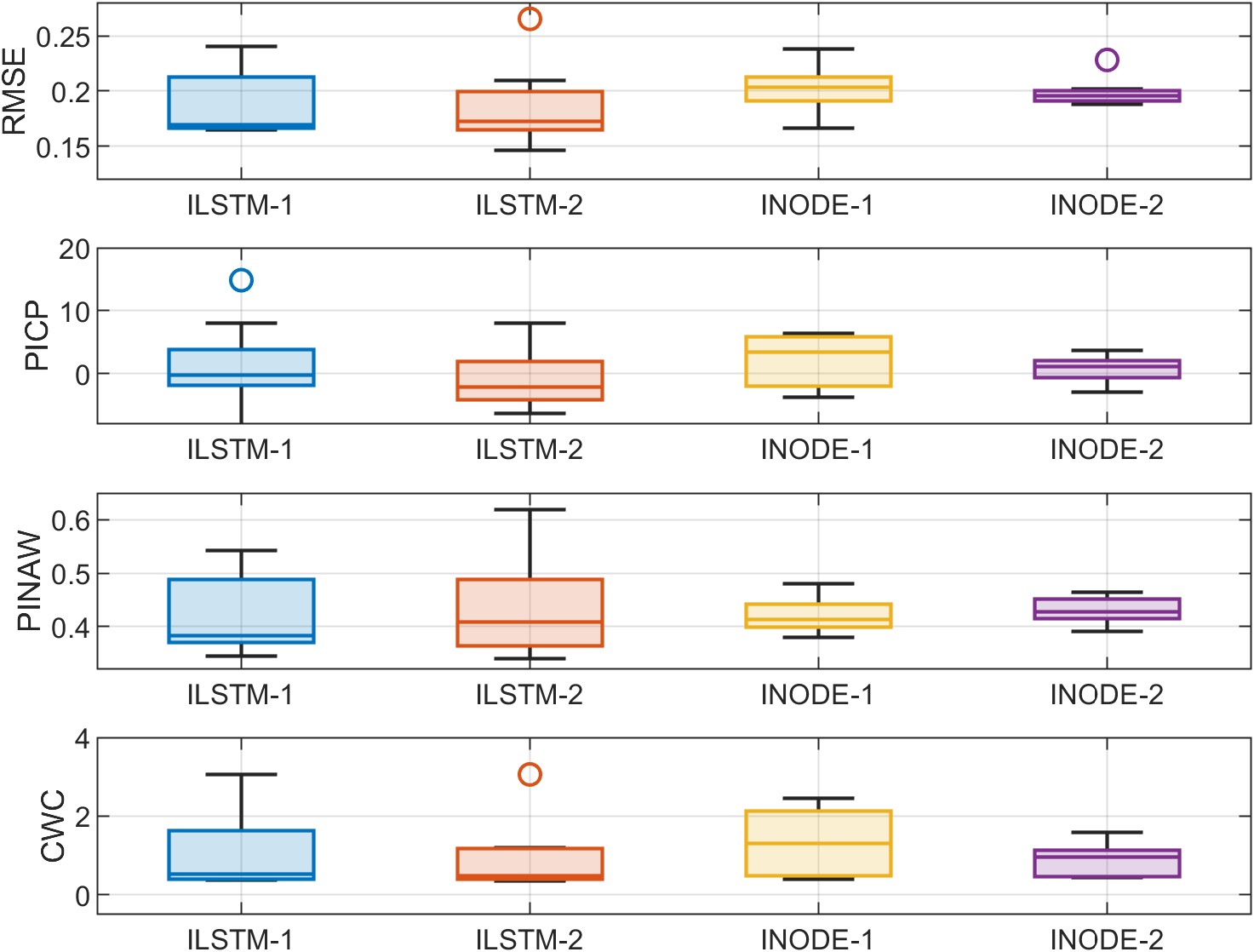}
}
\hfill
\subfloat[\(\alpha = 0.95\)]{%
    \includegraphics[width=0.48\textwidth]{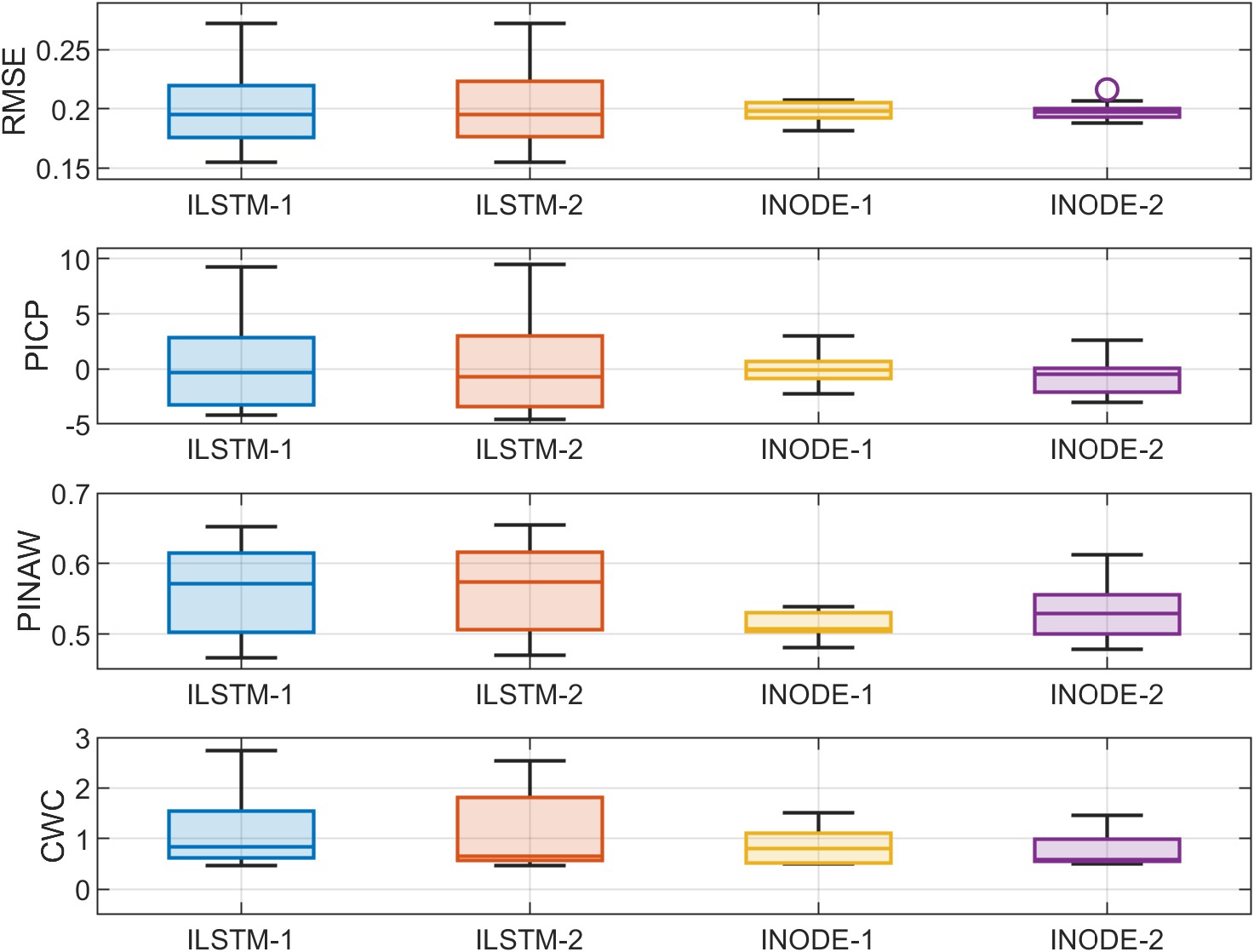}
}
\caption{Robot Arm Dataset: Performance Comparison for J-INN Strategy}
\label{fig:robot2}

\centering
\subfloat[\(\alpha = 0.90\)]{%
    \includegraphics[width=0.48\textwidth]{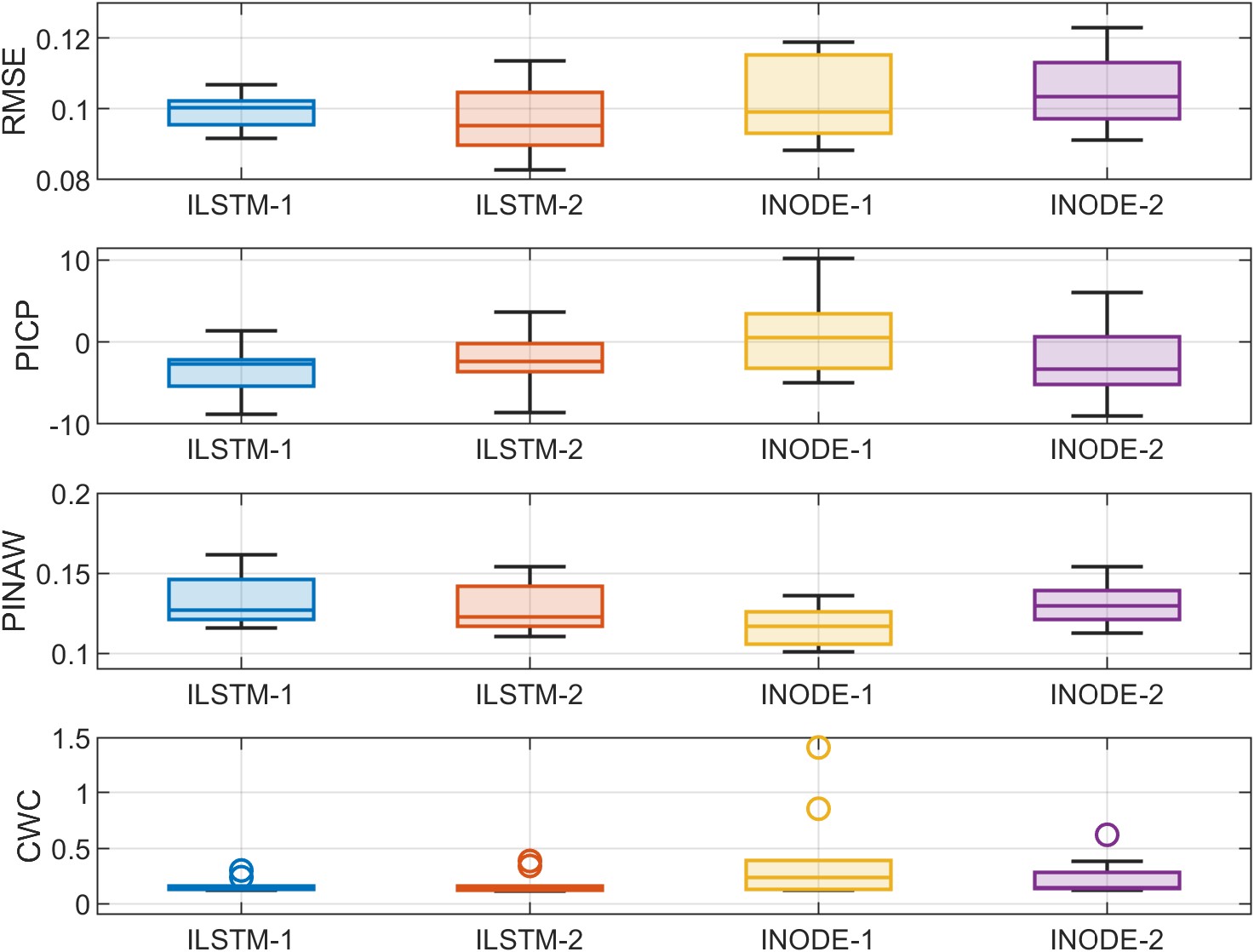}
}
\hfill
\subfloat[\(\alpha = 0.95\)]{%
    \includegraphics[width=0.48\textwidth]{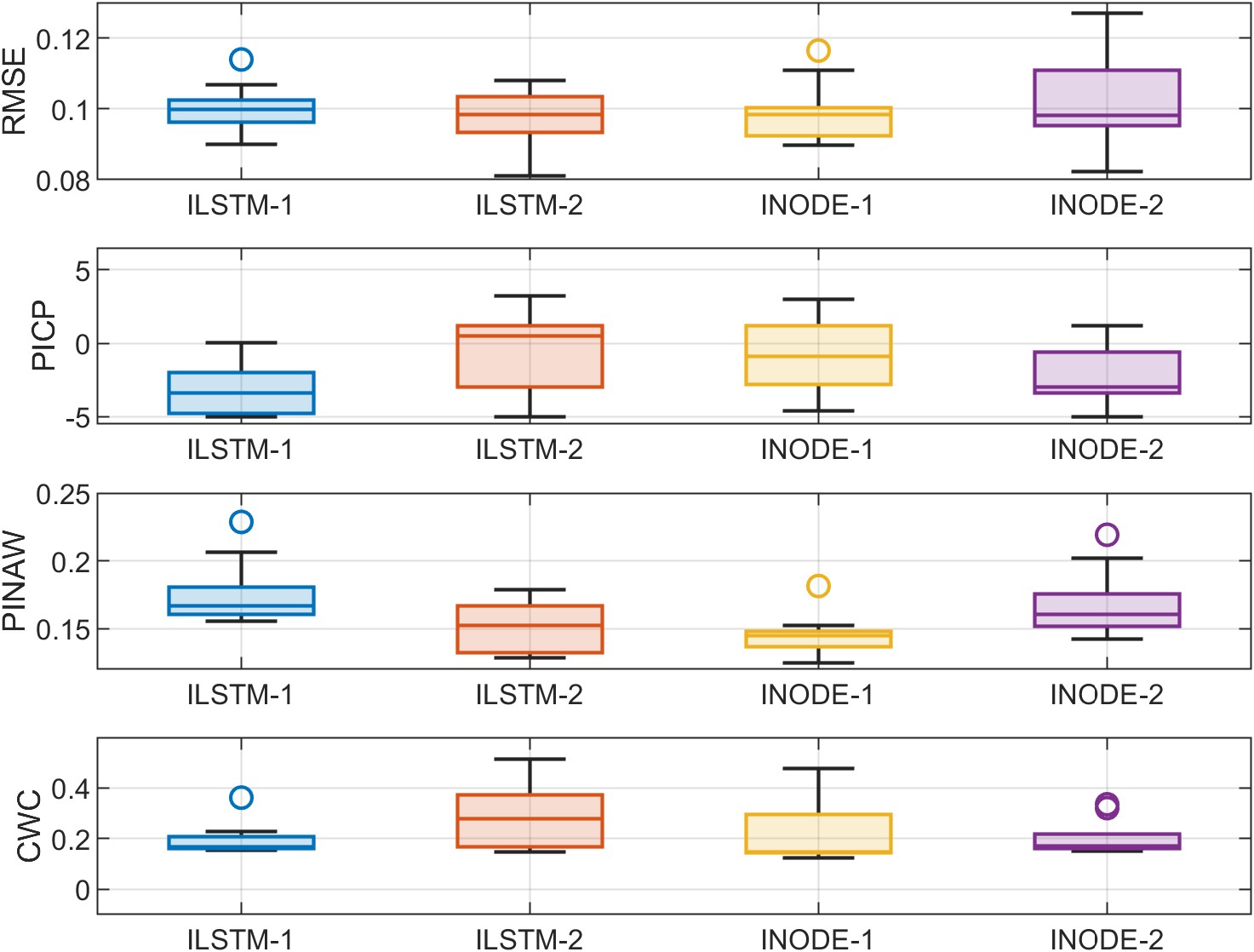}
}
\caption{Hair Dryer Dataset: Performance Comparison for J-INN Strategy}
\label{fig:dryer2}
\end{figure}

\begin{figure}[t]

\centering
\subfloat[\(\alpha = 0.90\)]{%
    \includegraphics[width=0.48\textwidth]{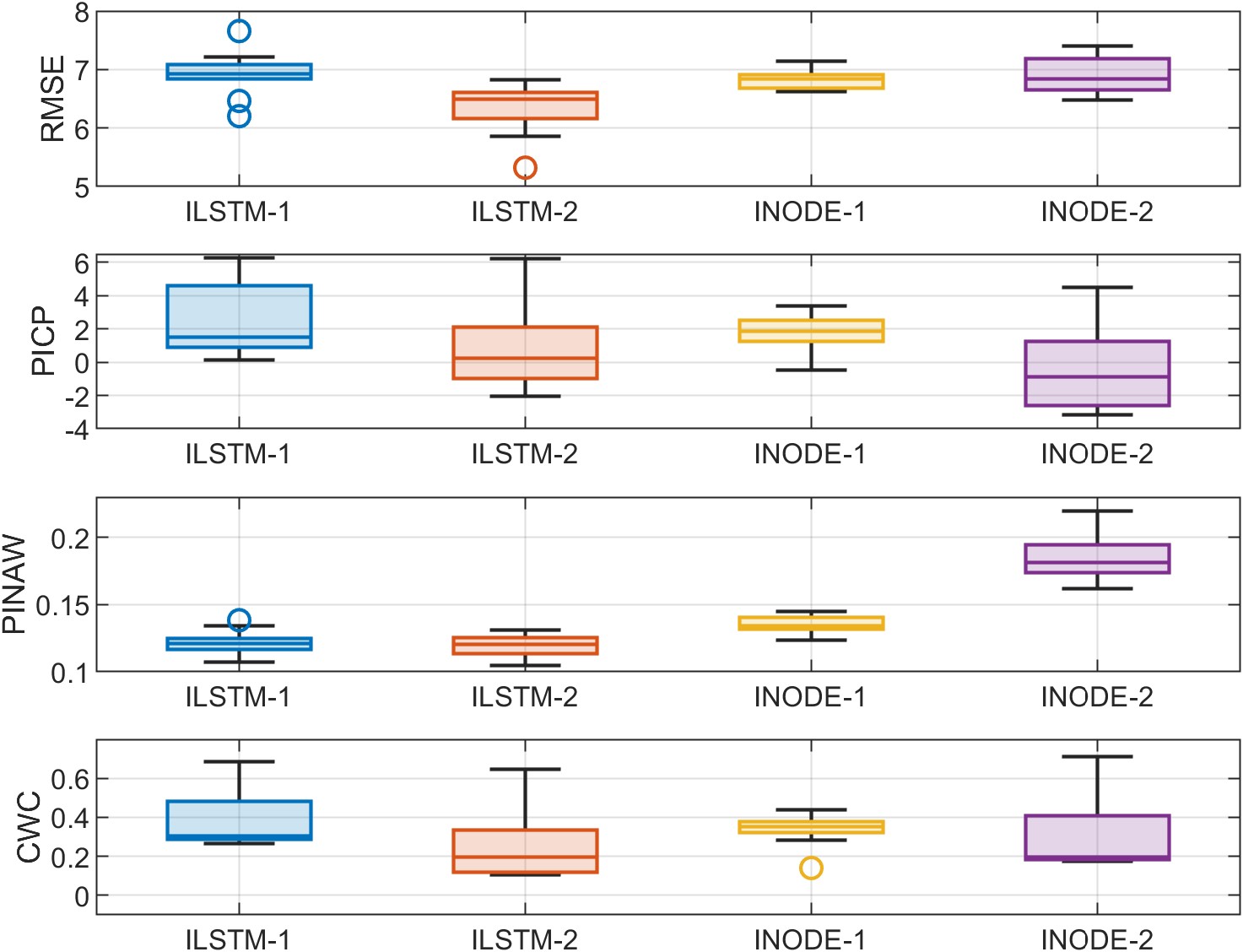}
}
\hfill
\subfloat[\(\alpha = 0.95\)]{%
    \includegraphics[width=0.48\textwidth]{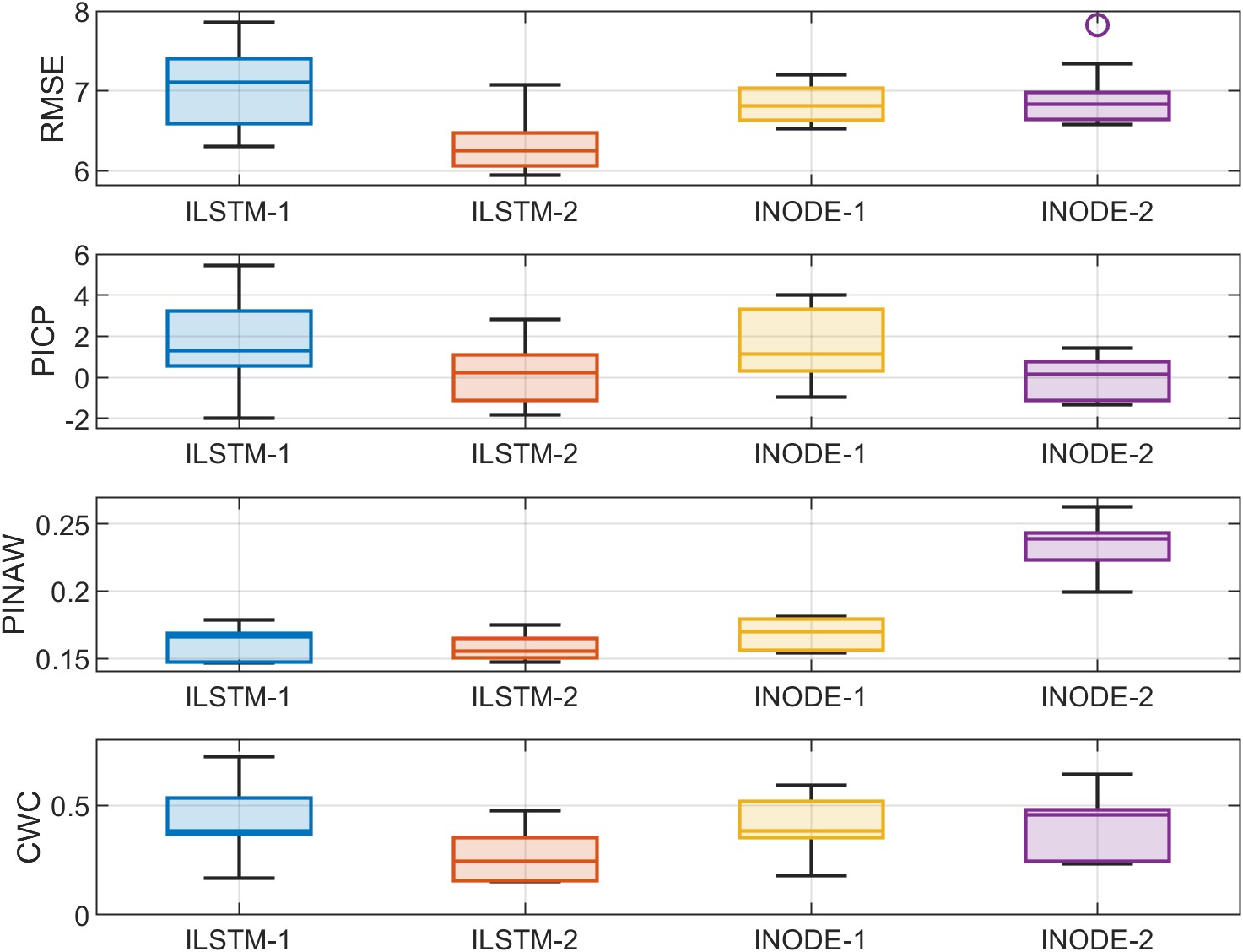}
}
\caption{MR-Damper Dataset: Performance Comparison for J-INN Strategy}
\label{fig:damper2}

\centering
\subfloat[\(\alpha = 0.90\)]{%
    \includegraphics[width=0.48\textwidth]{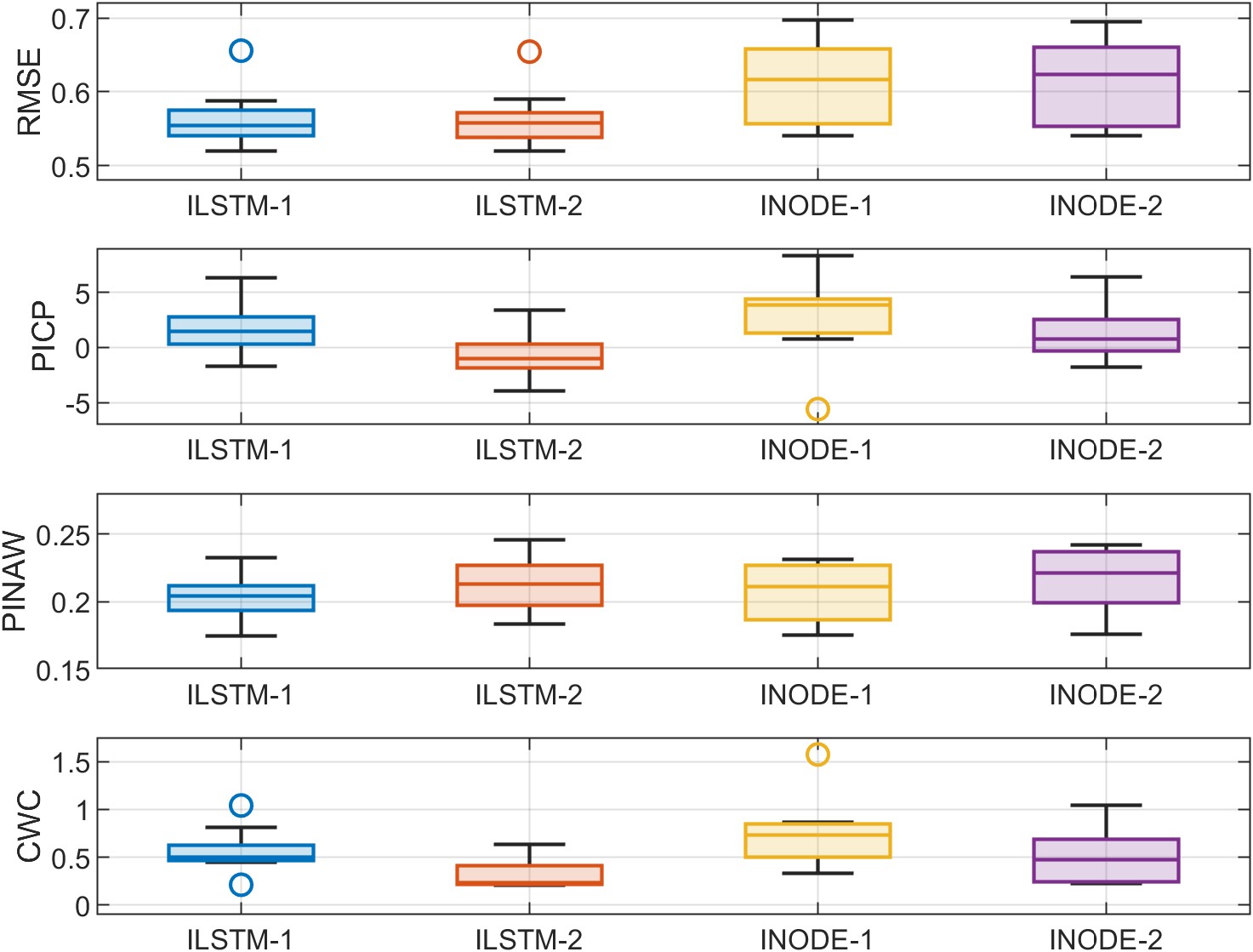}
}
\hfill
\subfloat[\(\alpha = 0.95\)]{%
    \includegraphics[width=0.48\textwidth]{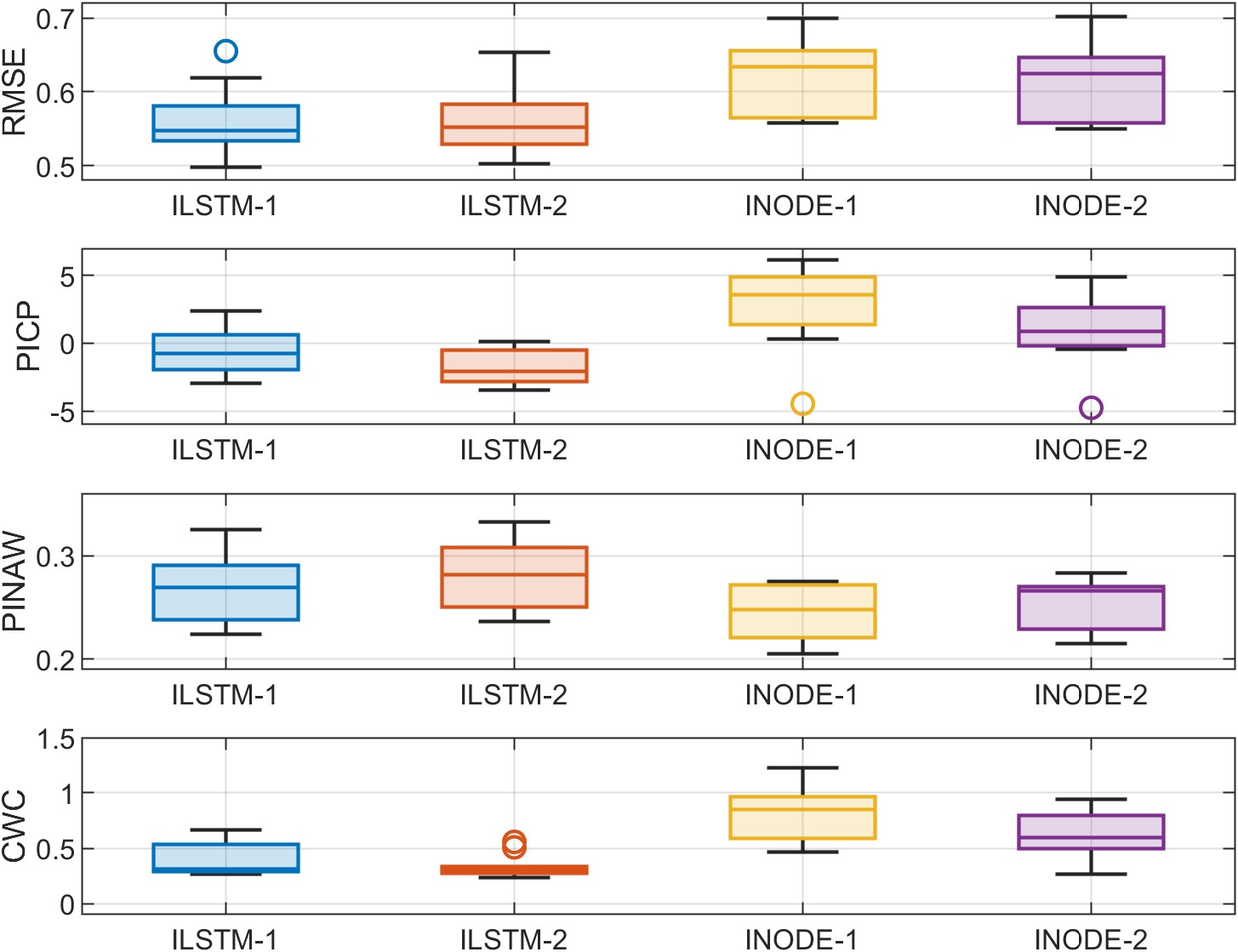}
}
\caption{Heat Exchanger Dataset: Performance Comparison for J-INN Strategy}
\label{fig:exchanger2}

\end{figure}

The overall performance results are presented in Table~\ref{tab:jnn_picp_pinaw_90} and Table~\ref{tab:jnn_picp_pinaw_95} for $\alpha=0.9$ and $\alpha=0.95$, respectively, while the corresponding boxplots for J-INN are shown in Figs.~\ref{fig:robot2}--\ref{fig:exchanger2}. We observe that: 
\begin{itemize}  
    \item The point prediction performance, measured by RMSE, is generally comparable across models, except for the MR-Damper dataset (Fig.~\ref{fig:damper2}) and the Heat Exchanger dataset (Fig.~\ref{fig:exchanger2}), where noticeable deviations are observed. For the remaining datasets, the models exhibit similar RMSE distributions, indicating consistent point prediction accuracy.
    
    \item Examining the box plot, ILSTM-2 and INODE-2 achieve the best PICP performance. While their PICP values are similar, INODE-2 demonstrates lower variance, especially for \(\alpha=0.95\) target coverage shown in Table~\ref{tab:jnn_picp_pinaw_95}. The J-INN framework exhibits consistent statistical robustness across all datasets, with only a few cases showing higher variance, such as ILSTM-1 in Fig.~\ref{fig:robot2}. Their PINAW performance is also comparable and closely aligned, which shows these models perform well.
   \item In terms of CWC results in Table~\ref{tab:jnn_picp_pinaw_90} and Table~\ref{tab:jnn_picp_pinaw_95}, as well as in the boxplots Figs. \ref{fig:robot2}-\ref{fig:exchanger2}, ILSTM-2 achieves the best performance across almost all datasets, with some exceptions. For example, in the Hair Dryer Dataset, ILSTM-1 shows the best performance. However, the results of ILSTM-2 for this dataset are also very close to those of ILSTM-1.
   
   \item Similar to C-INN, the results show a clear advantage of ILSTM-2 and INODE-2 over ILSTM-1 and INODE-1 across all datasets, indicating that using $\sigma^{\text{abs}}$ instead of $\sigma^{\text{ReLU}}$ during training is generally more effective. Yet, the improvement is less pronounced than in C-INN, because J-INN already achieves better UQ performance with ILSTM-1 and INODE-1.  
\end{itemize}

To sum up, J-INN delivers robust SysID performance across both ILSTM and INODE variants, with only minor differences observed between models. Unlike C-INN, it jointly optimizes point predictions and uncertainty intervals by learning the point prediction accuracy and PI Quality loss components simultaneously rather than sequentially. This one-stage training paradigm effectively balances accuracy and uncertainty estimation. In addition, the parameterization trick $\sigma^{\text{abs}}$ enhances the stability and reliability of the predictive intervals, yielding particularly strong results for the ILSTM-2 and INODE-2 variants.

\begin{table}[t]
\caption{Evaluation of ILSTM and INODE performance at 90\% coverage under the J-INN strategy.}
\label{tab:jnn_picp_pinaw_90}
\centering
\begin{adjustbox}{max width=\textwidth}
\footnotesize
\renewcommand{\arraystretch}{0.8}
\setlength{\tabcolsep}{4pt}
\begin{threeparttable}
\renewcommand{\arraystretch}{1.2}

\begin{tabular}{|l| l|c|c|c|c|}
\hline
{\textbf{Dataset}} & {\textbf{Metric}} 
& \textbf{ILSTM-1} & \textbf{ILSTM-2} & \textbf{INODE-1} & \textbf{INODE-2} \\
\hline

\multirow{4}{*}{Robot Arm}
& RMSE & 18.75\((\pm2.66)\) & 18.35\((\pm3.22)\) & 20.10\((\pm1.81)\) & 19.82\((\pm1.09)\) \\
& PICP & 88.88\((\pm6.53)\) & 91.04\((\pm4.18)\) & 87.67\((\pm3.59)\) & 89.20\((\pm1.84)\) \\
& PINAW & 42.25\((\pm6.88)\) & 43.07\((\pm8.09)\) & 41.90\((\pm2.93)\) & 43.04\((\pm2.18)\) \\
& CWC & 2.43\((\pm4.53)\) & 0.86\((\pm0.79)\) & 1.37\((\pm0.75)\) & 0.91\((\pm0.35)\) \\

\hline
\multirow{4}{*}{Hair Dryer}
& RMSE & 9.94\((\pm0.49)\) & 9.72\((\pm0.96)\) & 10.30\((\pm1.12)\) & 10.50\((\pm0.94)\) \\
& PICP & 93.59\((\pm3.19)\) & 92.08\((\pm3.25)\) & 89.16\((\pm4.84)\) & 92.24\((\pm4.14)\) \\
& PINAW & 13.32\((\pm1.52)\) & 12.86\((\pm1.42)\) & 11.60\((\pm1.09)\) & 13.14\((\pm1.30)\) \\
& CWC & 0.16\((\pm0.06)\) & 0.18\((\pm0.09)\) & 0.39\((\pm0.40)\) & 0.22\((\pm0.15)\) \\

\hline
\multirow{4}{*}{MR-Damper}
& RMSE & 691.93\((\pm37.46)\) & 632.46\((\pm43.40)\) & 684.74\((\pm16.44)\) & 689.90\((\pm28.88)\) \\
& PICP & 87.43\((\pm2.20)\) & 89.28\((\pm2.34)\) & 88.25\((\pm1.09)\) & 90.43\((\pm2.41)\) \\
& PINAW & 12.18\((\pm0.89)\) & 11.90\((\pm0.75)\) & 13.45\((\pm0.63)\) & 18.55\((\pm1.60)\) \\
& CWC & 0.39\((\pm0.15)\) & 0.25\((\pm0.16)\) & 0.34\((\pm0.08)\) & 0.29\((\pm0.17)\) \\

\hline
\multirow{4}{*}{Heat Exchanger}
& RMSE & 56.33\((\pm3.65)\) & 56.38\((\pm3.62)\) & 64.52\((\pm11.96)\) & 64.58\((\pm12.02)\) \\
& PICP & 88.09\((\pm2.26)\) & 90.59\((\pm2.03)\) & 87.15\((\pm3.44)\) & 89.20\((\pm3.77)\) \\
& PINAW & 20.30\((\pm1.69)\) & 21.31\((\pm1.81)\) & 21.73\((\pm4.12)\) & 22.63\((\pm4.45)\) \\
& CWC & 0.56\((\pm0.21)\) & 0.31\((\pm0.15)\) & 0.75\((\pm0.32)\) & 0.52\((\pm0.29)\) \\

\hline
\end{tabular}
\begin{tablenotes}
\footnotesize
\item Note: RMSE and PINAW values are scaled by 100.
\end{tablenotes}
\end{threeparttable}
\end{adjustbox}
\end{table}

\begin{table}[t]
\caption{Evaluation of ILSTM and INODE performance at 95\% coverage under the J-INN strategy.}

\label{tab:jnn_picp_pinaw_95}
\centering
\begin{adjustbox}{max width=\textwidth}
\footnotesize
\renewcommand{\arraystretch}{0.8}
\setlength{\tabcolsep}{4pt}
\begin{threeparttable}
\renewcommand{\arraystretch}{1.2}

\begin{tabular}{|l| l|c|c|c|c|}
\hline
{\textbf{Dataset}} & {\textbf{Metric}} 
& \textbf{ILSTM-1} & \textbf{ILSTM-2} & \textbf{INODE-1} & \textbf{INODE-2} \\
\hline

\multirow{4}{*}{Robot Arm}
& RMSE & 20.12\((\pm3.31)\) & 20.15\((\pm3.36)\) & 19.74\((\pm0.80)\) & 19.82\((\pm0.80)\) \\
& PICP & 94.48\((\pm4.25)\) & 94.54\((\pm4.38)\) & 95.01\((\pm1.36)\) & 95.68\((\pm1.60)\) \\
& PINAW & 55.82\((\pm6.33)\) & 56.01\((\pm6.23)\) & 51.34\((\pm1.72)\) & 53.15\((\pm3.94)\) \\
& CWC & 1.62\((\pm1.76)\) & 1.61\((\pm1.86)\) & 0.85\((\pm0.35)\) & 0.79\((\pm0.31)\) \\

\hline
\multirow{4}{*}{Hair Dryer}
& RMSE & 10.00\((\pm0.66)\) & 9.69\((\pm0.87)\) & 9.92\((\pm0.81)\) & 10.18\((\pm1.19)\) \\
& PICP & 98.14\((\pm1.64)\) & 95.37\((\pm2.57)\) & 95.87\((\pm2.43)\) & 97.21\((\pm1.98)\) \\
& PINAW & 17.55\((\pm2.27)\) & 15.09\((\pm1.72)\) & 14.44\((\pm1.52)\) & 16.76\((\pm2.35)\) \\
& CWC & 0.19\((\pm0.06)\) & 0.28\((\pm0.11)\) & 0.22\((\pm0.12)\) & 0.20\((\pm0.07)\) \\

\hline
\multirow{4}{*}{MR-Damper}
& RMSE & 709.56\((\pm46.97)\) & 634.59\((\pm38.33)\) & 683.15\((\pm22.31)\) & 692.71\((\pm37.01)\) \\
& PICP & 93.32\((\pm2.32)\) & 94.85\((\pm1.42)\) & 93.56\((\pm1.61)\) & 95.02\((\pm0.96)\) \\
& PINAW & 16.28\((\pm1.13)\) & 15.78\((\pm0.86)\) & 16.81\((\pm1.02)\) & 23.35\((\pm1.78)\) \\
& CWC & 0.42\((\pm0.17)\) & 0.27\((\pm0.12)\) & 0.40\((\pm0.14)\) & 0.40\((\pm0.14)\) \\

\hline
\multirow{4}{*}{Heat Exchanger}
& RMSE & 56.07\((\pm4.41)\) & 56.21\((\pm4.28)\) & 65.82\((\pm12.98)\) & 65.41\((\pm13.23)\) \\
& PICP & 95.58\((\pm1.57)\) & 96.72\((\pm1.25)\) & 92.32\((\pm2.91)\) & 93.93\((\pm2.56)\) \\
& PINAW & 27.08\((\pm3.05)\) & 28.23\((\pm3.01)\) & 26.41\((\pm7.16)\) & 27.62\((\pm7.70)\) \\
& CWC & 0.40\((\pm0.14)\) & 0.34\((\pm0.10)\) & 0.81\((\pm0.22)\) & 0.60\((\pm0.21)\) \\

\hline
\end{tabular}
\begin{tablenotes}
\footnotesize
\item Note: RMSE and PINAW values are scaled by 100.
\end{tablenotes}
\end{threeparttable}
\end{adjustbox}
\end{table}

\subsection{Comparison Between C-INN and J-INN Learning Strategies} \label{casjoin}
Sections 6.2 and 6.3 reveal complementary strengths of the two approaches. C-INN generally achieves lower RMSE due to its two-stage design, which first trains a crisp network solely for point prediction accuracy. In contrast, J-INN yields superior UQ, attaining higher PICP values with comparable PINAW. For instance, in the \(\alpha=0.9\) results shown in Table~\ref{tab:cnn_picp_pinaw-90} and Table~\ref{tab:jnn_picp_pinaw_90}, INODE-1 under J-INN shows markedly improved coverage across most datasets.

Both the C-INN and J-INN learning strategies identify ILSTM-2 and INODE-2 as the best-performing models in general. However, INODE-2 becomes the superior approach due to its more consistent and robust behavior across both strategies.
Although J-INN generally achieves superior performance, C-INN remains competitive. This is further illustrated in Fig.~\ref{fig:snap}, which shows the normalized test performances across four datasets using 100 test sample snapshots for our best model, INODE-2, under both strategies.  Both learning strategies produce PIs with narrow widths that successfully cover the true data. Even in regions with sharp transitions, spikes, or noisy behavior, the proposed strategies perform reliably and provide clear, well-defined uncertainty bounds. However, it is worth noting that the ILSTM-2 model also demonstrates strong performance, which is comparable to INODE-2 and even surpasses it in some cases. 

To provide a clear picture, Tables~\ref{tab:ilstm2_inode2_comparison} and \ref{tab:hierarchical_comparison_fixed_95} summarize the results of the best-performing models, ILSTM-2 and INODE-2, under the C-INN and J-INN strategies, together with comparisons against Bayesian NN, MC Dropout, and Deep Ensemble counterparts (details on these models are presented in ~\ref{AppendixA}.) From the comparative performance analyses, we observe that:  
\begin{itemize}
\item In terms of RMSE, the C-INN approach yields better point prediction performance. This is because C-INN is a two-stage strategy that first trains a crisp NN focused solely on minimizing RMSE, resulting in superior point predictions.
\item For UQ performance, the J-INN approach generally achieves superior results; however, the C-INN approach remains competitive. Since J-INN learns point predictions and PIs simultaneously, it produces better calibrated intervals, but this joint optimization sometimes leads to a slight degradation in RMSE.

\item Compared with benchmark methods reported in Tables~\ref{tab:ilstm2_inode2_comparison} and \ref{tab:hierarchical_comparison_fixed_95}, Bayesian NNs generally do not achieve competitive performance in terms of either RMSE or UQ performance compared to both C-INN and J-INN strategies. Although they can reach the target coverage in some cases, especially for LSTM Networks, NODE networks frequently exhibit over-coverage behavior, indicating overly conservative uncertainty estimates. For the MC Dropout approach, the RMSE performance is consistently inferior to both C-INN and J-INN models. Similar to Bayesian NNs, MC Dropout also tends to produce over-coverage in both LSTM- and NODE-based implementations. In the case of Deep Ensembles, improved RMSE performance is observed in certain datasets (e.g., the Hair Dryer dataset for NODE Networks); however, this improvement is often accompanied by over-coverage, indicating a trade-off between accuracy and interval precision. Although Deep Ensembles and MC Dropout exhibit lower variance, they perform worse in terms of both RMSE and UQ than the proposed C-INN and J-INN.

\end{itemize}

To sum up, the presented results illustrate a trade-off between the C-INN and J-INN approaches in terms of their strengths: C-INN excels in point prediction as measured by RMSE, while J-INN provides better UQ performance, as reflected by better PICP values. If one prioritizes better point predictions, C-INN is preferable; however, for improved coverage probabilities with a modest acceptance of slightly worse RMSE, J-INN is recommended.

\begin{table}[ht!]
\caption{Comparison of Model Performance at 90\% Coverage across Datasets.}
\label{tab:ilstm2_inode2_comparison}
\centering
\begin{adjustbox}{max width=\textwidth}
\footnotesize
\renewcommand{\arraystretch}{1.2}
\setlength{\tabcolsep}{4pt}
\begin{threeparttable}

\begin{tabular}{|l|l|l|c|c|c|c|}
\hline
\textbf{Approach} & \textbf{Base Model} & \textbf{Metric} & \textbf{Robot Arm} & \textbf{Hair Dryer} & \textbf{MR-Damper} & \textbf{Heat Exchanger} \\
\hline

\multirow{8}{*}{\textbf{C-INN}} 
& \multirow{4}{*}{\textbf{LSTM-2}} 
& RMSE  & 19.39\((\pm3.85)\) & 9.88\((\pm0.42)\) & 617.40\((\pm36.73)\) & 54.86\((\pm3.35)\) \\
& & PICP  & 89.75\((\pm4.87)\) & 90.32\((\pm6.55)\) & 89.79\((\pm0.89)\) & 89.51\((\pm3.24)\) \\
& & PINAW & 42.31\((\pm9.33)\) & 14.27\((\pm2.41)\) & 12.59\((\pm0.72)\) & 23.69\((\pm2.30)\) \\
& & CWC   & 1.28\((\pm1.62)\) & 0.53\((\pm0.67)\) & 0.21\((\pm0.09)\) & 0.52\((\pm0.20)\) \\
\cline{2-7}
& \multirow{4}{*}{\textbf{NODE-2}} 
& RMSE  & 18.57\((\pm1.82)\) & 10.27\((\pm1.72)\) & 686.86\((\pm21.36)\) & 59.06\((\pm4.13)\) \\
& & PICP  & 87.95\((\pm2.13)\) & 88.68\((\pm2.45)\) & 90.73\((\pm1.21)\) & 85.39\((\pm2.73)\) \\
& & PINAW & 39.86\((\pm3.88)\) & 12.68\((\pm1.45)\) & 15.70\((\pm1.25)\) & 19.95\((\pm2.10)\) \\
& & CWC   & 1.08\((\pm0.42)\) & 0.32\((\pm0.13)\) & 0.22\((\pm0.08)\) & 0.95\((\pm0.42)\) \\
\hline

\multirow{8}{*}{\textbf{J-INN}} 
& \multirow{4}{*}{\textbf{LSTM-2}} 
& RMSE  & 18.35\((\pm3.22)\) & 9.72\((\pm0.96)\) & 632.46\((\pm43.40)\) & 56.38\((\pm3.62)\) \\
& & PICP  & 91.04\((\pm4.18)\) & 92.08\((\pm3.25)\) & 89.28\((\pm2.34)\) & 90.59\((\pm2.03)\) \\
& & PINAW & 43.07\((\pm8.09)\) & 12.86\((\pm1.42)\) & 11.90\((\pm0.75)\) & 21.31\((\pm1.81)\) \\
& & CWC   & 0.86\((\pm0.79)\) & 0.18\((\pm0.09)\) & 0.25\((\pm0.16)\) & 0.31\((\pm0.15)\) \\
\cline{2-7}
& \multirow{4}{*}{\textbf{NODE-2}} 
& RMSE  & 19.82\((\pm1.09)\) & 10.50\((\pm0.94)\) & 689.90\((\pm28.88)\) & 64.58\((\pm12.02)\) \\
& & PICP  & 89.20\((\pm1.84)\) & 92.24\((\pm4.14)\) & 90.43\((\pm2.41)\) & 89.20\((\pm3.77)\) \\
& & PINAW & 43.04\((\pm2.18)\) & 13.14\((\pm1.30)\) & 18.55\((\pm1.60)\) & 22.63\((\pm4.45)\) \\
& & CWC   & 0.91\((\pm0.35)\) & 0.22\((\pm0.15)\) & 0.29\((\pm0.17)\) & 0.52\((\pm0.29)\) \\
\hline

\multirow{8}{*}{\textbf{Bayesian NN}} 
& \multirow{4}{*}{\textbf{LSTM}} 
& RMSE  & 19.63\((\pm3.25)\) & 12.64\((\pm0.58)\) & 706.31\((\pm102.21)\) & 62.69\((\pm2.92)\) \\
& & PICP  & 87.28\((\pm3.75)\) & 90.22\((\pm3.37)\) & 90.20\((\pm2.33)\) & 88.27\((\pm5.06)\) \\
& & PINAW & 40.39\((\pm4.24)\) & 14.90\((\pm1.04)\) & 11.42\((\pm1.40)\) & 23.05\((\pm4.81)\) \\
& & CWC   & 1.59\((\pm1.39)\) & 0.31\((\pm0.12)\) & 0.19\((\pm0.08)\) & 0.79\((\pm0.56)\) \\

\cline{2-7}
& \multirow{4}{*}{\textbf{NODE}} 
& RMSE  & 19.60\((\pm1.15)\) & 13.47\((\pm1.51)\) & 1355.30\((\pm571.02)\) & 70.14\((\pm2.14)\) \\
& & PICP  & 92.45\((\pm1.79)\) & 92.61\((\pm4.44)\) & 94.63\((\pm2.86)\) & 93.14\((\pm1.48)\) \\
& & PINAW & 46.45\((\pm3.41)\) & 17.57\((\pm2.42)\) & 32.74\((\pm13.49)\) & 29.65\((\pm1.74)\) \\
& & CWC   & 0.51\((\pm0.13)\) & 0.31\((\pm0.28)\) & 0.33\((\pm0.13)\) & 0.30\((\pm0.02)\) \\
\hline

\multirow{8}{*}{\textbf{MC Dropout}} 
& \multirow{4}{*}{\textbf{LSTM}} 
& RMSE  & 19.15\((\pm3.68)\) & 10.81\((\pm0.57)\) & 676.46\((\pm46.72)\) & 62.52\((\pm10.07)\) \\
& & PICP  & 90.86\((\pm3.19)\) & 98.76\((\pm0.96)\) & 95.75\((\pm1.35)\) & 90.72\((\pm1.29)\) \\
& & PINAW & 44.31\((\pm9.21)\) & 20.49\((\pm1.88)\) & 21.06\((\pm1.51)\) & 23.37\((\pm3.85)\) \\
& & CWC   & 0.75\((\pm0.45)\) & 0.20\((\pm0.02)\) & 0.21\((\pm0.02)\) & 0.33\((\pm0.17)\) \\
\cline{2-7}
& \multirow{4}{*}{\textbf{NODE}} 
& RMSE  & 16.66\((\pm0.75)\) & 9.31\((\pm0.62)\) & 810.60\((\pm144.06)\) & 56.37\((\pm2.60)\) \\
& & PICP  & 96.18\((\pm0.73)\) & 98.46\((\pm0.74)\) & 93.46\((\pm1.85)\) & 93.06\((\pm2.30)\) \\
& & PINAW & 49.54\((\pm1.41)\) & 16.00\((\pm0.78)\) & 18.29\((\pm1.69)\) & 24.04\((\pm2.07)\) \\
& & CWC   & 0.50\((\pm0.01)\) & 0.16\((\pm0.01)\) & 0.21\((\pm0.07)\) & 0.27\((\pm0.08)\) \\
\hline

\multirow{8}{*}{\textbf{Deep Ensemble}} 
& \multirow{4}{*}{\textbf{LSTM}} 
& RMSE  & 20.81\((\pm1.92)\) & 9.28\((\pm0.33)\) & 512.77\((\pm21.33)\) & 58.40\((\pm4.06)\) \\
& & PICP  & 96.03\((\pm1.44)\) & 92.91\((\pm2.02)\) & 91.27\((\pm1.23)\) & 91.04\((\pm1.78)\) \\
& & PINAW & 55.13\((\pm4.00)\) & 12.04\((\pm0.70)\) & 9.87\((\pm0.53)\) & 22.10\((\pm2.30)\) \\
& & CWC   & 0.55\((\pm0.04)\) & 0.13\((\pm0.03)\) & 0.11\((\pm0.04)\) & 0.28\((\pm0.11)\) \\
\cline{2-7}
& \multirow{4}{*}{\textbf{NODE}} 
& RMSE  & 16.57\((\pm0.43)\) & 9.41\((\pm0.43)\) & 702.72\((\pm135.69)\) & 58.57\((\pm3.26)\) \\
& & PICP  & 94.09\((\pm1.79)\) & 91.80\((\pm1.68)\) & 94.57\((\pm1.44)\) & 92.66\((\pm2.24)\) \\
& & PINAW & 43.74\((\pm2.41)\) & 12.24\((\pm0.52)\) & 16.22\((\pm3.35)\) & 24.94\((\pm2.59)\) \\
& & CWC   & 0.44\((\pm0.02)\) & 0.14\((\pm0.05)\) & 0.16\((\pm0.03)\) & 0.31\((\pm0.11)\) \\
\hline

\end{tabular}
\begin{tablenotes}
\footnotesize
\item Note: RMSE and PINAW values are scaled by 100.
\end{tablenotes}
\end{threeparttable}
\end{adjustbox}
\end{table}

\newcolumntype{C}{>{\centering\arraybackslash}p{2.5cm}}

\begin{table}[ht!]
\caption{Comparison of Model Performance at 95\% Coverage across Datasets.}
\label{tab:hierarchical_comparison_fixed_95}
\centering
\begin{adjustbox}{max width=\textwidth}
\footnotesize
\renewcommand{\arraystretch}{1.2}
\setlength{\tabcolsep}{3pt} 
\begin{threeparttable}

\begin{tabular}{|l|l|l|C|C|C|C|}
\hline
\textbf{Approach} & \textbf{Base Model} & \textbf{Metric} & \textbf{Robot Arm} & \textbf{Hair Dryer} & \textbf{MR-Damper} & \textbf{Heat Exchanger} \\
\hline

\multirow{8}{*}{\textbf{C-INN}} 
& \multirow{4}{*}{\textbf{LSTM-2}} 
& RMSE  & 19.39\((\pm3.85)\) & 9.88\((\pm0.42)\) & 617.40\((\pm36.73)\) & 54.86\((\pm3.35)\) \\
& & PICP  & 95.62\((\pm1.90)\) & 95.83\((\pm3.27)\) & 95.77\((\pm0.75)\) & 96.18\((\pm1.38)\) \\
& & PINAW & 52.23\((\pm9.61)\) & 17.92\((\pm2.43)\) & 17.32\((\pm1.23)\) & 30.47\((\pm2.72)\) \\
& & CWC   & 0.79\((\pm0.46)\) & 0.30\((\pm0.18)\) & 0.21\((\pm0.08)\) & 0.37\((\pm0.11)\) \\
\cline{2-7}
& \multirow{4}{*}{\textbf{NODE-2}} 
& RMSE  & 18.57\((\pm1.82)\) & 10.27\((\pm1.72)\) & 686.86\((\pm21.36)\) & 59.06\((\pm4.13)\) \\
& & PICP  & 94.81\((\pm1.35)\) & 93.93\((\pm2.49)\) & 94.55\((\pm1.76)\) & 92.40\((\pm1.92)\) \\
& & PINAW & 48.23\((\pm3.73)\) & 15.32\((\pm2.38)\) & 19.27\((\pm1.46)\) & 24.82\((\pm2.36)\) \\
& & CWC   & 0.78\((\pm0.38)\) & 0.35\((\pm0.17)\) & 0.37\((\pm0.14)\) & 0.76\((\pm0.21)\) \\
\hline

\multirow{8}{*}{\textbf{J-INN}} 
& \multirow{4}{*}{\textbf{LSTM-2}} 
& RMSE  & 20.15\((\pm3.36)\) & 9.69\((\pm0.87)\) & 634.59\((\pm38.33)\) & 56.21\((\pm4.28)\) \\
& & PICP  & 94.54\((\pm4.38)\) & 95.37\((\pm2.57)\) & 94.85\((\pm1.42)\) & 96.72\((\pm1.25)\) \\
& & PINAW & 56.01\((\pm6.23)\) & 15.09\((\pm1.72)\) & 15.78\((\pm0.86)\) & 28.23\((\pm3.01)\) \\
& & CWC   & 1.61\((\pm1.86)\) & 0.28\((\pm0.11)\) & 0.27\((\pm0.12)\) & 0.34\((\pm0.10)\) \\
\cline{2-7}
& \multirow{4}{*}{\textbf{NODE-2}} 
& RMSE  & 19.82\((\pm0.80)\) & 10.18\((\pm1.19)\) & 692.71\((\pm37.01)\) & 65.41\((\pm13.23)\) \\
& & PICP  & 95.68\((\pm1.60)\) & 97.21\((\pm1.98)\) & 95.02\((\pm0.96)\) & 93.93\((\pm2.56)\) \\
& & PINAW & 53.15\((\pm3.94)\) & 16.76\((\pm2.35)\) & 23.35\((\pm1.78)\) & 27.62\((\pm7.70)\) \\
& & CWC   & 0.79\((\pm0.31)\) & 0.20\((\pm0.07)\) & 0.40\((\pm0.14)\) & 0.60\((\pm0.21)\) \\
\hline

\multirow{8}{*}{\textbf{Bayesian NN}} 
& \multirow{4}{*}{\textbf{LSTM}} 
& RMSE  & 19.63\((\pm3.25)\) & 12.64\((\pm0.58)\) & 706.31\((\pm102.21)\) & 62.69\((\pm2.92)\) \\
& & PICP  & 92.72\((\pm3.43)\) & 95.53\((\pm2.09)\) & 93.42\((\pm2.29)\) & 93.41\((\pm3.09)\) \\
& & PINAW & 48.12\((\pm5.05)\) & 17.75\((\pm1.24)\) & 13.61\((\pm1.66)\) & 27.46\((\pm5.74)\) \\
& & CWC   & 1.54\((\pm1.38)\) & 0.31\((\pm0.10)\) & 0.34\((\pm0.13)\) & 0.69\((\pm0.30)\) \\
\cline{2-7}
& \multirow{4}{*}{\textbf{NODE}} 
& RMSE  & 19.60\((\pm1.15)\) & 13.47\((\pm1.51)\) & 1355.30\((\pm571.02)\) & 70.14\((\pm2.14)\) \\
& & PICP  & 96.65\((\pm1.48)\) & 96.55\((\pm2.98)\) & 98.15\((\pm1.39)\) & 96.84\((\pm0.89)\) \\
& & PINAW & 55.34\((\pm4.06)\) & 20.94\((\pm2.88)\) & 39.00\((\pm16.07)\) & 35.33\((\pm2.07)\) \\
& & CWC   & 0.67\((\pm0.21)\) & 0.31\((\pm0.21)\) & 0.39\((\pm0.16)\) & 0.35\((\pm0.02)\) \\
\hline

\multirow{8}{*}{\textbf{MC Dropout}} 
& \multirow{4}{*}{\textbf{LSTM}} 
& RMSE  & 19.15\((\pm3.68)\) & 10.81\((\pm0.57)\) & 676.46\((\pm46.72)\) & 62.52\((\pm10.07)\) \\
& & PICP  & 95.07\((\pm2.13)\) & 99.68\((\pm0.35)\) & 97.44\((\pm1.10)\) & 95.41\((\pm0.85)\) \\
& & PINAW & 52.79\((\pm10.97)\) & 24.41\((\pm2.24)\) & 25.09\((\pm1.80)\) & 27.85\((\pm4.58)\) \\
& & CWC   & 0.94\((\pm0.36)\) & 0.24\((\pm0.02)\) & 0.25\((\pm0.02)\) & 0.37\((\pm0.13)\) \\
\cline{2-7}
& \multirow{4}{*}{\textbf{NODE}} 
& RMSE  & 16.66\((\pm0.75)\) & 9.31\((\pm0.62)\) & 810.60\((\pm144.06)\) & 56.37\((\pm2.60)\) \\
& & PICP  & 98.63\((\pm0.30)\) & 99.50\((\pm0.48)\) & 96.95\((\pm1.58)\) & 96.65\((\pm1.39)\) \\
& & PINAW & 59.03\((\pm1.69)\) & 19.07\((\pm0.93)\) & 21.79\((\pm2.01)\) & 28.64\((\pm2.47)\) \\
& & CWC   & 0.59\((\pm0.02)\) & 0.19\((\pm0.01)\) & 0.26\((\pm0.12)\) & 0.35\((\pm0.11)\) \\
\hline

\multirow{8}{*}{\textbf{Deep Ensemble}} 
& \multirow{4}{*}{\textbf{LSTM}} 
& RMSE  & 20.81\((\pm1.92)\) & 9.28\((\pm0.33)\) & 512.77\((\pm21.33)\) & 58.40\((\pm4.06)\) \\
& & PICP  & 98.67\((\pm0.76)\) & 96.47\((\pm1.27)\) & 94.25\((\pm1.14)\) & 95.76\((\pm1.04)\) \\
& & PINAW & 65.69\((\pm4.76)\) & 14.34\((\pm0.83)\) & 11.76\((\pm0.64)\) & 26.33\((\pm2.74)\) \\
& & CWC   & 0.66\((\pm0.05)\) & 0.17\((\pm0.06)\) & 0.23\((\pm0.08)\) & 0.32\((\pm0.11)\) \\
\cline{2-7}
& \multirow{4}{*}{\textbf{NODE}} 
& RMSE  & 16.57\((\pm0.43)\) & 9.41\((\pm0.43)\) & 702.72\((\pm135.69)\) & 58.57\((\pm3.26)\) \\
& & PICP  & 97.55\((\pm1.26)\) & 96.23\((\pm1.36)\) & 97.33\((\pm0.90)\) & 96.40\((\pm1.31)\) \\
& & PINAW & 52.12\((\pm2.87)\) & 14.58\((\pm0.62)\) & 19.32\((\pm3.99)\) & 29.71\((\pm3.09)\) \\
& & CWC   & 0.52\((\pm0.03)\) & 0.17\((\pm0.07)\) & 0.19\((\pm0.04)\) & 0.34\((\pm0.10)\) \\
\hline

\end{tabular}
\begin{tablenotes}
\footnotesize
\item Note: RMSE and PINAW values are scaled by 100.
\end{tablenotes}
\end{threeparttable}
\end{adjustbox}
\end{table}

\subsection{Analyzing the Uncertainty Representation} \label{elas}
A key theoretical feature of the INN framework is its treatment of parameters as intervals, enabling intrinsic representation of uncertainty. Unlike conventional NNs that rely on external mechanisms (e.g., Bayesian priors or ensembles), INNs embed uncertainty directly within the interval LPs~$\tilde{\theta}$. This joint formulation integrates point prediction accuracy and UQ, and thus necessitates a careful analysis of how uncertainty is distributed across the network. For this purpose, the concept of \textit{elasticity} was introduced in prior work \cite{ferah2025introducing} to quantify the level of uncertainty associated with an INN parameter. The elasticity of an interval LP set is defined as
\begin{equation}\label{uncertainity}
    \bar{r}=\frac{\|\overline{\theta}-\underline{\theta}\|}{\|\theta\|}
\end{equation} 
 If $\bar{r}$ is close to 0, the uncertainty in the LP is minimal, indicating that the corresponding weight or bias is well-defined and does not contribute to uncertainty. On the other hand, if $\bar{r}$ is bigger than 1, the uncertainty is maximal, indicating that the parameter is highly uncertain. 

\begin{figure}[h!]
    \centering
    \setlength{\tabcolsep}{4pt} 
    \renewcommand{\arraystretch}{0}

    \subfloat[INODE-1]{
        \begin{tabular}{@{}c@{}}
            \includegraphics[width=0.48\textwidth]{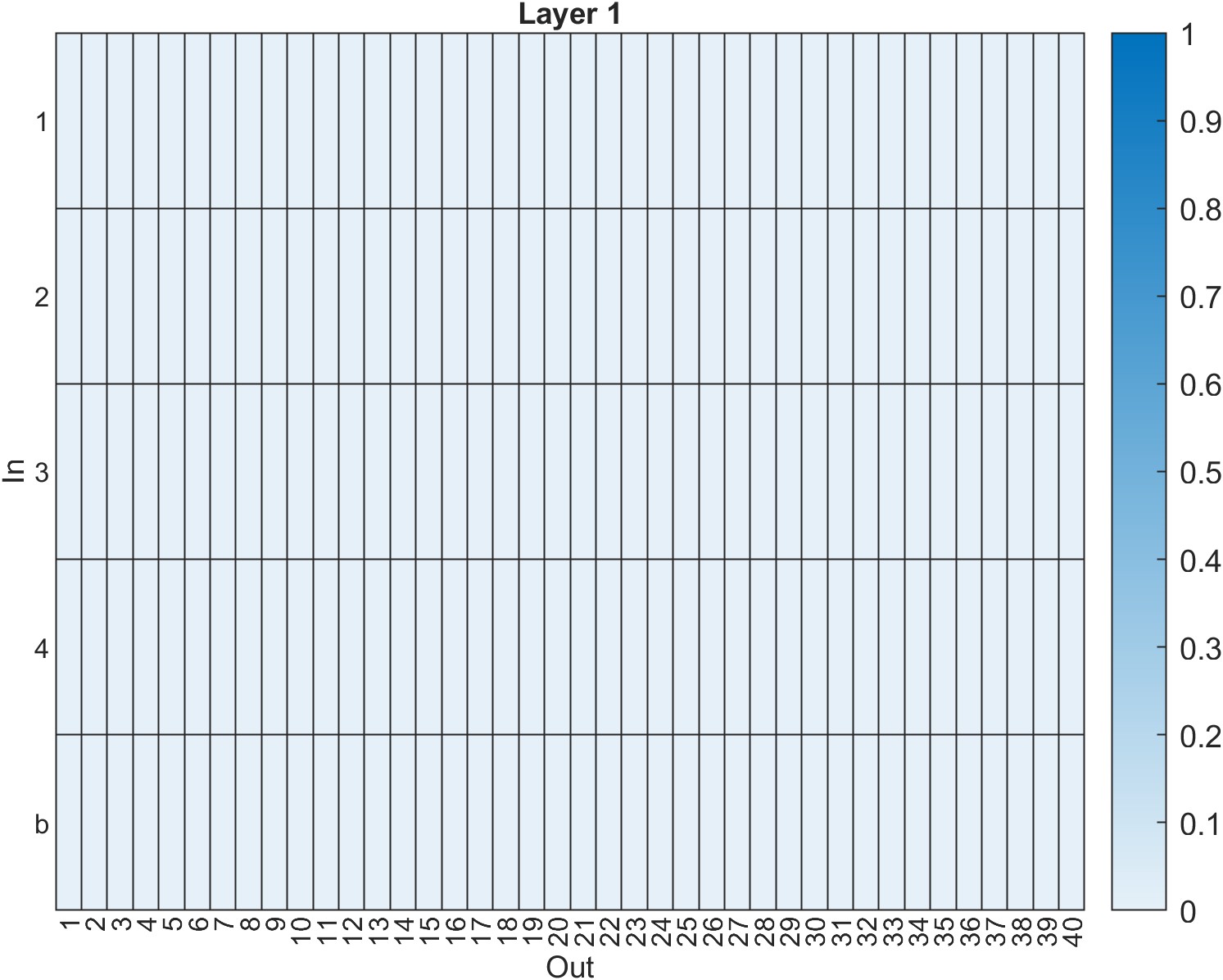} \\
            \includegraphics[width=0.48\textwidth]{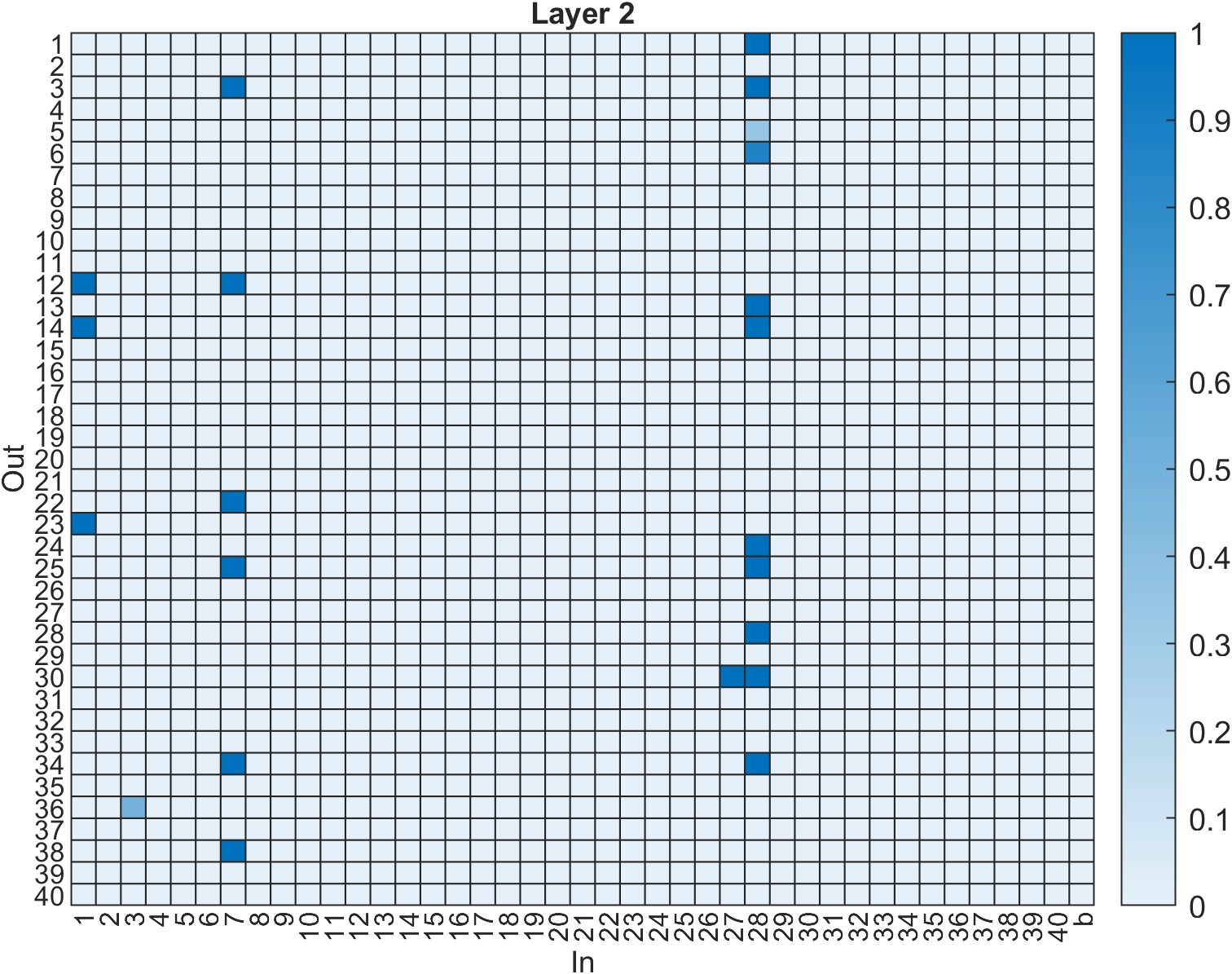} \\
            \includegraphics[width=0.48\textwidth]{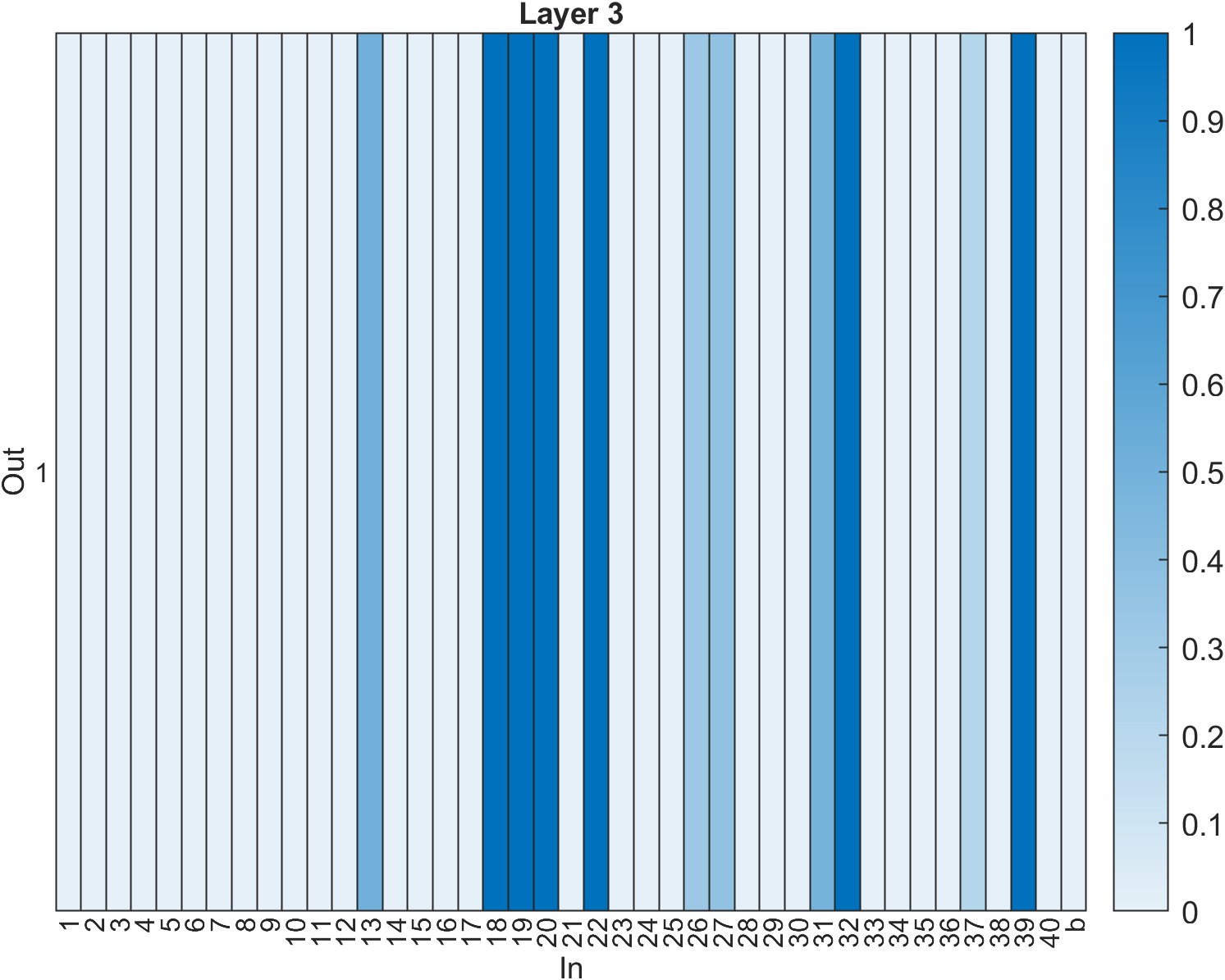}
        \end{tabular}
    }
    \hfill
    \subfloat[INODE-2]{
        \begin{tabular}{@{}c@{}}
            \includegraphics[width=0.48\textwidth]{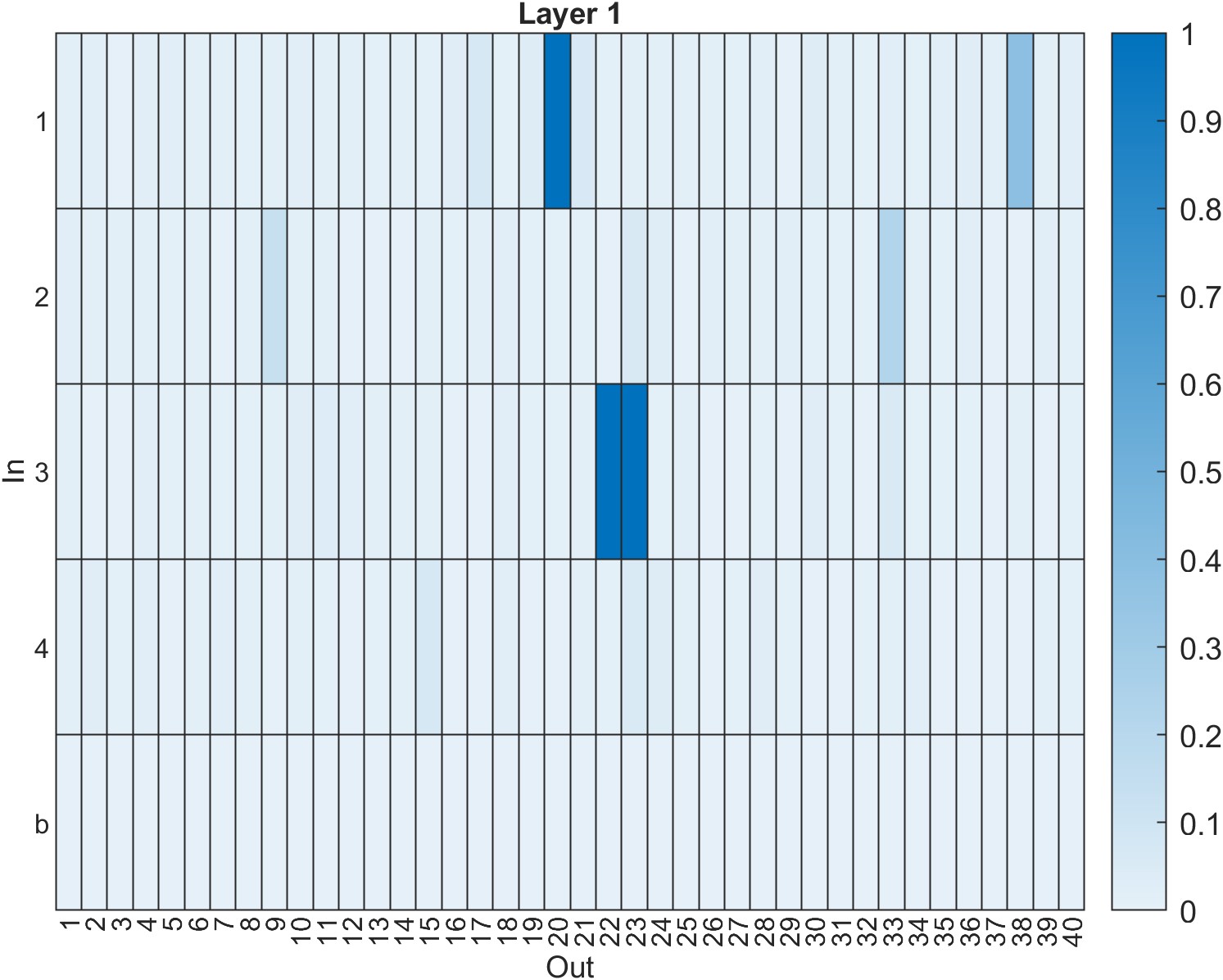} \\
            \includegraphics[width=0.48\textwidth]{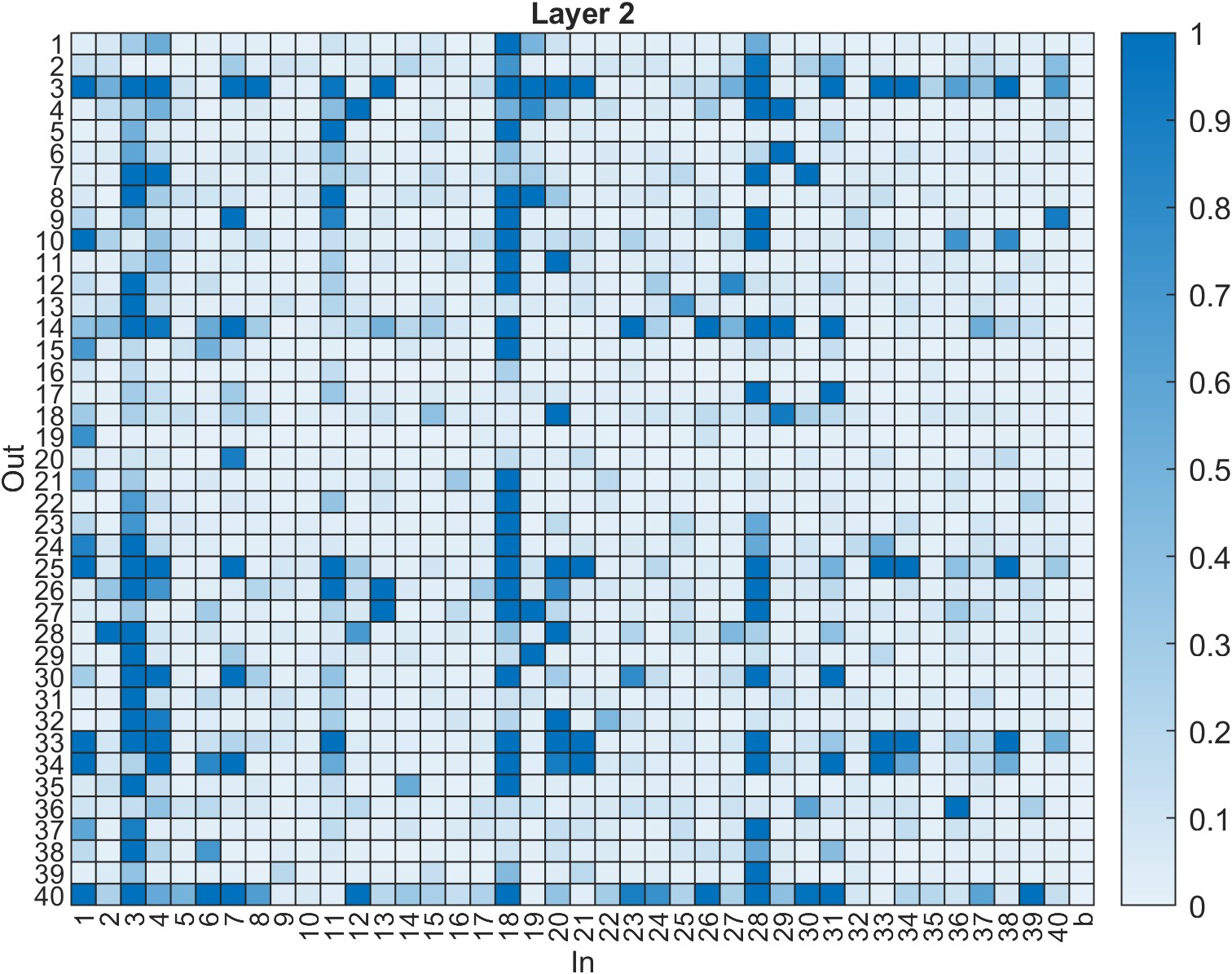} \\
            \includegraphics[width=0.48\textwidth]{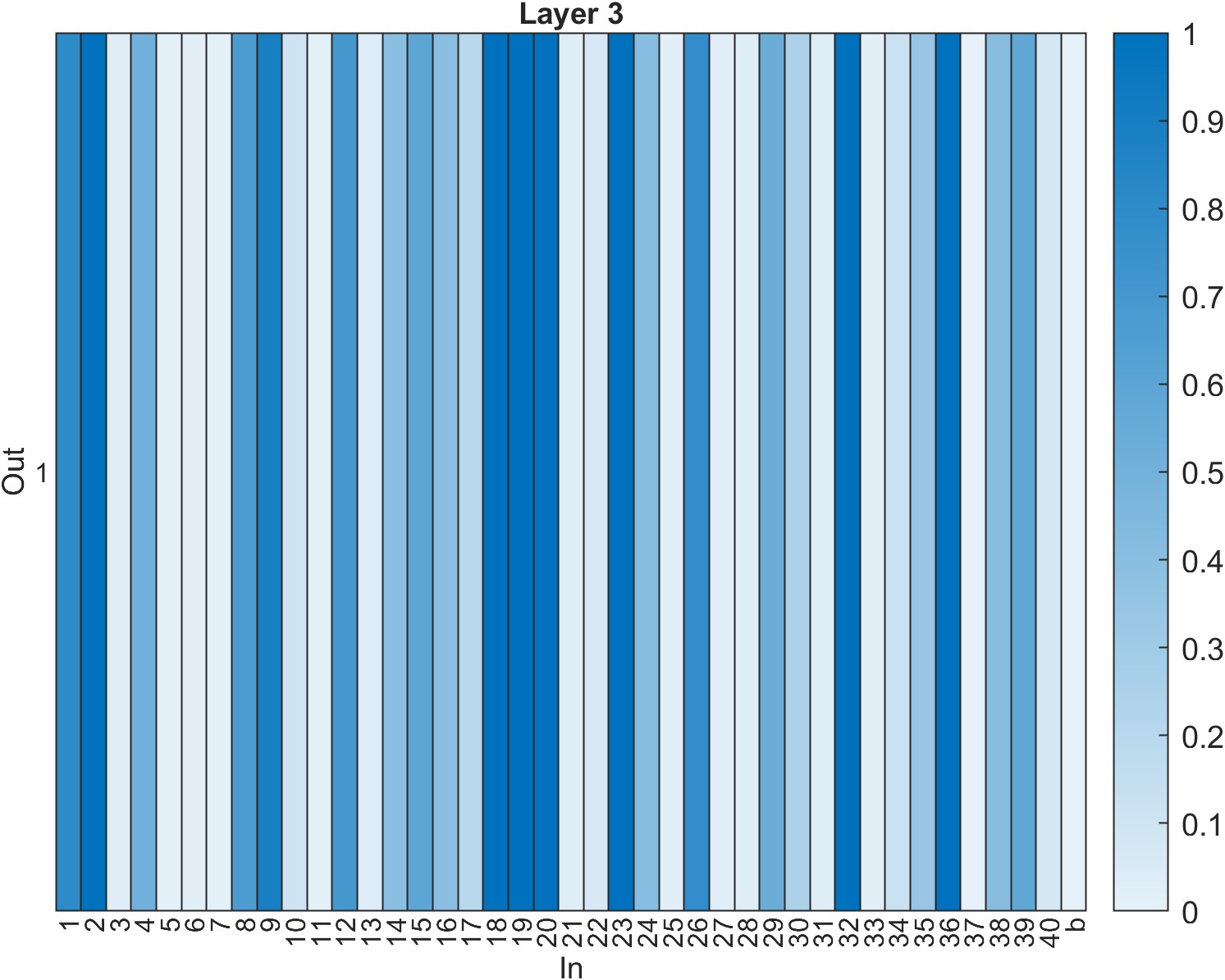}
        \end{tabular}
    }

    \caption{Heatmaps of LPs resulting from the MR-Damper dataset using the J-INN strategy. The regression input, defined in Eq.~\eqref{NNinput}, is $x(k) = [u(k), u(k-1), u(k-2), y(k-1)]$. In layer 1, the LPs corresponding to these inputs map to channel inputs 2, 3, 4, and 1, respectively.
}
    \label{heatmap}
\end{figure}

To facilitate a visual interpretation of the elasticity, Fig.~\ref{heatmap} presents the $\bar{r}$ values for each weight and bias as a heatmap, focusing on the INODE-1 and INODE-2 models trained on the MR-Damper dataset with $\text{seed} = 1$ and $\alpha=0.9$ for the J-INN strategy. The INODE models were selected for this analysis due to their relatively small number of parameters compared to their ILSTM counterparts, which simplifies visual examination. The J-INN strategy was chosen because its performance surpasses that of the C-INN counterpart for the given models in terms of PICP, with INODE-1 achieving 90.65 and INODE-2 achieving 90.56. Furthermore, analyzing both INODE-1 and INODE-2 enables the investigation of the influence of the activation function on the uncertainty distribution across all parameters in the trained models. It can be observed from Fig.~\ref{heatmap} that INODE-1 exhibits a sparser uncertainty pattern compared to INODE-2. This suggests that using $\sigma^{\mathrm{ReLU}}$ instead of $\sigma^{\mathrm{abs}}$ may yield a simpler and more interpretable uncertainty distribution in the parameters. However, this does not necessarily imply better performance, as shown in Section~\ref{joint}, where INODE-2 achieves the best results. Although INODE-2 presents a denser uncertainty distribution across model parameters, it still allows for a clear analysis of uncertainty behavior along the channels.

The heatmap representation of elasticity provides a powerful means of analyzing uncertainty within INNs; however, as network complexity increases, particularly with larger neuron sizes, it becomes difficult to disentangle the contribution of each input to the overall UQ. To address this limitation, we propose a projected version of elasticity, termed \emph{channel-wise elasticity}. We define the channel-wise elasticity as follows:
\begin{equation}
    \bar{R}_{i}(n_{in}) = \sum_{n_{out}=1}^{N^{out}_i} \bar{r}_{i}(n_{in},n_{out}),
\end{equation}
where $\bar{r}_{i}(m,n)$ denotes the elasticity associated with the connection from input channel $m$ to output channel $n$ in layer $i$. Here, $N^{out}_i$ is the number of output channels in layer $i$, and $n_{in}$ indexes the input channels. This formulation projects the full elasticity matrix onto the input channels, providing a compact measure of each channel’s contribution to the uncertainty in layer $i$.

\begin{figure}[t]
    \centering
    
    \begin{minipage}[t]{0.48\textwidth}
        \centering
        \includegraphics[width=\linewidth]{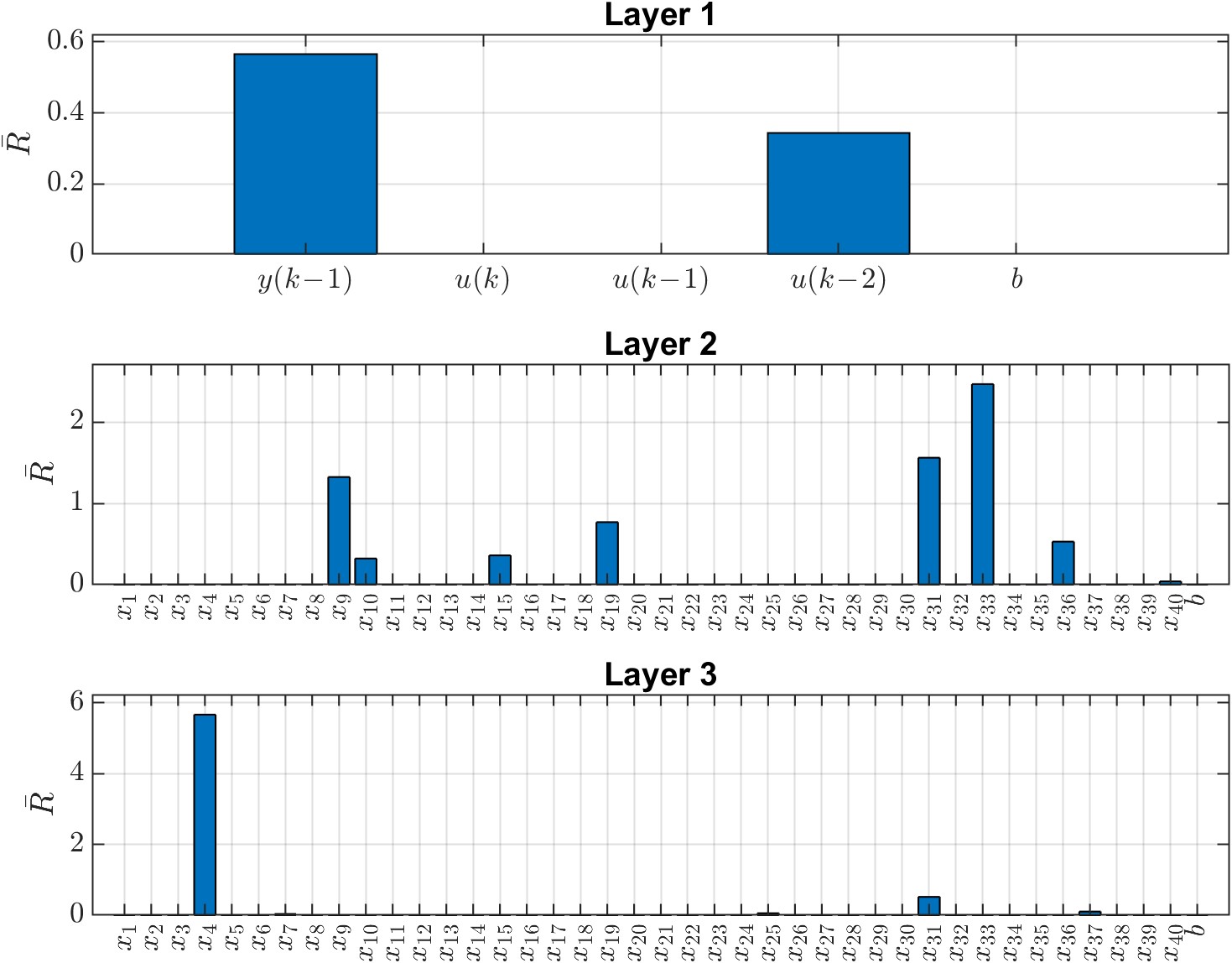}\par\vspace{0.4cm}
        \vspace{0.2cm}
        {\footnotesize (a) INODE-1: C-INN Strategy}
    \end{minipage}%
    \hfill
    \begin{minipage}[t]{0.48\textwidth}
        \centering
        \includegraphics[width=\linewidth]{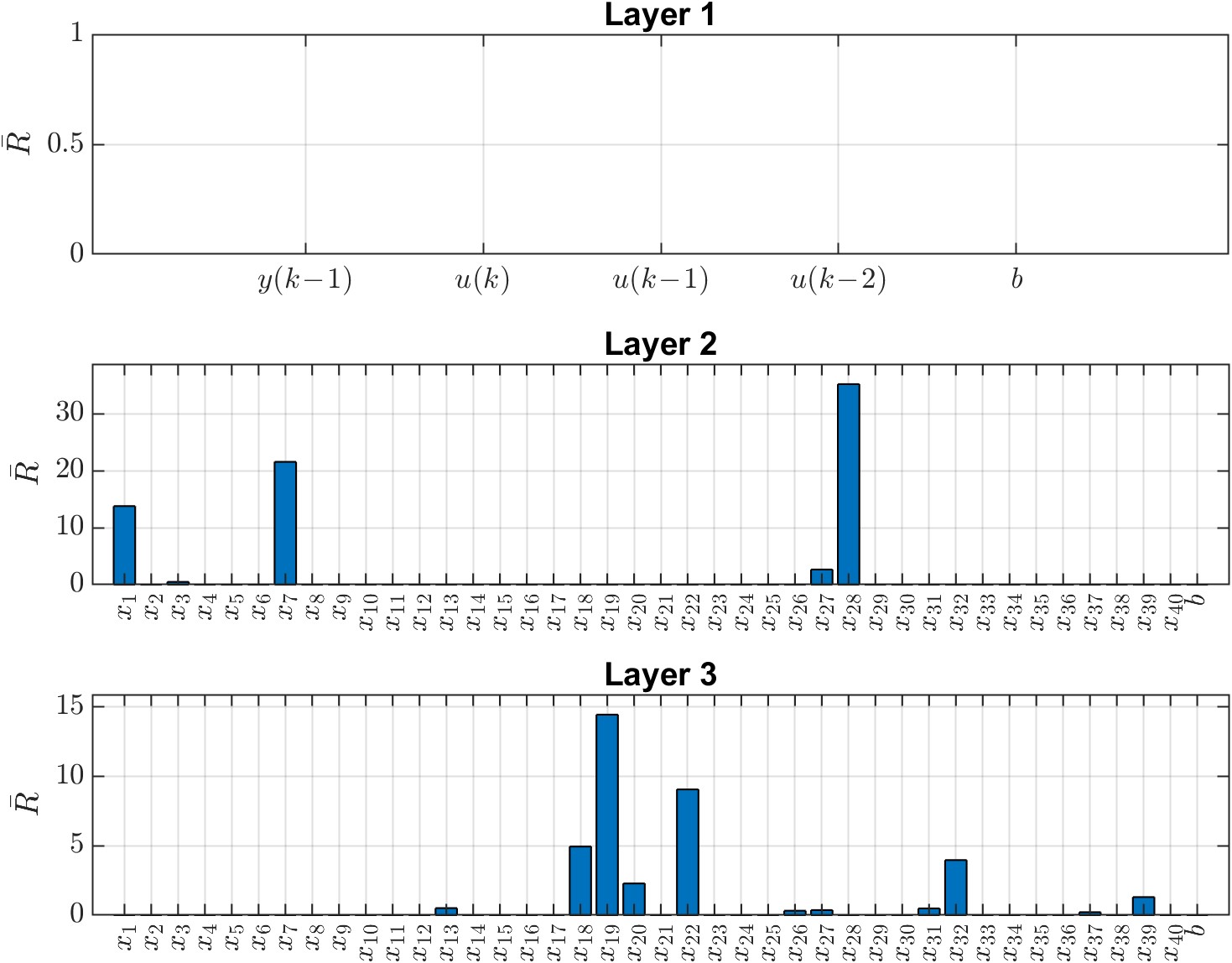}\par\vspace{0.4cm}
        \vspace{0.2cm}
        {\footnotesize (b) INODE-1: J-INN Strategy}
    \end{minipage}
    \vspace{0.6cm}
    
    \begin{minipage}[t]{0.48\textwidth}
        \centering
        \includegraphics[width=\linewidth]{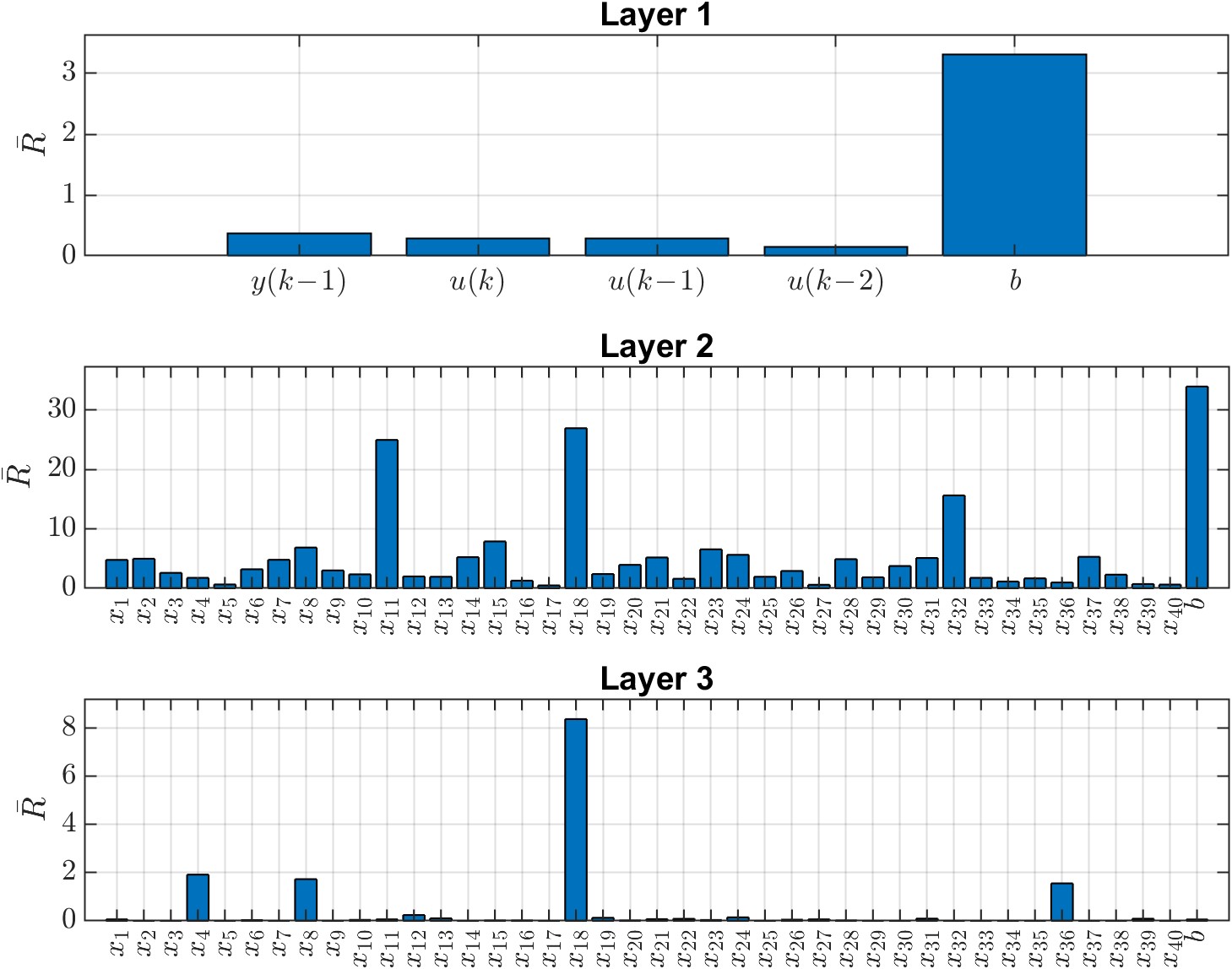}\par\vspace{0.4cm}
        \vspace{0.2cm}
            {\footnotesize (c) INODE-2: C-INN Strategy}
    \end{minipage}%
    \hfill
    \begin{minipage}[t]{0.48\textwidth}
        \centering
        \includegraphics[width=\linewidth]{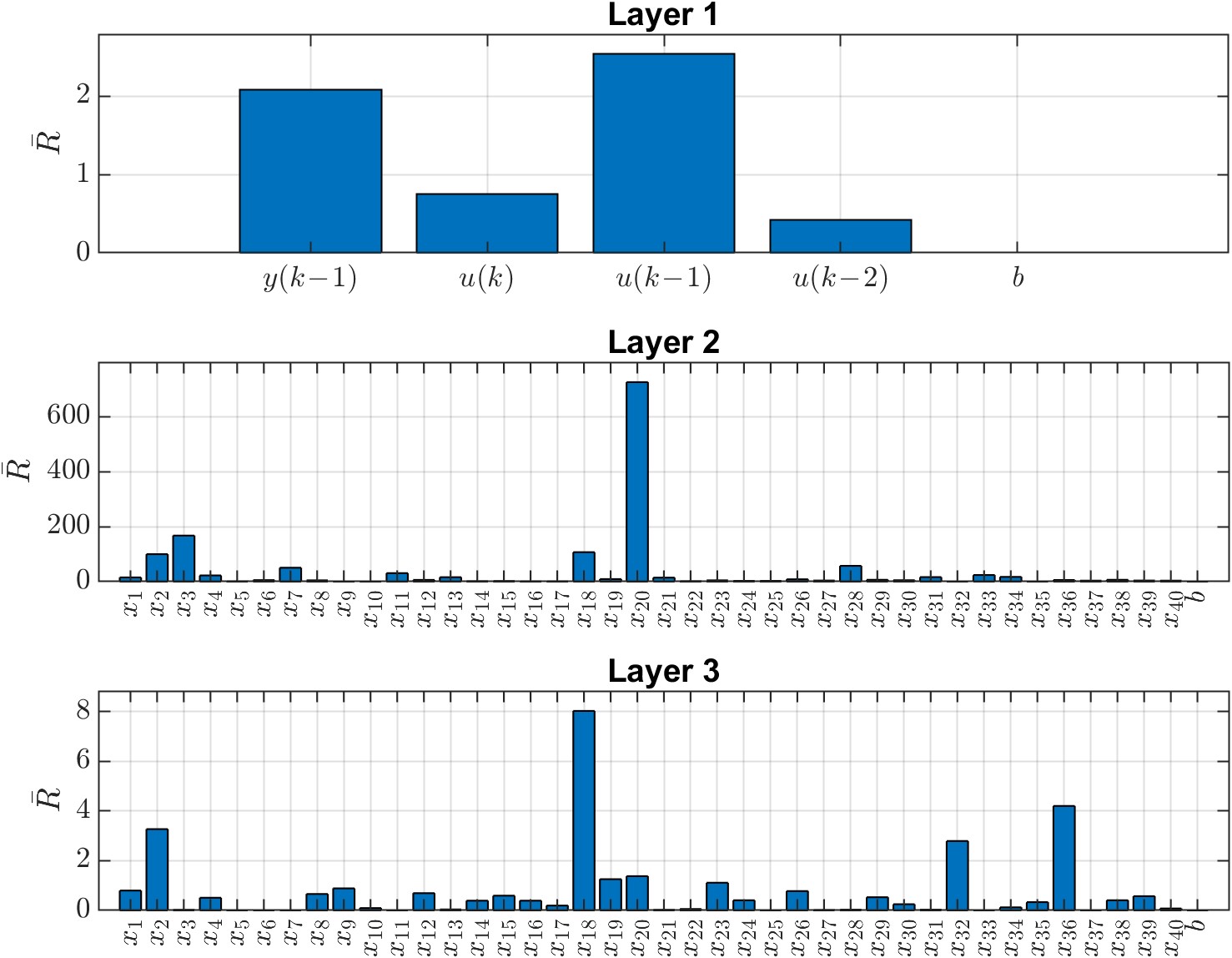}\par\vspace{0.4cm}
        \vspace{0.2cm}
        {\footnotesize (d) INODE-2: J-INN Strategy}
    \end{minipage}
    
    \caption{Channel-wise elasticity of models obtained for the MR-Damper Dataset}
    \label{heatmap_r}
\end{figure}

Fig.~\ref{heatmap_r} presents the channel-wise elasticity for the INODE-1 and INODE-2 models under both C-INN and J-INN training strategies. Overall, INODE-1 exhibits uncertainty concentrated in specific inputs and primarily in the last layer, whereas INODE-2 shows a broader distribution of elasticity across multiple inputs and layers, regardless of the training strategy. Focusing on the best performing INODE-2, the C-INN strategy concentrates elasticity mainly on the constant bias term, representing uncertainty as largely constant noise, while the J-INN strategy distributes uncertainty more evenly across regression inputs, with $y(k-1)$ and $u(k-1)$ exhibiting the largest contributions. In general, C-INN tends to localize uncertainty to specific weights or biases, particularly in the first and final layers, whereas J-INN encourages a more distributed representation of epistemic uncertainty, reflecting the differences in how these strategies propagate and encode uncertainty through the network.

\section{Conclusion and Future Work}\label{conc}

This paper introduced a systematic framework for constructing and training INNs for uncertainty-aware SysID. By extending crisp NN architectures into their interval counterparts, we developed the ILSTM and INODE, enabling predictive modeling that naturally incorporates uncertainty propagation without relying on probabilistic assumptions. To address the challenge of training INNs, we proposed two complementary learning strategies: C-INN, a two-stage learning strategy, and J-INN, a one-stage learning strategy. Both frameworks employ tailored UQ-specific loss functions and parameterization tricks to ensure robust and reliable performance. We provided comprehensive mathematical foundations and detailed implementation guidelines to support reproducibility and practical deployment.

Comparative evaluations across four benchmark SysID datasets demonstrated that both strategies are effective, yet they offer distinct advantages: C-INN excels in point prediction, while J-INN produces tighter and more reliable PIs. Beyond performance evaluation, we introduced a channel-wise elasticity analysis to provide an interpretable view of how uncertainty is distributed across network parameters and input features. This analysis revealed fundamental differences in how the two training strategies propagate and encode uncertainty, highlighting the trade-offs between localized versus distributed uncertainty representation.

Future research will further advance the concept of elasticity as an explanatory tool to strengthen its connection to system dynamics and input relevance. In parallel, extending INNs to physics-informed formulations and scaling the framework to large-scale, real-time applications remain key directions for promoting the practical adoption of uncertainty-aware SysID in safety-critical control system applications.

\section*{Acknowledgment}
This work was supported by the Scientific Research Projects Commission of Istanbul Technical University under the Research Universities Development Program (Project No. MGA-2025-46487) and by the Scientific and Technological Research Council of Türkiye (TÜBİTAK) (Project No. 125E231).

\section*{Declaration of generative AI and AI-assisted technologies in the writing process}
During the preparation of this work, the authors used ChatGPT in order to refine the grammar and enhance the English language expressions. After using ChatGPT, the authors reviewed and edited the content as needed and take full responsibility for the content of the publication

\appendix
\section{Baseline Probabilistic NN Models} \label{AppendixA}
The performances of the proposed INN models are benchmarked against commonly used probabilistic NN approaches as described within this section. For all benchmark models, we employ the negative log-likelihood (NLL) loss. Let the NN prediction $Y(m,k)$ parameterize a Gaussian distribution with mean $\mu(m,k)$ and variance $\sigma^2(m,k)$. The NLL loss is defined as:
\begin{equation}\label{nll_loss}
L_{\text{NLL}} = \frac{1}{BN} \sum_{m=1}^{B}\sum_{k=1}^{N} \left[ \frac{1}{2} \log \sigma^2(m,k) + \frac{\big(\hat{Y}(m,k) - \mu(m,k)\big)^2}{2\sigma^2(m,k)} \right]
\end{equation}
In both LSTM and NODE networks defined with probabilistic NNs, the model outputs a bivariate prediction $[\mu \quad \sigma]$. The predictive mean $\mu$ is used in a recursive manner to construct the model input at the next time step. The forward computations for the mean channel are performed as described in Sections~(\ref{ILSTM}) and (\ref{INODE}). The predictive variance is obtained directly from the network output and enforced to be positive via a softplus activation function \(\sigma^{\text{softplus}}\) applied to the corresponding output channel.

\subsection{Bayesian Neural Network}
We train a Bayesian NN trained via variational inference \cite{mahajan2024bayesian}. Unlike deterministic networks, the model parameters are treated as random variables with a prior distribution. The prior distribution over the network parameters is defined as a zero-mean Gaussian with fixed variance, i.e., $p(\theta) = \mathcal{N}(0, \rho^2 I)$, where $\rho$ is a fixed hyperparameter controlling the strength of regularization. The Bayesian NN outputs both predictive mean $\mu$ and variance $\bar{\sigma}^2$:
\begin{equation}
[\mu, \bar{\sigma}^2]= g(x; \theta), \quad \theta \sim p(\theta)
\end{equation}
Since the exact posterior $p(\theta \mid \mathcal{D})$, where $\mathcal{D}$ denotes the observed dataset,  is intractable, we approximate it with a variational distribution $q(\theta)$ by minimizing the negative evidence lower bound (ELBO), given by:
\begin{equation}
\mathcal{L}_{\text{BNN}} =
\mathbb{E}_{q(\theta)}[-\log p(y \mid x, \theta)]
+ \mathrm{KL}(q(\theta)\,\|\,p(\theta))
\end{equation}

In practice, the expectation is approximated using Monte Carlo sampling from $q(\theta)$. The first term corresponds to the NLL-related loss, as defined in Eq.~(\ref{nll_loss}) under a Gaussian likelihood assumption. The second term represents the Kullback–Leibler (KL) divergence, which acts as a regularizer by encouraging the variational posterior $q(\theta)$ to remain close to the prior $p(\theta)$ \cite{blundell2015weight}. At test time, predictive uncertainty is estimated using Monte Carlo sampling. Specifically, at each predicton step, we draw $S=100$ samples from the variational posterior $q(\theta)$ and compute the corresponding predictions:
\begin{equation}
\{ \mu^{(s)}, \bar{\sigma}^{2 (s)} \}_{s=1}^{S}, \quad \theta^{(s)} \sim q(\theta)
\end{equation}
The predictive mean and variance are obtained by aggregating these samples:
\begin{equation} \label{z_score}
\hat{\mu} = \frac{1}{S} \sum_{s=1}^{S} \mu^{(s)}, \quad
\hat{\sigma}^2(x) = \frac{1}{S} \sum_{s=1}^{S} \left(\bar{\sigma}^{2 (s)} + \mu^{2 (s)} \right) - \hat{\mu}^2
\end{equation}
PIs are constructed under the Gaussian assumption as:
\begin{equation} \label{pi_zscore}
\left[\underline{y}, \bar{y}\right] = \left[\hat{\mu} - z_{\alpha} \hat{\sigma}, \hat{\mu} + z_{\alpha} \hat{\sigma}\right]
\end{equation}
where $z_{\alpha}$ denotes the standard normal quantile corresponding to the target coverage level (e.g., $z_{0.9} = 1.645$).

\subsection{NN with Monte Carlo Dropout}
 
We employ MC Dropout, which approximates Bayesian inference by enabling dropout at inference time \cite{gal2016dropout}. Unlike Bayesian NNs, NNs with MC Dropout do not require a variational posterior or KL divergence term. During training, dropout is applied to the hidden layers to prevent overfitting. At test time, dropout is kept active, and stochastic forward passes are performed to approximate the predictive distribution.

The predictive model under MC Dropout is defined as:
\begin{equation}
[\mu^{(s)}, \bar{\sigma}^{2(s)}] = g(x; \theta, m^{(s)}),
\end{equation}
where $m^{(s)} \sim \mathrm{Bernoulli}(1-p)$ represents the stochastic dropout mask applied to the hidden layers.

Similar to the Bayesian NN, predictive uncertainty is estimated using $S=100$ Monte Carlo samples at test time. The final predictive mean and variance are computed as defined in Eq.~(\ref{z_score}). PIs are constructed as in Eq.~(\ref{pi_zscore}).

\subsection{Deep Ensemble NN}
  
To benchmark UQ performance, we employ a Deep Ensemble consisting of $M$ independently trained NNs \cite{lakshminarayanan2017simple}. Each network is initialized with different random seeds and trained separately using the same architecture and training procedure. Unlike Bayesian NNs, Deep Ensemble NNs do not rely on variational inference or KL divergence, and uncertainty is captured through diversity among independently trained models. Each ensemble member outputs a predictive mean and variance:
\begin{equation}
[\mu^{(m)}, \bar{\sigma}^{2(m)}] = g(x; \theta_m),
\end{equation}
where $\theta_m$ denotes the parameters of the $m$-th network.

At test time, each model is evaluated independently, and predictive statistics are aggregated over $M$ ensemble members. The final predictive mean is computed as:
\begin{equation}
\hat{\mu} = \frac{1}{M} \sum_{m=1}^{M} \mu^{(m)}.
\end{equation}
The total predictive variance is obtained using the law of total variance:
\begin{equation}
\hat{\sigma}^2 = \frac{1}{M} \sum_{m=1}^{M} \bar{\sigma}^{2(m)} + \mathrm{Var}\left(\mu^{(m)}\right),
\end{equation}
In this study, we use $M=5$ ensemble models. PIs are constructed as in Eq.~\ref{pi_zscore}.

\section{Hyperparameters} \label{AppendixB}

The dataset configurations together with the corresponding model hyperparameters—prior scale $\rho$, dropout rate $m$, and ensemble size $M$—are provided in Table~\ref{tab:dataset_configs}.

\begin{table}[ht!]
\caption{Dataset configurations and NN settings.}

\label{tab:dataset_configs}
\centering
\scriptsize
\setlength{\tabcolsep}{3pt}
\renewcommand{\arraystretch}{1.2}

\begin{tabular}{|l|c|c|c|c|}
\hline
\textbf{} & \textbf{MR-Damper} & \textbf{Heat Exchanger} & \textbf{Hair Dryer} & \textbf{Robot Arm} \\
\hline
$K$ & \(3499\) & \(4000\) & \(1000\) & \(1024\) \\
Train.–Val.–Test (\%) & \(54\text{–}13\text{–}33\) & \(20\text{–}5\text{–}75\) & \(40\text{–}10\text{–}50\) & \(40\text{–}10\text{–}50\) \\
$N$ & \(40\) & \(80\) & \(30\) & \(30\) \\
$n_{\text{step}}$ & \(5\) & \(5\) & \(2\) & \(1\) \\
$n_x - n_d - n_y$ & \(2 - 0 - 1\) & \(2 - 0 - 3\) & \(2 - 2 - 3\) & \(2 - 1 - 3\) \\
\hline
\multicolumn{5}{|l|}{\textbf{INODE}} \\ 
\hline

Number of Layers & \(3\) & \(3\) & \(3\) & \(3\) \\
Hidden Layer Sizes & \(40,40\) & \(40,40\) & \(40,40\) & \(40,40\) \\
$r_{\text{o}} - r_{\text{h}}$ & \(1-1\) & \(1-1\) & \(1-1\) & \(1-1\) \\
$\beta$ & \(0.1\) & \(1\) & \(0.1\) & \(0.1\) \\
\hline
\multicolumn{5}{|l|}{\textbf{ILSTM}} \\ 
\hline

Number of Layers & \(3\) & \(3\) & \(3\) & \(3\) \\
Hidden Layer Sizes & \(30,10\) & \(10,10\) & \(30,10\) & \(10,10\) \\
$r_{\text{o}} - r_{\text{h}}$ & \(1-0.2\) & \(1-0.2\) & \(1-0.2\) & \(1-0.2\) \\
$\beta$ & \(0.5\) & \(1\) & \(1\) & \(0.1\) \\
\hline

\multicolumn{5}{|l|}{\textbf{Bayesian NODE Network }} \\ 
\hline

Number of Layers & \(3\) & \(3\) & \(3\) & \(3\) \\
Hidden Layer Sizes & \(40,40\) & \(40,40\) & \(40,40\) & \(40,40\) \\
$\rho$ & \(1e-1\) & \(1\) & \(1\) & \(1\) \\
\hline

\multicolumn{5}{|l|}{\textbf{Bayesian LSTM Network}} \\ 
\hline
Number of Layers & \(3\) & \(3\) & \(3\) & \(3\) \\
Hidden Layer Sizes & \(30,10\) & \(10,10\) & \(30,10\) & \(10,10\) \\
$\rho$ & \(1e-1\) & \(1\) & \(1e-1\) & \(1e-1\) \\

\hline

\multicolumn{5}{|l|}{\textbf{NODE Network with MC Dropout }} \\ 
\hline

Number of Layers & \(3\) & \(3\) & \(3\) & \(3\) \\
Hidden Layer Sizes & \(40,40\) & \(40,40\) & \(40,40\) & \(40,40\) \\
$m$ & \(5e-2\) & \(5e-2\) & \(5e-2\) & \(5e-2\) \\
\hline

\multicolumn{5}{|l|}{\textbf{LSTM Network with MC Dropout}} \\ 
\hline
Number of Layers & \(3\) & \(3\) & \(3\) & \(3\) \\
Hidden Layer Sizes & \(30,10\) & \(10,10\) & \(30,10\) & \(10,10\) \\
$m$ & \(5e-2\) & \(5e-2\) & \(5e-2\) & \(5e-2\) \\

\hline

\multicolumn{5}{|l|}{\textbf{Deep Ensemble NODE Network}} \\ 
\hline

Number of Layers & \(3\) & \(3\) & \(3\) & \(3\) \\
Hidden Layer Sizes & \(40,40\) & \(40,40\) & \(40,40\) & \(40,40\) \\
$M$ & \(5\) & \(5\) & \(5\) & \(5\) \\
\hline

 \multicolumn{5}{|l|}{\textbf{Deep Ensemble LSTM Network}} \\ 
 \hline
Number of Layers & \(3\) & \(3\) & \(3\) & \(3\) \\
 Hidden Layer Sizes & \(30,10\) & \(10,10\) & \(30,10\) & \(10,10\) \\
$M$ & \(5\) & \(5\) & \(5\) & \(5\) \\

\hline
\end{tabular}
\end{table}

\bibliographystyle{elsarticle-num} 
\bibliography{cites}
\end{document}